\documentclass[fleqn, 10pt]{wlscirep}
\usepackage[utf8]{inputenc}
\usepackage[T1]{fontenc}
\usepackage{newunicodechar}
\newunicodechar{）}{)}
\newunicodechar{，}{,}
\usepackage{colortbl}
\usepackage{bm}
\usepackage{marvosym}

\usepackage{xcolor}

\usepackage{multirow}
\usepackage[htt]{hyphenat}
\usepackage{booktabs}
\usepackage{tabularx}
\usepackage{url}
\usepackage{fancyvrb}
\usepackage{fvextra} 
\usepackage[most]{tcolorbox}
\usepackage{booktabs, threeparttable, adjustbox, caption, float}

\title{A Multimodal Agentic Pathology Co-pilot via Evidence Grounded Reasoning}
\author[1,*]{Zhe Xu}
\author[2,3,*]{Zhengyu Zhang} 
\author[1]{Zhiyuan Cai}
\author[1]{Jiahao Xu}
\author[1]{Yijie Lin}
\author[1]{Ziyi Liu}
\author[1]{Junlin Hou}
\author[1]{Hongyi Wang}
\author[1]{Yuxiang Nie}
\author[1]{Yihui Wang}
\author[1]{Jiabo Ma}
\author[1]{Ling Liang}
\author[1]{Yingxue Xu}
\author[1]{Zhengrui Guo}
\author[1]{Guanghao Wu}
\author[2,3]{Danyi Li} 
\author[2,3]{Ziqi Zhou} 
\author[2,3]{Donglin Tan} 
\author[2,3]{Zhijian Cen} 
\author[2,3]{Ying Tan} 
\author[2,3]{Xiaolin Liu} 
\author[2,3]{Qi Xie} 
\author[6]{Xiaoying Tang} 
\author[7,8]{Xi Peng} 
\author[9]{Cheng Deng}
\author[5]{Lijuan Qu}
\author[4]{Ronald Cheong Kin Chan}
\author[2,3,10,11 \Letter ]{Li Liang}
\author[1,12,13,14,15 \Letter]{Hao Chen}

\affil[1]{Department of Computer Science and Engineering, Hong Kong University of Science and Technology, Hong Kong, China}
\affil[2]{Department of Pathology, Nanfang Hospital, Southern Medical University, Guangzhou, China}
\affil[3]{Department of Pathology, School of Basic Medical Sciences, Southern Medical University, Guangzhou, China} 
\affil[4]{Department of Anatomical and Cellular Pathology, Chinese University of Hong Kong, Hong Kong, China}
\affil[5]{Department of Pathology, The 900th Hospital of Joint Logistic Support Force, Fuzhou, China}
\affil[6]{Department of Electronic and Electrical Engineering, Southern University of Science and Technology, Shenzhen, China}
\affil[7]{School of Artificial Intelligence, Sichuan University, China} 
\affil[8]{Tianfu Jincheng Laboratory, Chengdu, China.}
\affil[9]{College of Information Science and Engineering, Hohai University, Nanjing, China}
\affil[10]{Guangdong Provincial Key Laboratory of Molecular Tumor Pathology, Guangzhou, China}
\affil[11]{Jinfeng Laboratory, Chongqing, China}
\affil[12]{Department of Chemical and Biological Engineering, Hong Kong University of Science and Technology, Hong Kong SAR, China}
\affil[13]{Division of Life Science, Hong Kong University of Science and Technology, Hong Kong SAR, China}
\affil[14]{State Key Laboratory of Nervous System Disorders, The Hong Kong University of Science and Technology, Hong Kong SAR, China}
\affil[15]{HKUST Shenzhen-Hong Kong Collaborative Innovation Research Institute, The Hong Kong University of Science and Technology, Futian, Shenzhen, China}

\affil[*]{Contributed Equally (Co-first)}
\affil[\Letter]{Corresponding Authors}
\affil[ ]{\textbf{Lead Contact: Hao Chen (jhc@ust.hk)}}

\begin{abstract}
Pathology is the cornerstone of modern medicine, where accurate decision-making relies heavily on evidence-based practices. While artificial intelligence (AI) shows high potential in pathology, most existing models generate diagnostic outputs without providing verifiable clinical evidence, which limits clinician trust and hinders real-world translation. To overcome these limitations, we present PathPocket, a multimodal agentic co-pilot designed specifically for evidence grounded pathology. We construct the most comprehensive pathology evidence corpus to date, encompassing approximately 110,472 public and authorized documents structured across a rigorous hierarchy of evidence from clinical guideline to expert opinion. From this meticulously graded foundation, we build a large-scale multimodal pathology hypergraph containing over 4.55 million entities and 7.10 million relations. Serving as a robust knowledge engine, this hypergraph provides traceable evidence for a collaborative multi-agent reasoning framework integrating case understanding, evidence retrieval, evidence filtering, and response generation. This enables PathPocket to seamlessly resolve a wide spectrum of clinical tasks, ranging from text-only queries to complex multimodal diagnosis involving region-of-interest (ROI) and gigapixel whole-slide image (WSI). We rigorously evaluate the system on a multidimensional benchmark of over 21,000 real-world cases, where it significantly outperforms existing state-of-the-art methods. Crucially, a randomized, two-stage crossover reader study involving both senior and junior pathologists demonstrates that PathPocket substantially enhances clinical workflows, yielding a +6.8\% relative improvement in diagnostic accuracy, a +11.6\% boost in clinician confidence, and a +32.6\% relative reduction in the senior-junior performance gap. By grounding pathology reasoning directly in verifiable literature, PathPocket bridges the gap between algorithmic predictions and clinical practice, demonstrating substantial clinical and translational value for evidence-grounded computational pathology.
\end{abstract}

\begin{document}

    \maketitle

\section{Introduction}

\begin{figure*}
    \centering
    \includegraphics[width=0.93\linewidth]{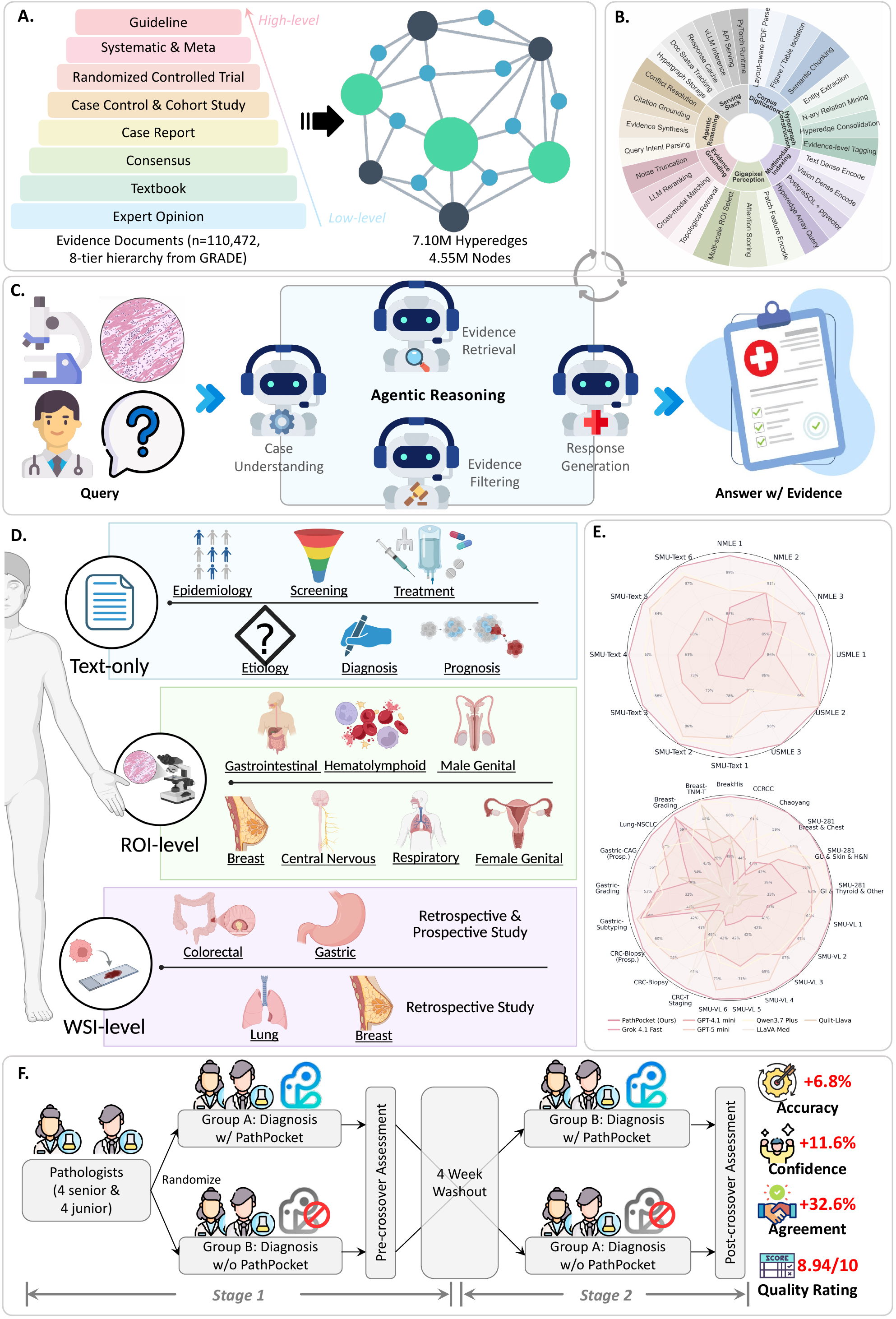}
\end{figure*}

\begin{figure*}
    \centering
\caption{\textbf{Overview of PathPocket, a multimodal AI agentic co-pilot for evidence grounded computational pathology.} \textbf{A.} Pathology evidence hierarchy and hypergraph construction. PathPocket is founded on a comprehensive pathology evidence corpus, structured according to a rigorous hierarchy of medical evidence ranging from clinical guidelines to expert opinions. This graded foundation is used to construct a large-scale multimodal pathology hypergraph that maps complex multi-entity relationships. \textbf{B.} Supporting multimodal infrastructure. Essential tools and utilities, including document parsing, database management (PostgreSQL), vision/text embeddings, patch selection, and reranking mechanisms. This infrastructure provides the necessary foundational support for both the large-scale construction of the pathology hypergraph and the operational execution of the multi-agent reasoning framework. \textbf{C.} Collaborative multi-agent reasoning framework. The clinical workflow where pathologist queries and multimodal inputs (such as region-of-interest or whole-slide images) are processed by a cooperative multi-agent system to generate accurate, evidence grounded diagnostic results. \textbf{D.} Benchmark task taxonomy. Schematic overview of the evaluation suite spanning text-only clinical reasoning, ROI-level multimodal interpretation across anatomical systems, and WSI-level diagnostic endpoints (grading, staging, and subtyping; retrospective and prospective). \textbf{E.} Multidimensional benchmark evaluation. Stacked radar charts comparing PathPocket with state-of-the-art baselines: the upper chart summarizes text-only clinical tasks (pathologist-generated USMLE-style and NMLE-style items, and SMU-Text), and the lower chart summarizes multimodal ROI- and WSI-level tasks, highlighting robust generalization across diverse real-world datasets. \textbf{F.} Human--AI collaboration reader study. A randomized, two-stage crossover reader study evaluating the impact of PathPocket on pathologist performance. A cohort of 8 pathologists (4 senior and 4 junior) was randomized into two groups. In Stage 1, Group A performed diagnoses with the assistance of PathPocket, while Group B diagnosed without the tool. Following a 4-week washout period to mitigate carryover effects, the groups crossed over for Stage 2, where Group B utilized PathPocket and Group A did not. Pre- and post-crossover assessments demonstrate that PathPocket significantly enhances clinical workflows, yielding a +6.8\% relative improvement in diagnostic accuracy, a +11.6\% boost in clinician confidence, a +32.6\% relative reduction in the senior-junior performance gap, and a high overall quality rating of 8.94/10.}
\label{fig:overview}
\end{figure*}

Evidence-based medicine integrates the best available research evidence with clinical expertise and patient values to guide healthcare decisions \cite{sackett1996evidence,guyatt2008grade,howick2011oxford}. Pathology, as the cornerstone of modern medicine, is inherently evidence-driven \cite{hanahan2000hallmarks,hanahan2011hallmarks,nowell1976clonal,marshall1984unidentified,kumar2014robbins,rosai2011rosai,song2023artificial,shmatko2022artificial}. From biopsy triage and frozen-section consultation to definitive surgical reporting, every morphological observation, immunohistochemical interpretation, and molecular finding must be contextualized within graded medical knowledge before it can support diagnosis, treatment selection, or prognosis. In routine practice, this requirement is intensified by rising specimen volumes, morphologic heterogeneity, overlapping histologic patterns, and substantial interobserver variability, especially for diagnostically ambiguous cases. As a result, pathologists routinely need not only visual recognition of tissue patterns, but also rapid, trustworthy access to guidelines, textbooks, and primary literature that justify and constrain their conclusions.

Artificial intelligence (AI) has transformed computational pathology by improving recognition and prediction from histology images. Pathology foundation models now achieve strong performance on various tasks \cite{conch,chief,gpfm,uni,virchow,gigapath,plip,xu2025versatile,xu2025discovering}. Conversational systems such as PathChat \cite{lu2024multimodal} and SlideSeek \cite{chen2025evidence} further enable vision-language interaction over regions of interest (ROIs) and whole-slide images (WSIs). Despite this progress, most pathology AI systems remain weakly grounded in explicit clinical evidence. Their knowledge is stored implicitly in model parameters rather than retrieved from citable sources, so outputs are difficult to audit, hard to update when guidelines change, and vulnerable to hallucinated morphologic or molecular interpretations. This gap is clinically consequential: in pathology, an answer without a traceable evidentiary basis is difficult to trust or to incorporate into regulated diagnostic workflows.

In parallel, evidence-oriented clinical AI tools have begun to couple large language models with medical literature retrieval. Systems such as OpenEvidence \cite{open2024evidence} and the Baichuan-M series \cite{baichuan2024m2} demonstrate that explicit evidence grounding can improve verifiability in general medicine. However, these tools remain largely confined to text-only settings. They are not designed to ingest the multimodal inputs that constitute the diagnostic language of pathology, including multi-image ROIs and gigapixel WSIs, nor do they encode the fine-grained relationships among diseases, histologic patterns, immunohistochemical panels, and molecular alterations that structure pathological reasoning. Consequently, evidence-based AI and multimodal pathology AI have largely evolved as separate lineages, leaving a critical unmet need for systems that jointly support visual interpretation and literature-grounded justification.

More recently, pathology-specific retrieval-augmented frameworks have attempted to close this gap. YpathRAG \cite{wang2024ypathrag} retrieves from a corpus of approximately 1,000 papers using flat chunk embeddings, while Patho-AgenticRAG \cite{zhang2026pathoagenticrag} builds page-level embeddings from about 600 textbooks. These efforts mark an important step toward evidence-aware pathology AI, but several limitations remain. First, their corpora are limited in scale and not stratified by evidence hierarchy. Second, flat chunk or page retrieval cannot readily capture structured multi-entity relations, for example how a candidate diagnosis is jointly constrained by anatomic site, morphologic pattern, and a panel of immunohistochemical markers. Third, validation has typically been restricted to small or single-modality datasets, with limited assessment of real-world clinical utility through pathologist-AI collaboration. Bridging these gaps requires an evidence engine that is large-scale, graded, multimodal, and relational, together with an agentic reasoning framework that can operate across text-only queries, ROI-level interpretation, and WSI-level diagnosis under clinically realistic evaluation.

\begin{figure}[H]
    \centering
    \includegraphics[width=\textwidth]{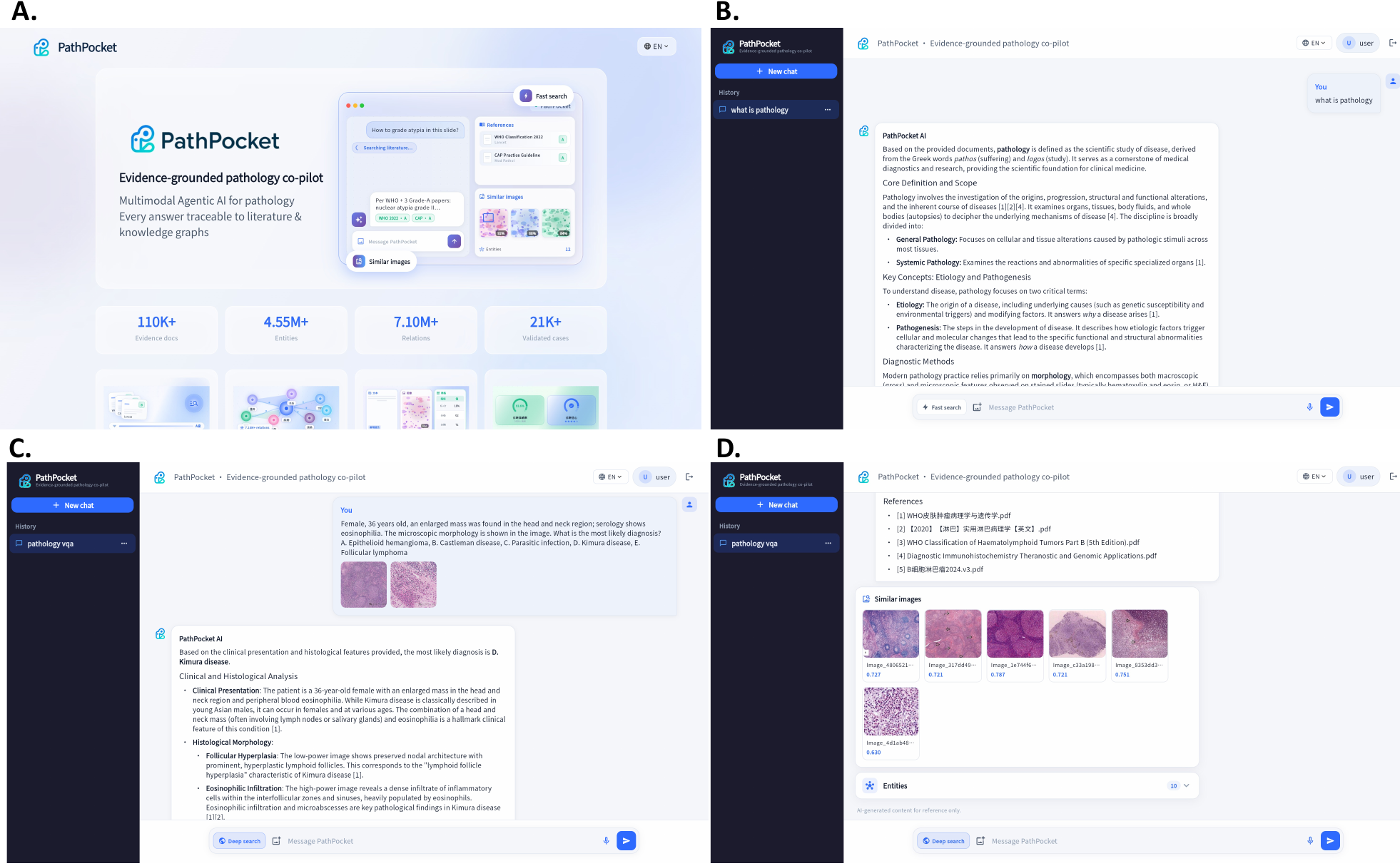}
    \caption{\textbf{Web interface of the PathPocket agentic co-pilot.}
    \textbf{(A)}~Landing page introducing PathPocket as an evidence-grounded multimodal pathology co-pilot, with corpus-scale statistics (evidence documents, entities, relations, and validated cases).
    \textbf{(B)}~Text-only chat: PathPocket returns structured, citation-linked answers to pathology knowledge queries.
    \textbf{(C)}~Multimodal visual question answering: users upload histopathology images with a clinical stem; PathPocket provides a diagnosis and evidence-backed clinical/histological analysis.
    \textbf{(D)}~Evidence panel associated with an assisted answer, including source references, visually similar images with similarity scores, and hypergraph evidences (e.g., pathology entities).}
    \label{fig:PP_website}
\end{figure}

Here we present PathPocket, a multimodal agentic co-pilot for evidence grounded reasoning in computational pathology (Figure~\ref{fig:overview}; Appendix Figure~\ref{fig:PP_website}). First, we curate the most comprehensive pathology evidence corpus assembled to date, containing over 110,000 documents from guidelines, systematic reviews, randomized trials, observational studies, case reports, consensus statements, textbooks, and expert opinions, each tagged with an explicit level of evidence following an eight-tier adaptation of the GRADE hierarchy \cite{guyatt2008grade}. From this graded corpus, we construct a large-scale multimodal hypergraph with 4.55 million entities and 7.10 million relations (Figure~\ref{fig:overview}A). Unlike pairwise knowledge graphs, hyperedges can connect multiple entities simultaneously, enabling PathPocket to model the multivariate clinical relations that underlie differential diagnosis and workup. Second, we establish a multimodal infrastructure for both offline hypergraph construction and online inference (Figure~\ref{fig:overview}B), encompassing document parsing, vision and text embeddings, gigapixel patch selection, and evidence reranking. Third, this infrastructure and hypergraph drive a collaborative multi-agent reasoning framework (Figure~\ref{fig:overview}C) in which specialized agents perform case understanding, evidence retrieval, evidence filtering, and response generation, with every final answer accompanied by direct, verifiable citations.

We systematically evaluate PathPocket on a multidimensional benchmark of more than 21,000 cases spanning 33 clinical tasks across text-only reasoning, ROI-level multimodal interpretation, and WSI-level diagnosis (Figure~\ref{fig:overview}D). Because the multimodal pathology hypergraph is constructed from public and authorized literature as the evidence source, most evaluation tasks are conducted on private clinical data to reduce the risk of evidence leakage into the test queries. Across these settings, PathPocket substantially outperforms strong general-purpose and pathology-specific baselines (Figure~\ref{fig:overview}E). Beyond automated benchmarking, a randomized two-stage crossover reader study with senior and junior pathologists shows that PathPocket assistance improves diagnostic accuracy by 6.8\%, increases clinician confidence by 11.6\%, and strengthens senior-junior performance gap by 32.6\%, with a high content-quality rating of 8.94/10 (Figure~\ref{fig:overview}F). Collectively, these results establish PathPocket as a practical pathway toward evidence grounded computational pathology, in which pathology reasoning is inseparable from graded, citable medical knowledge.

\section{Results}

In this study, we develop and comprehensively evaluate PathPocket, an evidence grounded multimodal agentic co-pilot designed for computational pathology workflows. Our results are structured around four core pillars. First, we construct a large-scale, evidence-graded multimodal pathology hypergraph, establishing a rigorous and traceable knowledge engine that forms the foundation of the system. Second, we rigorously evaluate PathPocket's diagnostic reasoning capabilities across an exhaustive benchmark of 33 distinct clinical tasks. This benchmark spans text-only clinical queries, multimodal ROI interpretations, and complex gigapixel WSI diagnosis, across which PathPocket consistently outperforms state-of-the-art baselines. Because the hypergraph evidence corpus is derived from public and authorized documents, the majority of these tasks use private hospital cohorts to mitigate potential data leakage between the evidence base and the evaluation set. Third, detailed qualitative analyses demonstrate PathPocket's ability to effectively mitigate AI hallucinations by providing transparent, evidence-backed diagnoses (Figure~\ref{fig:qualitative_examples}; Appendix Figure~\ref{fig:qualitative_full}). 
Finally, a randomized crossover reader study involving board-certified pathologists confirmed the practical clinical utility of the system, demonstrating substantial improvements in real-world diagnostic accuracy, user confidence, and agreement. 
The detailed findings for each of these pillars are presented in the following sections.

\begin{figure}[htbp]
    \centering
    \includegraphics[width=\textwidth]{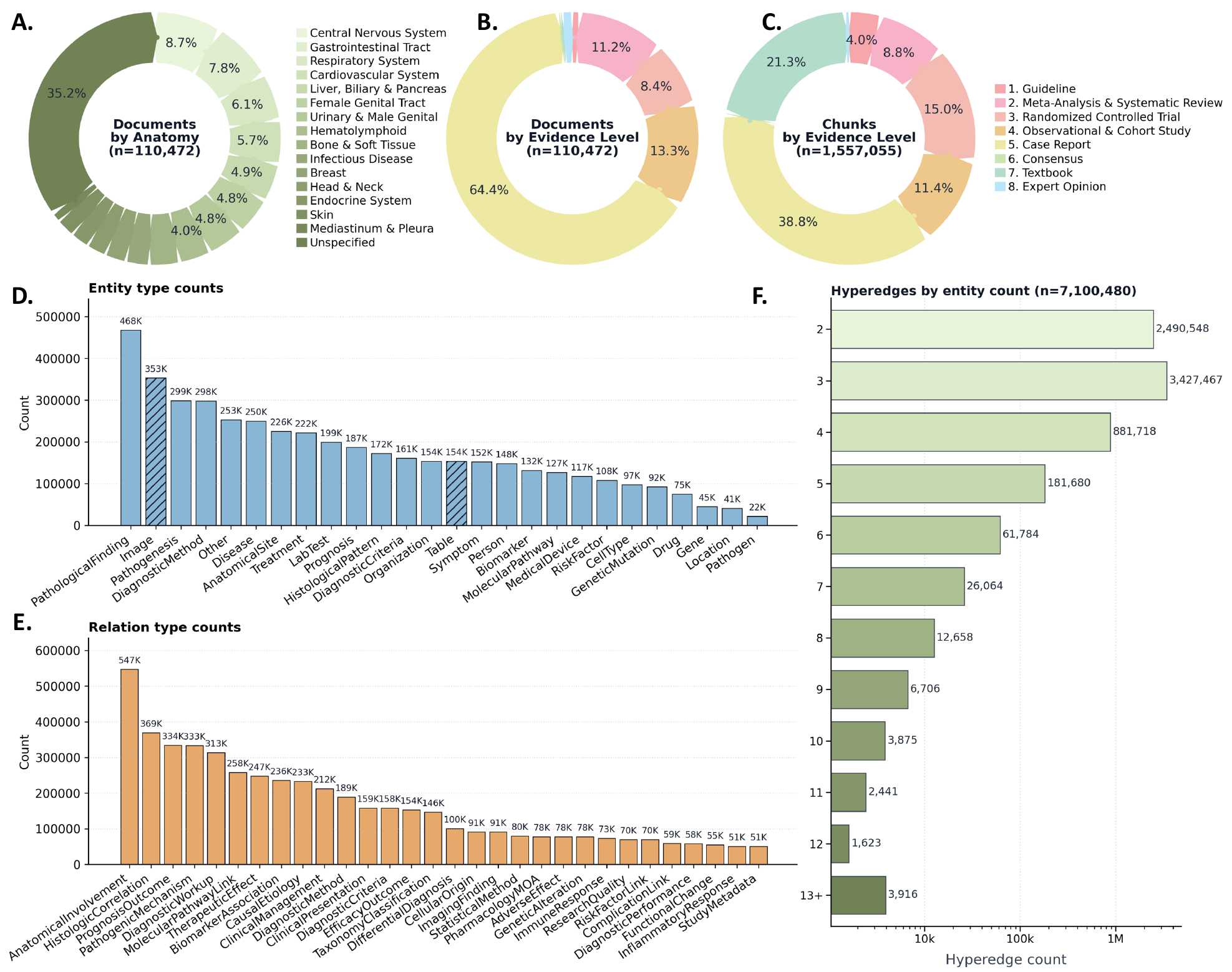} 
    \caption{\textbf{Construction and characterization of the large-scale multimodal pathology evidence hypergraph.}
    \textbf{A.} Distribution of the curated 110,472 source documents across various anatomical systems, demonstrating comprehensive coverage of human pathology.
    \textbf{B--C.} Distribution of medical evidence levels (graded according to an 8-tier hierarchy) at the document level (B) and the parsed semantic chunk level (C). While Case Reports (Level 5) constitute the majority of individual documents (64.4\%), dense sources such as Textbooks (Level 7) and Guidelines (Level 1) yield a disproportionately large share of the 1,557,055 semantic knowledge chunks (21.3\% and 4.0\%, respectively).
    \textbf{D.} Distribution of extracted entity types within the hypergraph. Crucially, multimodal elements such as ``Image'' ($n>350,000$) and ``Table'' are integrated with clinical entities within the hypergraph.
    \textbf{E.} Distribution of the relation types connecting the entities. The network is heavily populated with clinically actionable relationships, forming a densely connected knowledge topology. \textbf{F.} Hyperedges by entity count. Distribution of the total 7,100,480 hyperedges categorized by the number of connected entities per hyperedge. The x-axis is shown on a log scale.}
    \label{fig:hypergraph_stats}
\end{figure}

\begin{figure*}
    \centering
    \includegraphics[width=1\linewidth]{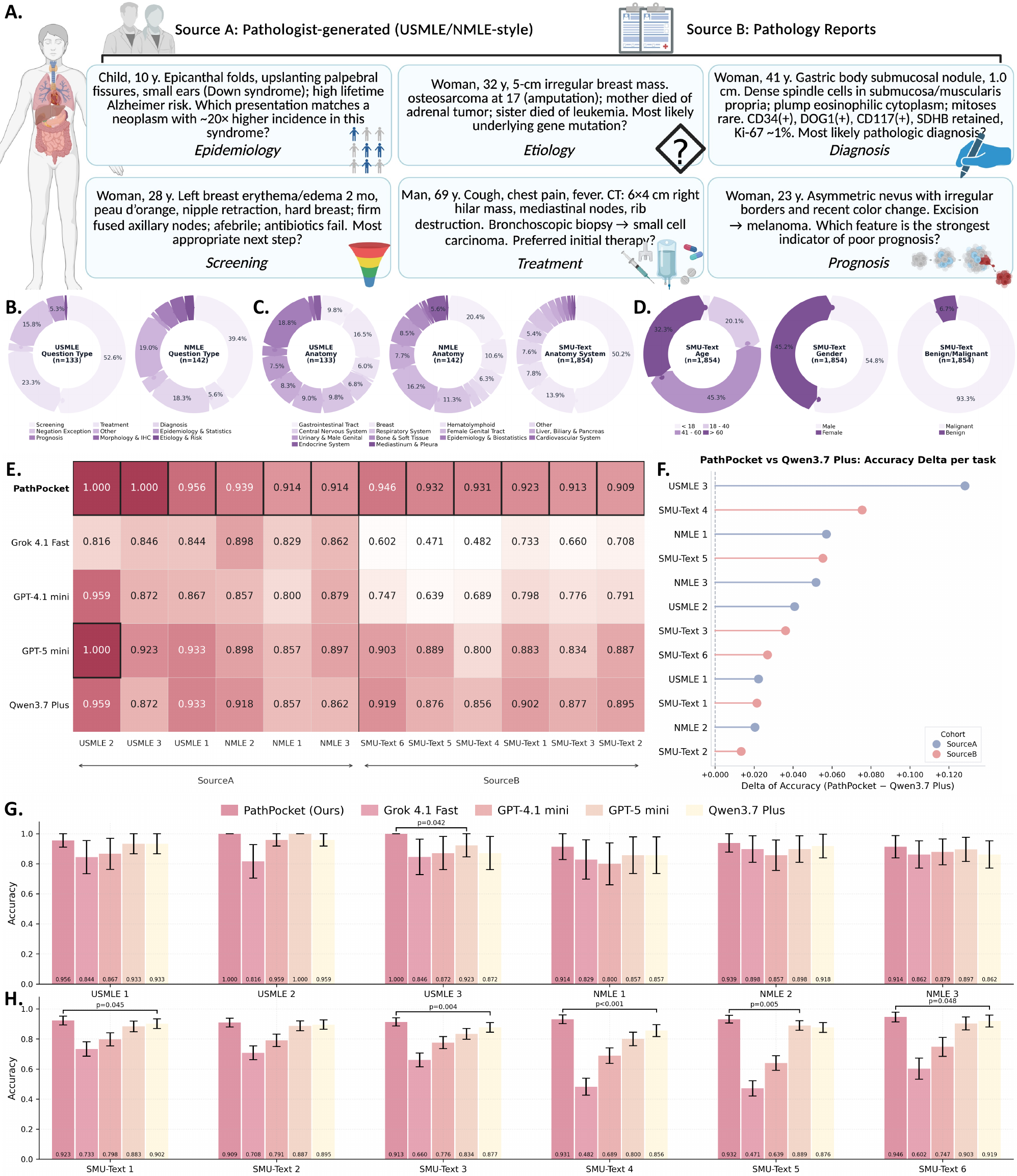}
\end{figure*}

\begin{figure*}
    \centering
\caption{\textbf{Dataset characteristics and performance evaluation of PathPocket on text-only pathology benchmarks.} \textbf{A.} Dual-source construction of the text-only benchmark, spanning pathologist-generated USMLE/NMLE-style items (Source~A) and questions derived from real-world pathology reports (Source~B; SMU-Text), with representative examples across epidemiology, etiology, screening, treatment, diagnosis, and prognosis. \textbf{B.} Question-type distributions for the pathologist-generated USMLE-style ($n=133$) and NMLE-style ($n=142$) items. \textbf{C.} Anatomical-system distributions for the pathologist-generated USMLE-style and NMLE-style items and the private SMU-Text cohort ($n=1,854$). \textbf{D.} SMU-Text demographic and clinical characteristics, including age, gender, and benign/malignant status. \textbf{E.} Accuracy heatmap across twelve text-only tasks grouped into Source~A (USMLE/NMLE-style) and Source~B (SMU-Text). Within each cohort group, columns are sorted by PathPocket accuracy (descending); black outlines mark the best-performing model(s) in each column. \textbf{F.} Per-task accuracy delta of PathPocket relative to Qwen3.7 Plus, colored by cohort (Source~A vs.\ Source~B). \textbf{G.} Accuracy comparison on six pathologist-generated USMLE-style and NMLE-style subsets (USMLE~1--3 and NMLE~1--3). \textbf{H.} Accuracy comparison on six independent SMU-Text sub-tasks (SMU-Text~1--6). Error bars represent 95\% confidence intervals, and $p$-values indicate statistical significance between PathPocket and the strongest baseline in each task.}
\label{fig:text_benchmark}
\end{figure*}

\subsection{Construction and characterization of a large-scale multimodal pathology hypergraph}

The foundation of PathPocket's reasoning capabilities is a meticulously curated, evidence grounded knowledge base. To overcome the limitations of standard large language models that often hallucinate medical facts or lack verifiable traceability, we construct the most comprehensive multimodal pathology hypergraph assembled to date. 

We initiate this pipeline by collating a large-scale corpus of 110,472 public or authorized pathology documents. As shown in Figure~\ref{fig:hypergraph_stats}A, this corpus encompasses a broad spectrum of human anatomy, ensuring comprehensive coverage across specialized subfields (e.g., Central Nervous System, Gastrointestinal Tract, Respiratory System) as well as general pathological principles. To support evidence-based pathology reasoning, every document is rigorously tagged with an explicit evidence level based on an 8-tier adaptation of the GRADE hierarchy (Figure~\ref{fig:hypergraph_stats}B).
To enable fine-grained evidence retrieval and traceable reasoning, we parse the corpus into 1,557,055 distinct semantic chunks. This conversion reveals a critical characteristic of medical knowledge density (Figure~\ref{fig:hypergraph_stats}C): although individual case reports dominate the raw document count (64.4\%), Textbooks and RCTs are highly dense, generating a disproportionately large-scale volume of semantic chunks (21.3\% and 15.0\%, respectively). 

From these graded semantic chunks, we extract a complex topology comprising 4.55 million entities and 7.10 million multi-entity relations. As illustrated in Figure~\ref{fig:hypergraph_stats}D, multimodal entities ``Image'' and ``Table'' are integrated with text-based clinical entities such as ``Pathological Finding'', ``Diagnostic Method'', and ``Disease.'' 
These nodes are interconnected by clinically meaningful edges. The relation type distribution (Figure~\ref{fig:hypergraph_stats}E) demonstrates a high density of actionable medical linkages, predominantly featuring ``AnatomicalInvolvement'', ``HistologicCorrelation'', ``PrognosisOutcome'', and ``PathogenicMechanism''. Notably, unlike traditional graphs that are restricted to binary relations, a hypergraph naturally accommodates multivariate relationships by allowing a single hyperedge to connect an arbitrary number of entities. We present the distribution of hyperedges based on their entity counts in Figure~\ref{fig:hypergraph_stats}F. Out of the total 7,100,480 hyperedges, while binary relations account for a substantial portion (2,490,548), the vast majority of interactions involve three or more entities. Ternary relations (entity count =3) form the largest group with 3,427,467 hyperedges and a significant number of complex, higher-order relationships span across 4 to over 13 entities. This large-scale, graded, and multimodal hypergraph serves as the robust knowledge engine that directly powers PathPocket's multi-agent diagnostic reasoning.

\subsection{PathPocket excels in text-only clinical reasoning through evidence grounding}

To evaluate the fundamental pathology reasoning and pathology knowledge retrieval capabilities of PathPocket, we first benchmark our system on a comprehensive suite of 12 text-only pathology tasks. These tasks are carefully curated to encompass both examination-style clinical vignettes and complex real-world clinical scenarios. The benchmark includes six pathologist-generated subsets in the style of the United States Medical Licensing Examination (USMLE~1--3) and the China National Medical Licensing Examination (NMLE~1--3), alongside six private datasets originating from a large-scale real-world clinical cohort (SMU-Text~1--6). 

As illustrated in Figure~\ref{fig:text_benchmark}A--D, the text-only benchmark combines pathologist-generated USMLE/NMLE-style items (Source~A) with report-derived SMU-Text questions (Source~B). These USMLE-style and NMLE-style items cover diverse question types, with screening the most frequent category, and a broad spectrum of anatomical systems (Figure~\ref{fig:text_benchmark}B,C). To ensure our evaluation reflects actual clinical practice, we assemble the private SMU-Text dataset comprising 1,854 real-world pathology cases. This private cohort is highly heterogeneous, featuring a near-balanced gender distribution (54.8\% male), a wide age range (predominantly 41--60 years), a high prevalence of malignant cases (93.3\%), and comprehensive anatomical coverage dominated by gastrointestinal tract cases (50.2\%; Figure~\ref{fig:text_benchmark}C,D). We compare PathPocket against the state-of-the-art large language models Grok~4.1 Fast, GPT-4.1 mini, GPT-5 mini, and Qwen3.7 Plus.

Across the full text-only tasks, the sorted accuracy heatmap (Figure~\ref{fig:text_benchmark}E) shows PathPocket ranking first (or tied for first) on nearly all USMLE/NMLE-style and SMU-Text tasks, and the corresponding per-task deltas versus Qwen3.7 Plus are uniformly positive (Figure~\ref{fig:text_benchmark}F). On the six pathologist-generated USMLE-style and NMLE-style subsets (Figure~\ref{fig:text_benchmark}G), PathPocket demonstrates robust clinical knowledge, consistently achieving high accuracy across all tasks. Notably, PathPocket reaches 100\% on USMLE~3, exceeding GPT-5 mini (92.3\%, $p<0.05$) and Qwen3.7 Plus (87.2\%).
The advantages of PathPocket's evidence-based agentic framework become even more pronounced on the six private, real-world clinical tasks (Figure~\ref{fig:text_benchmark}H). Real-world pathology reports and queries are complex, often containing highly specialized information that challenges the parametric memory of standard LLMs. Across all six SMU-Text subsets, PathPocket consistently outperforms all baselines. For instance, on SMU-Text~4, PathPocket achieves an accuracy of 93.1\%, significantly surpassing Qwen3.7 Plus (85.6\%, $p<0.001$). Similar statistically significant improvements are observed in SMU-Text~1 ($p<0.05$), SMU-Text~3 ($p<0.05$), SMU-Text~5 ($p<0.05$), and SMU-Text~6 ($p<0.05$). These results demonstrate that integrating a rigorous evidence hypergraph with a collaborative multi-agent reasoning framework enables PathPocket to resolve complex clinical queries with greater accuracy and reliability.

\begin{figure}[htbp]
    \centering
    \includegraphics[width=\textwidth]{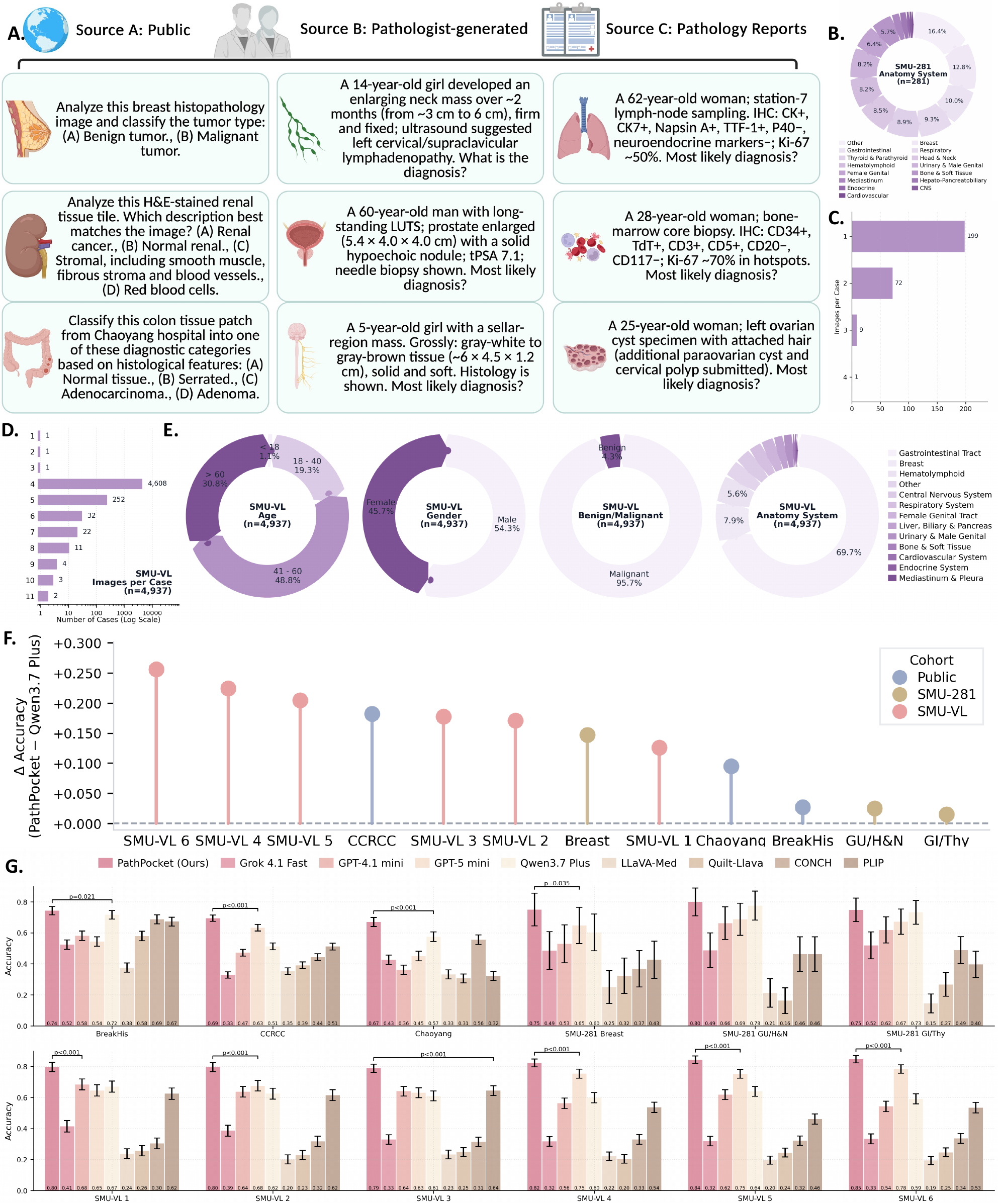} 
\end{figure}

\begin{figure}[htbp]
    \centering
    \caption{\textbf{Performance of PathPocket on ROI-level multimodal clinical pathology benchmarks.}
    \textbf{A.} Triple-source construction of the ROI benchmark, spanning public VQA datasets (Source~A: BreakHis, CCRCC, Chaoyang), pathologist-generated clinical items (Source~B: SMU-281), and questions derived from real-world pathology reports (Source~C: SMU-VL), with representative multimodal prompts.
    \textbf{B--C.} Characteristics of the private SMU-281 cohort ($n=281$), curated by 15 expert pathologists: anatomical-system distribution (B) and number of ROI images per case (C).
    \textbf{D--E.} Characteristics of the large-scale private SMU-VL cohort ($n=4,937$), with tasks formulated directly from pathology reports: images-per-case distribution on a log scale, with most cases containing 4 images (D); and demographic/clinical distributions including age, gender, benign/malignant status, and anatomical systems (E).
    \textbf{F.} Per-task accuracy delta of PathPocket relative to Qwen3.7 Plus across public, SMU-281, and SMU-VL cohorts.
    \textbf{G.} Accuracy comparison across twelve ROI-level tasks arranged in two rows (public datasets and SMU-281 organ strata on top; SMU-VL~1--6 below). PathPocket is compared with general MLLMs, medical-specific MLLMs (LLaVA-Med and Quilt-LLaVA), and pathology foundation models (CONCH and PLIP). For CONCH and PLIP, each task shows the higher-accuracy zero-shot prompting mode between option-only (O) and question+option (Q+O); full results for both modes are provided in Appendix Tables~\ref{tab:bench:roi_public}, \ref{tab:bench:roi_smu281}, and \ref{tab:bench:roi_smuvl}. Error bars represent 95\% confidence intervals, and $p$-values denote significance versus the strongest baseline in each task.}
    \label{fig:roi_benchmark}
\end{figure}

\subsection{PathPocket demonstrates superior multimodal reasoning on ROI-level tasks}

In clinical practice, pathology is inherently multimodal: accurate diagnosis cannot rely on text alone, but fundamentally requires fusing clinical data with the visual interpretation of tissue morphology. To evaluate PathPocket's capability in multimodal clinical reasoning, we construct a rigorous region-of-interest (ROI) level benchmark comprising 12 distinct visual-language tasks. This benchmark includes three widely used public datasets and nine challenging private tasks designed to closely mimic real-world pathology workflows.

We first assess PathPocket on three public multimodal datasets: BreakHis (breast cancer), CCRCC (renal cell carcinoma), and Chaoyang (colon cancer). As shown in Figure~\ref{fig:roi_benchmark}A and \ref{fig:roi_benchmark}G, we compare our system against state-of-the-art MLLMs as well as two pathology foundation models, CONCH\cite{conch} and PLIP\cite{plip}. Because CONCH and PLIP are dual-encoder models evaluated under option-only (O) and question+option (Q+O) zero-shot prompts, main-text figures show the higher-accuracy mode for each model and task, with both modes reported in Appendix Tables~\ref{tab:bench:roi_public}, \ref{tab:bench:roi_smu281}, and \ref{tab:bench:roi_smuvl}. PathPocket leads on all the datasets. For instance, on BreakHis, PathPocket reaches an accuracy of 74.3\%, significantly outperforming the best baseline, Qwen3.7 Plus (71.6\%, $p<0.05$), and exceeding the medical-specific Quilt-LLaVA (58.0\%). On CCRCC and Chaoyang, PathPocket attains 69.5\% and 67.0\% and significantly surpasses baselines ($p<0.001$). These findings suggest that pathology reasoning benefits substantially from explicit evidence grounding beyond generic visual instruction tuning.

Furthermore, we introduce 9 private tasks divided into two distinct cohorts. The first cohort, SMU-281, consists of three highly challenging organ strata meticulously curated by a board of 15 expert pathologists. This dataset features a diverse distribution of anatomical systems (Figure~\ref{fig:roi_benchmark}B) and varied numbers of input images per case (Figure~\ref{fig:roi_benchmark}C). When evaluated on these expert-crafted questions (Figure~\ref{fig:roi_benchmark}G), PathPocket consistently outperforms all baselines across the Breast \& Chest, GU/Skin/H\&N, and GI/Thyroid/Other strata. In the overall evaluation, PathPocket achieves an accuracy of 76.0\%, significantly higher than the top baseline Qwen3.7 Plus (71.0\%, $p<0.05$), supporting its capacity to address complex expert-level pathology queries. Complementary retrieval ablations on SMU-281 further isolate the contribution of hypergraph multi-entity evidence relative to naive chunk retrieval and a no-retrieval backbone (Appendix~\ref{sec:appendix_ablation}; Appendix Figure~\ref{fig:ablation_smu281}).

The second private cohort, SMU-VL, comprises six sub-tasks formulated directly from 4,937 real-world pathology reports, representing the daily diagnostic burden of a pathology department. This dataset is characterized by cases containing multiple ROI images (predominantly 4 images per case, Figure~\ref{fig:roi_benchmark}D) and covers a highly representative clinical demographic (Figure~\ref{fig:roi_benchmark}E), with a strong concentration in gastrointestinal cases (69.7\%). Per-task deltas versus Qwen3.7 Plus are positive across public, SMU-281, and SMU-VL cohorts (Figure~\ref{fig:roi_benchmark}F), and the combined accuracy bar plots (Figure~\ref{fig:roi_benchmark}G) show PathPocket marking the best outcome on nearly all tasks. Across all six report-derived tasks (SMU-VL~1 to 6), PathPocket maintains stable and high performance, consistently edging out the baselines. Notably, on SMU-VL~6, PathPocket achieves an accuracy of 84.7\% compared with GPT-5 mini at 78.4\% ($p<0.001$). 

Together, these results across 12 diverse multimodal tasks demonstrate that PathPocket integrates visual morphology with text-based clinical queries, leveraging its evidence hypergraph to deliver accurate multimodal diagnostic reasoning.

\begin{figure}[]
    \centering
    \includegraphics[width=\textwidth]{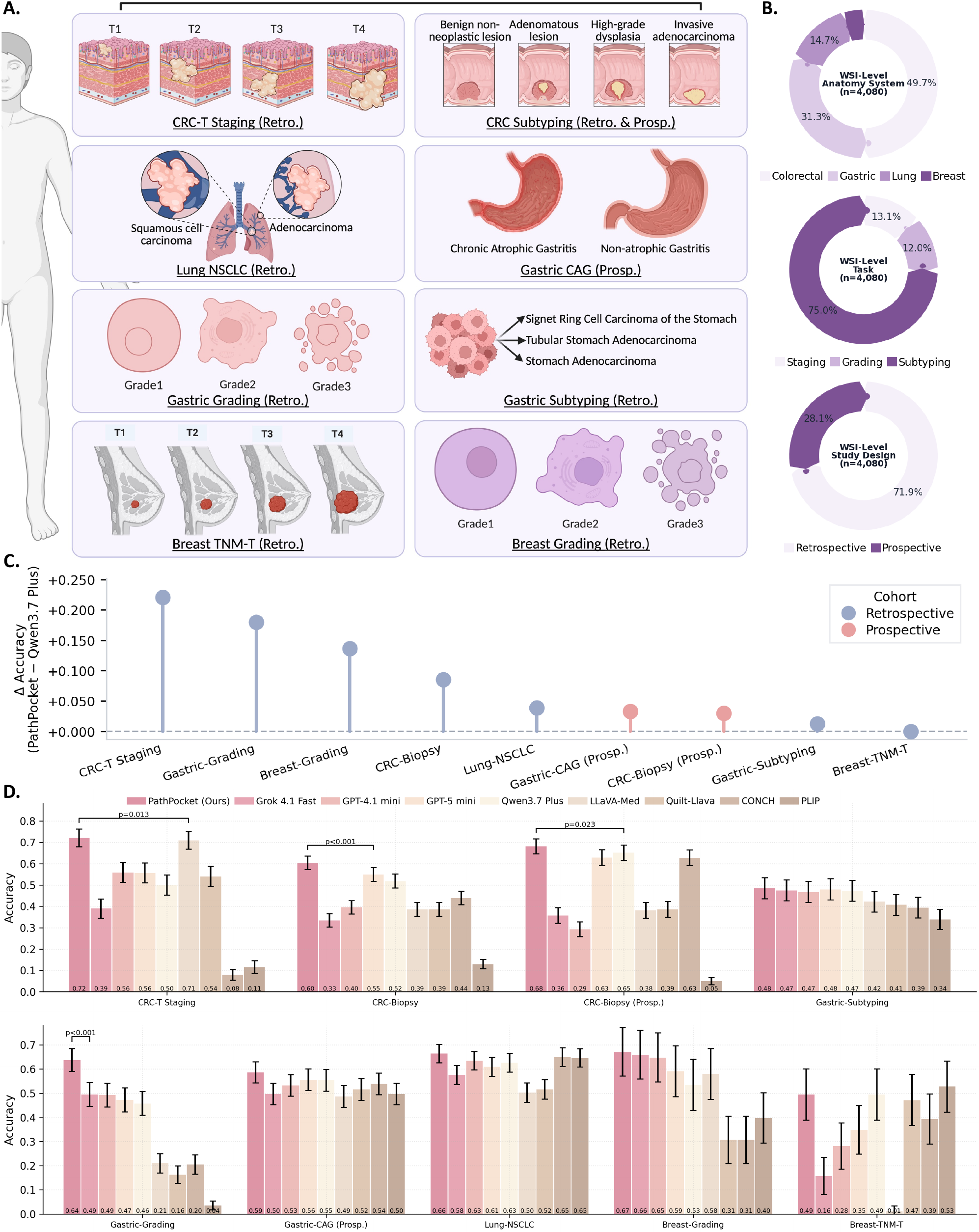} 
\end{figure}

\begin{figure}[]
    \centering
    \caption{\textbf{Performance of PathPocket on gigapixel whole-slide image (WSI) multimodal clinical tasks.}
    \textbf{A.} Schematic overview of the nine WSI-level diagnostic endpoints, spanning colorectal, gastric, lung, and breast cancers and covering tumor grading, staging, and subtyping under retrospective and prospective study designs.
    \textbf{B.} Characteristics of the private WSI-level clinical dataset ($n=4,080$): anatomical-system distribution (colorectal, gastric, lung, and breast), task-type distribution (subtyping, staging, and grading), and study-design split (retrospective vs.\ prospective).
    \textbf{C.} Per-task accuracy delta of PathPocket relative to Qwen3.7 Plus across retrospective and prospective WSI cohorts.
    \textbf{D.} Accuracy comparison between PathPocket and baseline models across the same nine WSI-level tasks. For CONCH and PLIP, each task shows the higher-accuracy zero-shot prompting mode between option-only (O) and question+option (Q+O); full results for both modes are provided in Appendix Table~\ref{tab:bench:wsi_private}. Error bars represent 95\% confidence intervals, and $p$-values denote significance versus the strongest baseline in each task.}
    \label{fig:wsi_benchmark}
\end{figure}

\subsection{PathPocket scales evidence-based reasoning to gigapixel whole-slide images}

While ROI evaluation demonstrates core multimodal capabilities, real-world pathology workflows require navigating and interpreting gigapixel whole-slide images (WSIs). WSI analysis remains challenging for conventional MLLMs owing to the scale of visual information and the need to integrate localized morphological features into a coherent diagnostic interpretation. To evaluate PathPocket's utility at this scale, we curate a private dataset of 4,080 WSIs encompassing nine diagnostic tasks.

As detailed in Figure~\ref{fig:wsi_benchmark}A,B, this WSI cohort covers major oncology domains, including gastric, colorectal, lung, and breast cancers, and requires complex clinical determinations such as tumor grading, staging, and subtyping. Crucially, alongside a robust retrospective majority (74.6\%), the dataset includes prospective clinical subsets (25.4\%), enabling evaluation under prospective real-world clinical conditions.

We benchmark PathPocket against general MLLMs, LLaVA-Med, Quilt-LLaVA, CONCH, and PLIP across the nine WSI-level tasks (Figure~\ref{fig:wsi_benchmark}C,D). As in the ROI evaluation, CONCH and PLIP are shown in main-text figures using the higher-accuracy prompting mode per task, with complete (O) and (Q+O) results in Appendix Table~\ref{tab:bench:wsi_private}. Per-task deltas versus Qwen3.7 Plus are positive on nearly all endpoints (Figure~\ref{fig:wsi_benchmark}C), with the largest gains on CRC-T fine staging and gastric grading, while the corresponding bar plots (Figure~\ref{fig:wsi_benchmark}D) quantify absolute accuracies and significance. PathPocket achieves superior performance on diverse specific tasks. For instance, in retrospective gastric cancer grading, PathPocket achieves an accuracy of 63.8\%, significantly surpassing GPT-4.1 mini (49.2\%, $p<0.05$). 
When evaluated on prospective CRC biopsy diagnosis, PathPocket maintains an accuracy of 68.1\%, exceeding Qwen3.7 Plus at 65.1\% ($p<0.05$). Even in breast cancer grading, a task characterized by high inter-observer variability, PathPocket maintains an accuracy of 67.0\%. 

These results suggest an important advantage of evidence-guided multimodal reasoning: by utilizing its TITAN-guided multi-scale patch-selection mechanism (Appendix~\ref{sec:appendix_wsi_patch}) together with an evidence-grounded hypergraph, PathPocket effectively bridges the gap between massive WSI visual contexts and precise clinical decision-making, enabling more reliable reasoning over gigapixel pathology images than conventional parametric vision-language models.

\begin{figure}[]
    \centering
    \includegraphics[width=0.95\textwidth]{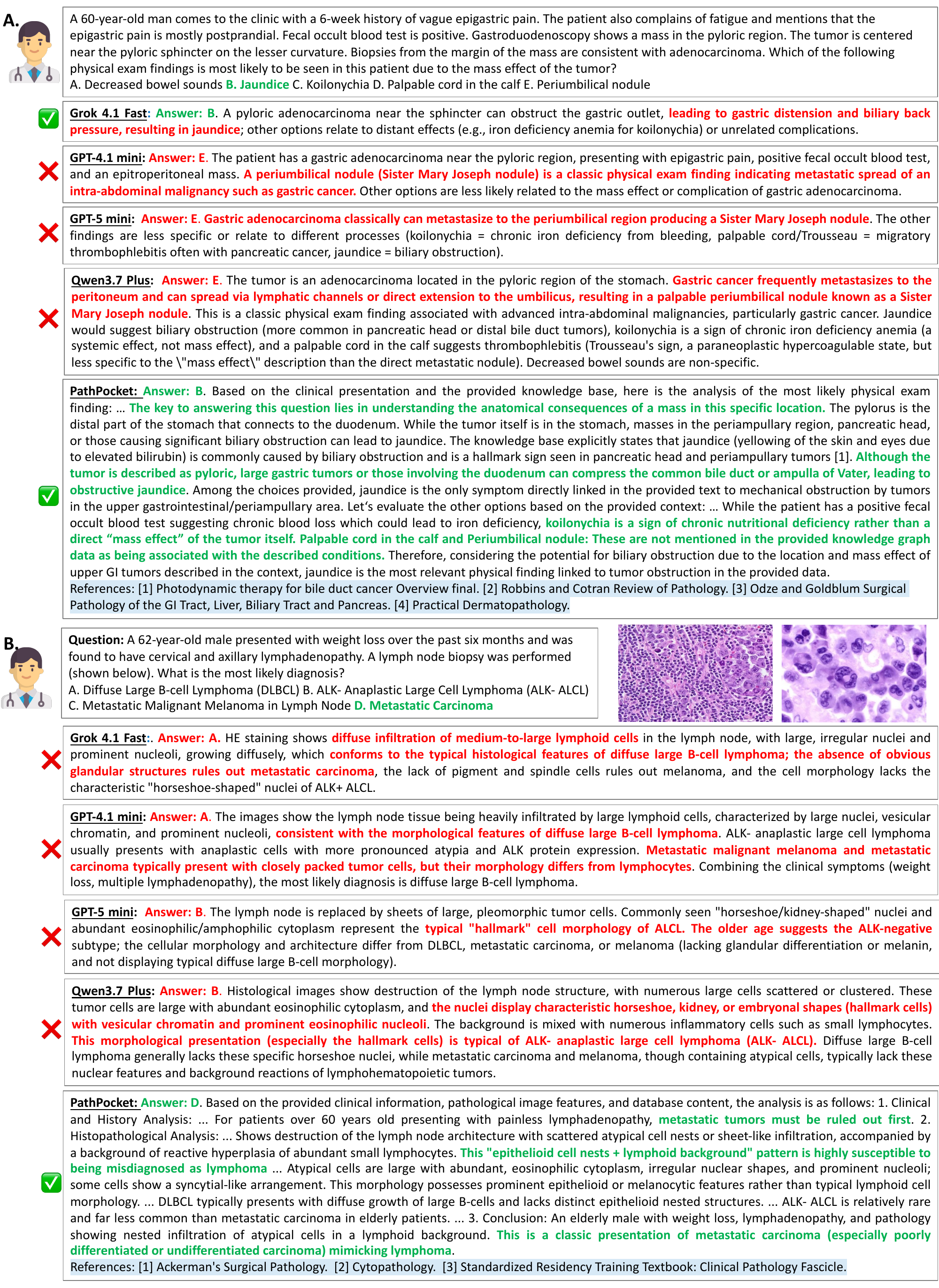}
\end{figure}

\begin{figure}[]
    \centering
    \caption{\textbf{Qualitative comparison of diagnostic reasoning between multiple standard baselines and the evidence-grounded PathPocket.}
    \textbf{A.} Text-only clinical case (pyloric mass): Three of the four baseline models fail, confidently hallucinating a periumbilical metastasis (Sister Mary Joseph nodule) to incorrectly justify Gastric Adenocarcinoma (E). In contrast, PathPocket correctly deduces the anatomical consequences, identifying Jaundice (B) as the direct result of the pyloric mass compressing the common bile duct, fully grounded in retrieved GI tract pathology reference books.
    \textbf{B.} Multimodal clinical case (lymph node lesion): Given microscopic ROI images showing an ``epithelioid cell nests + lymphoid background'' pattern, the baseline models fall into distinct diagnostic traps, with two misdiagnosing it as Diffuse Large B-cell Lymphoma (A) and the other two as Anaplastic Large Cell Lymphoma (B). PathPocket accurately decodes the visual features to correctly identify a Metastatic Carcinoma (D), explicitly refuting the lymphoma differentials by citing standard surgical pathology textbooks.
    Complete PathPocket response transcripts for both cases are provided in Appendix Figure~\ref{fig:qualitative_full}.}
    \label{fig:qualitative_examples}
\end{figure}

\subsection{Qualitative analysis highlights the necessity of traceable evidence in complex diagnostics}

To understand the mechanics behind PathPocket's superior quantitative performance and to evaluate its clinical interpretability, we conduct a qualitative analysis comparing its reasoning trajectories against a cohort of standard baseline models. Standard MLLMs rely exclusively on internal parametric memory, which frequently leads to collective hallucinations and uniform diagnostic blind spots when confronted with complex pathology cases. PathPocket mitigates this vulnerability by grounding its answers in retrieved, explicit medical evidence.

Figure~\ref{fig:qualitative_examples}A illustrates a highly specific text-only diagnostic challenge involving a mass located in the pyloric region. The cohort of standard baseline models fails to reason through the spatial anatomy correctly. All four baselines repeatedly hallucinate a periumbilical metastasis (Sister Mary Joseph nodule) to incorrectly select Option E (Gastric Adenocarcinoma), despite no such physical examination finding being provided in the case description. PathPocket, leveraging its Evidence Retrieval Agent, avoids this collective trap and correctly identifies the primary clinical manifestation as Jaundice (Option B). It solves the case by accurately reasoning through the anatomical consequences of a mass in this specific location, noting that large gastric tumors or those involving the duodenum can compress the common bile duct or the ampulla of Vater, leading to obstructive jaundice. Most importantly, PathPocket appends traceable references, allowing clinicians to instantly verify the anatomical logic applied.

The advantages of evidence-grounding extend to multimodal tasks, as demonstrated in Figure~\ref{fig:qualitative_examples}B, where a lymph node biopsy is presented with accompanying ROI images. The baseline models fall into two distinct diagnostic pitfalls due to the dense lymphoid background: the first two baselines superficially align the morphology with Diffuse Large B-cell Lymphoma (DLBCL, Option A), while the remaining two misinterpret the cellular features as Anaplastic Large Cell Lymphoma (ALCL, Option B). PathPocket, however, accurately identifies the distinct ``epithelioid cell nests within a lymphoid background'' pattern and recognizes that this specific presentation is highly susceptible to being misdiagnosed as lymphoma. It systematically rules out these lymphoma differentials by noting the absence of hallmark horseshoe/kidney-shaped cells, correctly arriving at the diagnosis of Metastatic Carcinoma (Option D). By explicitly citing top-tier evidence, PathPocket provides traceable, evidence-supported diagnostic reasoning that mimics a seasoned pathologist.

These qualitative examples underscore that PathPocket does not merely act as a standalone answer generator; it functions as an authentic clinical co-pilot, enhancing diagnostic safety and building user trust through rigorous, evidence-backed reasoning that remains robust where multiple baseline models fail. The corresponding full PathPocket responses, including stepwise differential reasoning and cited references for both cases, are shown in Appendix Figure~\ref{fig:qualitative_full}.

\begin{figure}[]
    \centering
    \includegraphics[width=\textwidth]{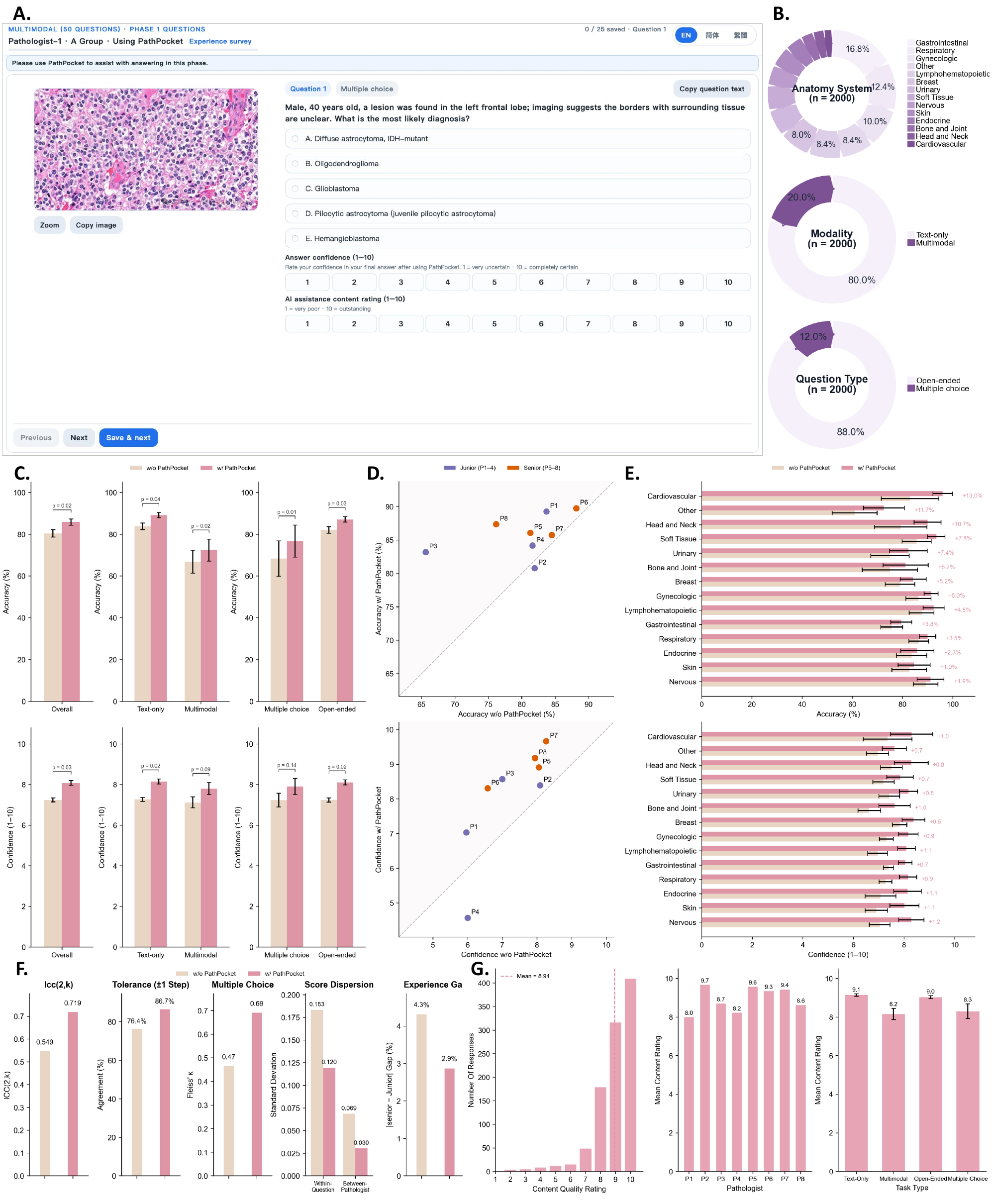} 
\end{figure}

\begin{figure}[]
    \centering
    \caption{\textbf{PathPocket reader study: study interface, question composition, and comparative outcomes with versus without PathPocket assistance.} Eight board-certified pathologists (four junior, P1–P4; four senior, P5–P8) complete a crossover reader study in which each participant answered matched question sets under two assistance conditions: without PathPocket (w/o; diagnosis independently or using other search tools) and with PathPocket (w/). The item bank comprised 150 unique questions (100 text-only and 50 multimodal). Each text-only question contains two independently scored sub-questions and is therefore treated as two scoring units (100$\times$2 = 200), which together with the 50 multimodal items yields 250 scored items and 2,000 pathologist-level response records (250$\times$8). Answer quality was evaluated with GPT-5 mini–based scoring of normalized accuracy (0–100\%) and self-reported confidence (1–10). Bars show condition means with error bars; paired one-sided t-tests were performed on pathologist-level means where indicated.
\textbf{A.} The web-based reader-study interface, showing a representative multimodal item with histopathology images, multiple-choice options, and sliders for answer confidence and PathPocket content quality.
\textbf{B.} Composition of the 2,000 scored response records by anatomy system, modality (text-only vs.\ multimodal), and question type (open-ended vs.\ multiple choice); text-only counts reflect two sub-questions per parent item.
\textbf{C.} Accuracy (top row) and confidence (bottom row) comparing w/o versus w/ PathPocket, overall and stratified by modality and question type. PathPocket is associated with higher accuracy across all subsets (overall 80.4\% → 85.8\%; p = 0.02) and significantly higher confidence overall (7.24 → 8.08; p = 0.03).
\textbf{D.} Paired outcomes for each pathologist (P1–P8): accuracy (top) and confidence (bottom) with PathPocket (y-axis) versus without (x-axis). Points above the diagonal indicate superiority; junior (purple) and senior (orange) pathologists are distinguished.
\textbf{E.} Accuracy (top) and confidence (bottom) by anatomy system. Percentage-point accuracy gains with PathPocket are annotated at right and confidence increases are shown as absolute score differences.
\textbf{F.} Inter-rater consistency and experience-related performance gap. Open-ended average-measure ICC(2,$k$) increased from 0.549 to 0.719; pairwise tolerance agreement ($\leq$1 rubric step) rose from 76.4\% to 86.7\%; multiple-choice Fleiss’ $\kappa$ increased from 0.47 to 0.69. Score dispersion (within-question and between-pathologist standard deviation) decreased with PathPocket. The absolute accuracy gap between senior and junior pathologists narrowed from 4.3\% to 2.9\%.
\textbf{G.} PathPocket content quality ratings (1–10) reported by pathologists: overall distribution (left), mean rating by pathologist (middle), and mean rating by task type (right; mean 8.94 across conditions).}
    \label{fig:user_study_design}
\end{figure}

\subsection{Crossover reader study demonstrates the clinical utility of PathPocket for pathologists}

While automated benchmarks establish the foundational capabilities of an AI model, the ultimate test of a clinical co-pilot is its tangible impact on human decision-making. To rigorously assess whether PathPocket's evidence-based reasoning translates into measurable clinical benefits, we conduct a randomized, crossover reader study (Figure~\ref{fig:user_study_design}). This study simulates real-world diagnostic workflows using a curated item bank to evaluate human-AI collaboration.

To ensure our findings are generalizable across different experience levels, we recruit 8 board-certified pathologists (see Appendix Table \ref{tab:RS_pathologist} for details), equally stratified into two tiers of expertise: 4 senior pathologists (P5--P8) with > 5 years experience and 4 junior pathologists (P1--P4) with < 5 years. We adopt a crossover design so that each pathologist acts as their own control, neutralizing confounding variables such as individual baseline diagnostic skills. The total item bank comprises 150 unique questions, including 100 text-only items and 50 multimodal items containing histopathology images (Figure~\ref{fig:user_study_design}B; Appendix Table \ref{tab:RS_questions}). Each text-only question is structured as two open-ended sub-questions that are scored independently; we therefore expand these 100 parent items into 200 scoring units (together with 50 multimodal items, 250 scored items in total). To prevent memory bias while maintaining balanced test conditions, independent pathologists partition these questions into two parallel, balanced sets, matched precisely by anatomical system, question type (open-ended vs.\ multiple-choice), and difficulty.

\begin{figure}[H]
    \centering
    \includegraphics[width=0.92\textwidth]{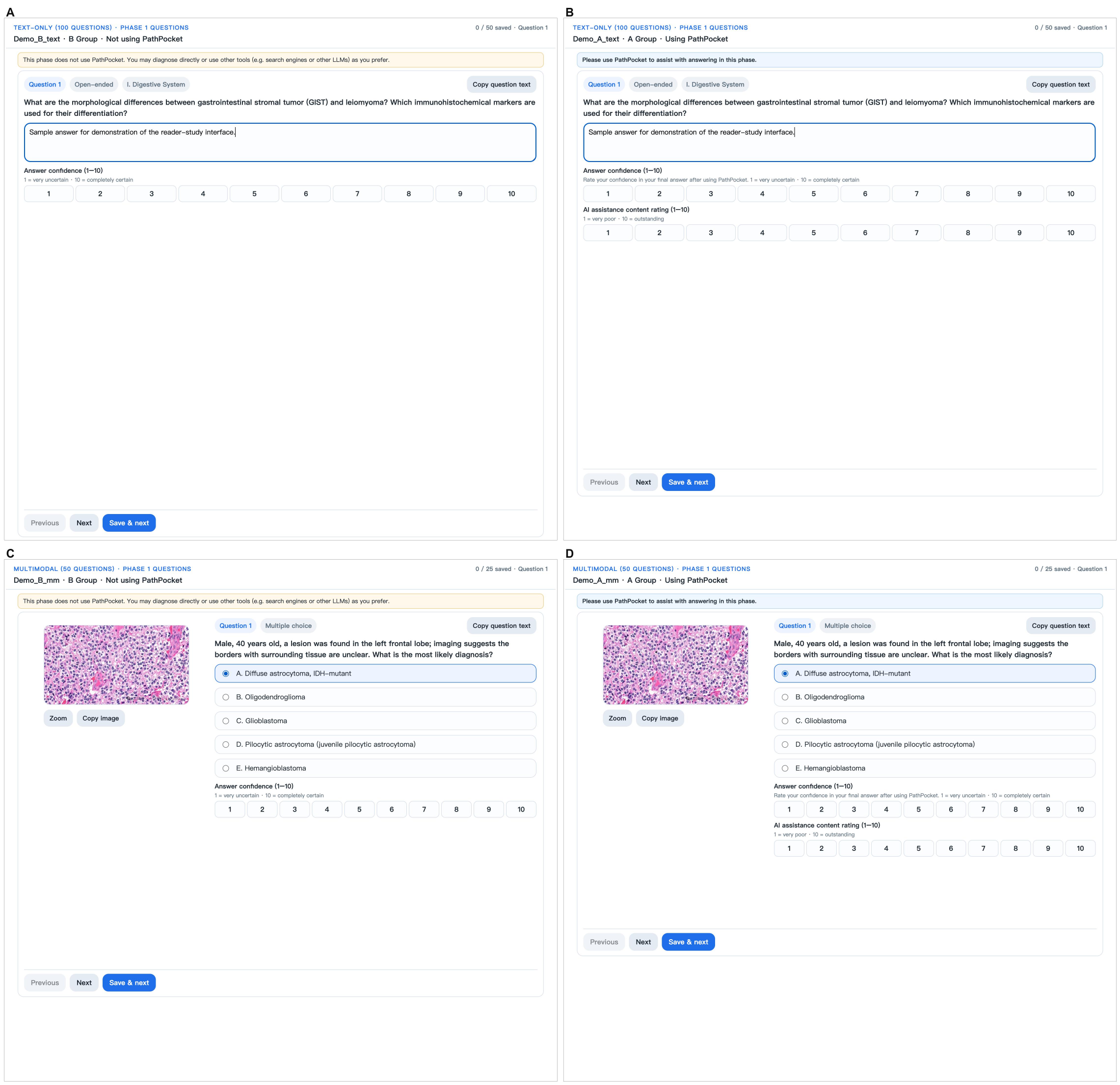}
    \caption{\textbf{Web answering interfaces of the PathPocket reader-study platform.}
    Rows contrast question banks; columns contrast assistance conditions.
    \textbf{(A)}~Text-only \textbf{without PathPocket}; \textbf{(B)}~Text-only \textbf{with PathPocket} (adds content-quality rating, 1--10).
    \textbf{(C)}~Multimodal \textbf{without PathPocket}; \textbf{(D)}~Multimodal \textbf{with PathPocket}.
    Under the without-PathPocket condition, pathologists may diagnose independently or use other search tools.
    Registration (name, Group~A/B, bank, and phase) is completed on a separate landing page before answering.}
    \label{fig:RS_interface}
\end{figure}

Participants are randomized into Group A and Group B (Appendix Figure~\ref{fig:RS_consort}) and complete the study on a web platform that enforces the assigned assistance condition (Appendix Figure~\ref{fig:RS_interface}). Within each stage, participants in both groups evaluate the exact same set of clinical cases, ensuring strict control over case difficulty within each testing phase. In Stage 1, Group A utilizes the PathPocket agentic co-pilot alongside their workflow (Diagnosis w/ PathPocket), while Group B establishes the human baseline by diagnosing the cases independently or using standard search tools (Diagnosis w/o PathPocket). Following a 4-week washout period, the conditions cross over in Stage 2: Group B receives PathPocket assistance, and Group A returns to unassisted diagnosis. Across both stages, this design yields 2,000 pathologist-level response records (250 scored items $\times$ 8 pathologists), which are evaluated on diagnostic \textit{Accuracy} (via GPT-5 mini scoring for open-ended questions; Appendix~\ref{sec:appendix_reader_scoring_prompts}), self-reported diagnostic \textit{Confidence} (1--10 scale), inter-observer \textit{Consistency}, and subjective \textit{Content Quality} (collected exclusively during PathPocket-assisted sessions).

The reader study demonstrates that PathPocket can enhance diagnostic performance across four key clinical and user-experience dimensions.

First, PathPocket assistance drives significant objective \textbf{Accuracy Gains}. The overall diagnostic accuracy rises from 80.4\% to 85.8\% ($p < 0.05$, Figure~\ref{fig:user_study_design}C; Appendix Table \ref{tab:RS_primary}). Stratified analysis reveals robust improvements across both text-only items (83.8\% $\rightarrow$ 89.1\%) and challenging multimodal items (66.8\% $\rightarrow$ 72.4\%, $p < 0.05$). This accuracy enhancement is highly consistent, appearing across almost all individual pathologists (Figure~\ref{fig:user_study_design}D; Appendix Table \ref{tab:RS_individual}) and anatomical systems (Figure~\ref{fig:user_study_design}E; Appendix Table \ref{tab:RS_organ}), with the most pronounced percentage-point improvements occurring in Cardiovascular (+12.6\%) and Head and Neck (+10.7\%) cases.

Second, PathPocket significantly bolsters practitioners' subjective \textbf{Diagnostic Confidence}. The overall self-reported confidence score increases from 7.24 to 8.08 ($p < 0.05$, Figure~\ref{fig:user_study_design}C; Appendix Table \ref{tab:RS_secondary}). This positive shift in certainty is observable in both text-only and multimodal tasks, indicating that the evidence-based reasoning provided by the AI effectively alleviates diagnostic hesitation in diverse clinical scenarios.

Third, human-AI collaboration successfully \textbf{Bridges the Experience Gap and Enhances Consistency} among readers. PathPocket acts as an expert equalizer, narrowing the absolute diagnostic accuracy margin between senior and junior pathologists from 4.3\% down to 2.9\% (Figure~\ref{fig:user_study_design}F; Appendix Table \ref{tab:RS_interrater} and \ref{tab:RS_experience_gap}). Concurrently, it reduces diagnostic dispersion and increases consensus. Inter-rater alignment improves substantially: the average-measure intraclass correlation for open-ended questions, $\text{ICC}(2,k)$, increases from 0.549 to 0.719; pairwise tolerance agreement within one rubric step rises from 76.4\% to 86.7\%; and Fleiss' $\kappa$ for multiple-choice questions rises from 0.47 to 0.69 (Figure~\ref{fig:user_study_design}F). Both within-question and between-pathologist standard deviations decrease under the assisted condition.

Fourth, the system achieves exceptional user satisfaction in \textbf{Content Quality Evaluation}. During the assisted phases, pathologists award PathPocket an outstanding average content quality rating of 8.94 out of 10 (Figure~\ref{fig:user_study_design}G; Appendix Table \ref{tab:RS_content_rating}). The high ratings remain stable across different individual pathologists and across both text-only and multimodal task types (Figure~\ref{fig:user_study_design}G), confirming that the retrieved multimodal evidence is highly valued by clinical practitioners.

\section{Discussion}

\noindent\textbf{PathPocket supports evidence grounding beyond purely parametric prediction.}
In this study, we introduce PathPocket, a multimodal agentic co-pilot for evidence grounded computational pathology. Clinical use of large language models and multimodal large language models is often constrained by limited interpretability, hallucination risk, and the lack of citable justification. PathPocket couples graded medical literature with collaborative multi-agent reasoning so that answers can be retrieved, filtered, and synthesized from explicit evidence rather than relying solely on parametric memory. Across text-only, ROI-level, and WSI-level settings, this design is associated with improved automated performance and measurable gains in pathologist-AI collaboration.

\noindent\textbf{A graded multimodal evidence hypergraph provides a structured and auditable knowledge base.}
From a data perspective, PathPocket is built on a stratified pathology evidence corpus. Medical LLMs are commonly trained on heterogeneous internet-scale text with uneven evidence quality, which can lead to weak traceability. Our corpus comprises over 110,000 public and authorized documents spanning guidelines, systematic reviews, randomized trials, observational studies, case reports, consensus statements, textbooks, and expert opinions, each tagged with an eight-tier adaptation of the GRADE hierarchy. From this foundation, we construct a multimodal pathology hypergraph containing 4.55 million entities and 7.10 million relations, including images and tables alongside clinical concepts. Hyperedges allow a single relation to connect multiple entities, which helps represent multivariate diagnostic dependencies that flat chunk retrieval and pairwise graphs capture less directly. New literature can also be added to the hypergraph without retraining a monolithic model, which facilitates continual updating.

\noindent\textbf{Collaborative multi-agent reasoning produces more transparent diagnostic trajectories.}
Methodologically, PathPocket separates pathology reasoning into specialized agents for case understanding, evidence retrieval, evidence filtering, and response generation. This organization mirrors common steps in diagnostic workup and makes intermediate decisions more explicit than single-pass generation. Supporting infrastructure for document parsing, dense embeddings, gigapixel patch selection, and reranking enables the same framework to handle clinical text, multi-image ROI panels, and WSIs. Final responses include direct citations so that clinicians can inspect the evidentiary basis of each answer. Qualitative analyses further suggest that evidence retrieval can reduce certain shared failure modes observed among strong parametric baselines on ambiguous cases.

\noindent\textbf{Multimodal benchmarking shows improved performance across text, ROI, and WSI tasks.}
We evaluate PathPocket on a benchmark of more than 21,000 cases spanning 33 clinical tasks, including pathologist-generated USMLE-style and NMLE-style items, report-derived SMU-Text questions, public and private ROI cohorts, and nine private WSI endpoints. The WSI benchmark includes both retrospective slides and prospective clinical subsets (25.4\% of the WSI cohort), enabling assessment on newly accrued cases such as CRC biopsy diagnosis and gastric CAG assessment. Across these settings, PathPocket achieves higher accuracy than the compared general-purpose and pathology-specific baselines. 

\noindent\textbf{Reader-study results indicate benefit for pathologist-AI collaboration.}
Beyond automated benchmarks, a randomized two-stage crossover reader study shows that PathPocket assistance is associated with improved pathologist performance under controlled conditions. Overall diagnostic accuracy increased from 80.4\% to 85.8\%, self-reported confidence from 7.24 to 8.08, the senior-junior accuracy gap narrowed from 4.3\% to 2.9\%, and inter-rater agreement improved for both open-ended and multiple-choice items, with a mean content-quality rating of 8.94/10. These results support an assistive role for PathPocket as an evidence-providing co-pilot, rather than a replacement for pathologist judgment.

\noindent\textbf{Limitations and future work.}
Three limitations are particularly relevant. First, although PathPocket outperforms baselines on WSI-level tasks, absolute accuracies remain lower than those for many text-only and ROI-level endpoints. Gigapixel reasoning depends on upstream tools such as multi-scale patch selection; errors or incomplete coverage in these modules can omit diagnostically important regions or introduce irrelevant context before agent reasoning. Future work should therefore evaluate more accurate patch-selection and multi-scale aggregation tools. Second, the multi-agent architecture incurs higher latency than single-turn LLMs because of multi-hop retrieval and agent coordination. This overhead is generally compatible with routine pathology turnaround measured in hours or days, but additional systems optimization will be needed for near-real-time settings such as frozen-section consultation. Third, evidence grounding does not eliminate residual errors on intrinsically hard near-neighbor differentials that remain challenging even for expert pathologists (see failure cases in Appendix Figure~\ref{fig:failure_cases}).

In summary, PathPocket shows that combining a graded multimodal hypergraph with multi-agent collaboration can improve both automated pathology reasoning and pathologist-AI collaboration. Validation across automated benchmarks, including prospective WSI evaluation, and a crossover reader study supports its potential as an auditable, literature-backed decision-support approach for computational pathology.

\section{Method}

The methodology for developing and comprehensively validating the PathPocket framework is structured into six sequential components, moving from foundational data curation to system architecture and rigorous clinical evaluation. 

First, we detail the Collection and Stratification of Pathology Evidence, establishing a medically grounded knowledge repository wherein diverse literature is graded by its authoritative consensus level. To process this large-scale influx of multimodal data, we outline the Infrastructure and Tooling for Multi-Agent Operations, introducing the specialized vision-language foundation models, parsing engines, and database systems that serve as the ``hands and eyes'' of our framework. Leveraging these tools, we describe the Multimodal Pathology Hypergraph Construction via Multi-Agent Collaboration, detailing how autonomous agents systematically extract, verify, and topologically link complex clinical entities into a high-dimensional graph. 

For the inference phase, we present the Multi-Agent Pathology Reasoning pipeline, illustrating how specialized agents collaborate for case understanding, evidence retrieval, evidence filtering, and response generation. To rigorously assess the diagnostic capabilities of this framework, we introduce a Comprehensive Pathology Evaluation Benchmark, which encompasses over 21,000 real-world evaluation items scaling from text-only reasoning to gigapixel WSI diagnosis across multiple private hospital cohorts. Finally, we provide the Implementation Details, specifying the hardware configurations, model deployment strategies, and computational frameworks utilized throughout our experiments.

\subsection{Collection and Stratification of Pathology Evidence}
\label{sec:data_collection}

To establish a highly authoritative, hallucination-free knowledge engine for PathPocket, we curate the most comprehensive corpus of multimodal pathology evidence to date. The dataset comprises over 110,000 public and authorized medical documents ($N = 110,472$). To operationalize evidence-based medicine (EBM) within our AI framework, we systematically stratified all collected documents into an 8-tier evidence hierarchy, adapted from standard evidence-based medicine framework GRADE\cite{guyatt2008grade,howick2011oxford}. 

The collection spans a diverse array of highly credible medical sources across the 8 evidence tiers:
\begin{itemize}
    \item \textbf{Level 1: Guidelines ($n=981$).} We aggregated gold-standard clinical and pathological guidelines from premier international and top-tier regional organizations, including the World Health Organization (WHO), College of American Pathologists (CAP), European Society for Medical Oncology (ESMO), Chinese Anti-Cancer Association (CACA), Chinese Society of Clinical Oncology (CSCO), and the National Health Commission (NHC).
    \item \textbf{Level 2: Systematic Reviews \& Meta-Analyses ($n=12,354$).} High-quality synthesis literature was systematically retrieved from PubMed Central (PMC).
    \item \textbf{Level 3: Randomized Controlled Trials ($n=9,305$).} Peer-reviewed clinical trials with rigorous controls were sourced from PMC.
    \item \textbf{Level 4: Observational \& Cohort Studies ($n=14,751$).} Including large-scale case-control and cohort studies, primarily extracted from PMC.
    \item \textbf{Level 5: Case Reports ($n=71,176$).} The largest segment of our corpus, providing vast long-tail and rare-disease morphological examples, sourced from PMC.
    \item \textbf{Level 6: Consensus Statements ($n=279$).} Expert consensus reports retrieved from PMC.
    \item \textbf{Level 7: Textbooks ($n=263$).} Classic and modern medical textbooks that provide foundational diagnostic criteria and dense multimodal knowledge.
    \item \textbf{Level 8: Expert Opinions ($n=1,396$).} Editorials and expert viewpoints sourced from PMC.
\end{itemize}

All documents are acquired through authorized institutional access or open-access repositories. To facilitate reproducibility and transparency, a comprehensive list of the source databases, specific textbook titles, and exact URLs accessed during the dataset curation is provided in \textbf{Appendix Table S1}.

\subsection{Infrastructure and Tooling for Multi-Agent Operations}
\label{sec:method-tools}

To empower the multi-agent workflow during both the hypergraph construction phase and the dynamic reasoning phase, PathPocket integrates a suite of state-of-the-art infrastructure and foundational models. These specialized tools function as the ``hands and eyes'' of the LLM agents, enabling them to process complex multimodal data, execute efficient retrievals, and handle gigapixel images.

\paragraph{High-Accuracy Document Parsing.} 
During the initial stage of hypergraph construction, the Parsing Agent utilizes \textbf{MinerU}\cite{wang2024mineru}, an accurate layout-aware document parsing engine. Medical documents are notoriously complex, containing nested tables, multi-column layouts, and interwoven figures. MinerU structurally decomposes these PDFs, precisely isolating text blocks, hierarchical headers, and multimodal elements (images/tables) while preserving their reading order and structural semantics, providing a clean data stream for subsequent LLM extraction.

\paragraph{Hypergraph Storage and Topological Querying.} 
The hypergraph is stored and queried using \textbf{PostgreSQL}, augmented with the \texttt{pgvector} extension. Rather than relying on specialized but rigid graph databases, we leverage PostgreSQL's robust array operations to model hyperedges. For instance, the Evidence Retrieval Agent dynamically generates SQL commands utilizing array overlap operators (e.g., \texttt{WHERE entities \&\& \$2::text[]}) to rapidly fetch hyperedges that intersect with the queried clinical entities, ordering the results by the degree of entity overlap (Appendix~\ref{sec:appendix_hypergraph_sql}). This ensures mathematically rigorous topological traversal alongside semantic vector search.

\paragraph{WSI Patch Selection.} 
WSIs are gigapixel-scale and cannot be directly ingested by standard MLLMs. During inference, PathPocket converts each slide into a compact set of multi-scale representative ROIs using a TITAN-guided selection pipeline (Appendix~\ref{sec:appendix_wsi_patch}).
Briefly, tissue patches are first encoded with CONCH~v1.5\cite{conch}; the TITAN\cite{ding2025multimodal} vision encoder then assigns attention-based importance scores to all patches.
On the resulting attention map, we select the highest-scoring square window at each of four nested scales ($1\times$, $2\times$, $4\times$, and $8\times$ the base patch size), proceeding from coarse to fine while enforcing IoU and spatial-margin constraints across scales.
The selected level-0 crops are resized and passed to the multimodal agents as the visual evidence for WSI reasoning.

\paragraph{Multimodal Embedding Foundations} 
To map all textual and visual data into high-dimensional vector space for cross-modal semantic retrieval, our system relies on two specialized foundation models. 
\textbf{Text Embedding (\textbf{BGE-m3}\cite{bge-m3}):} Utilized by the Consolidation Agent to generate dense vector representations for raw text chunks, entity attributes, and hyperedge descriptions. Its multilingual capabilities ensure robust alignment across diverse medical nomenclatures. \textbf{Vision Embedding (\textbf{Virchow2}\cite{virchow2}):} A pathology-specific vision foundation model utilized to encode all visual modalities. During hypergraph construction, it embeds parsed textbook figures; during inference, it embeds the patient's WSI ROIs selected by TITAN. This enables the Evidence Retrieval Agent to perform highly accurate image-to-image and text-to-image semantic matching.

\paragraph{Semantic Context Filtering.} 
Given the large-scale of the evidence retrieved from the hypergraph, the Evidence Filtering Agent employs \textbf{Qwen3-Reranker}\cite{qwen3embedding} as its primary scoring engine. It analyzes the semantic interplay between the structured query intent and the retrieved evidence chunks. By providing highly calibrated relevance scores, Qwen3-Reranker allows the Evidence Filtering Agent to confidently discard noise and truncate the context window, ensuring that the Response Generation Agent is fed only the most pertinent and high-quality evidence.

\subsection{Multimodal Pathology Hypergraph Construction via Multi-Agent Collaboration}
\label{sec:hypergraph_pipeline}

We conceptualize the construction of the multimodal pathology hypergraph as a collaborative multi-agent workflow. Over pathology documents, four specialized agents operate sequentially: (i) a Parsing and Chunking Agent for data structuralization, (ii) an Extraction Agent for initial entity and $n$-ary relation generation, (iii) a Rectifying Agent for refinement, and (iv) a Consolidation and Indexing Agent for topological assembly and multimodal dense embedding.

Let $\mathcal{D}$ denote the set of source documents. Each document is split into a set of text chunks $\mathcal{C}^{\text{txt}}=\{c_1,\dots,c_{N}\}$ and multimodal chunks $\mathcal{C}^{\text{mm}}=\{u_1,\dots,u_M\}$ (images/tables). A chunk $x$ carries content $s_x$, provenance, and identifier $\mathrm{id}(x)$. The final pathology hypergraph is $\mathcal{H}=(\mathcal{V},\mathcal{E})$, where vertices $\mathcal{V}$ are typed entities (including image/table nodes) and each hyperedge $e\in\mathcal{E}$ is an unordered set of incident entities $\mathrm{inc}(e)\subseteq\mathcal{V}$ with attributes (keywords, natural-language description, source chunk ids).

\subsubsection*{Parsing and Chunking Agent}
\label{sec:stage1}

To accurately extract knowledge from complex medical PDFs, the \textbf{Parsing and Chunking Agent} utilizes MinerU\cite{wang2024mineru} for layout-aware document parsing. MinerU structurally decomposes each document, isolating text paragraphs, hierarchical titles, figures, and tables into a structured content stream. The agent separates the textual stream $S$ from multimodal items $\mathcal{U}$. 
Text is segmented by a tokenizer-based windowing function under a token budget $T_{\max}$:
\begin{equation}
\mathcal{C}^{\text{txt}} = \mathrm{Chunk}(S;\, T_{\max}).
\end{equation}
For each multimodal unit $u\in\mathcal{U}$, the parsed textual description and metadata are stored together as a multimodal chunk record. Full document text and chunk inventories are persisted for downstream extraction.

\subsubsection*{Extraction Agent}
\label{sec:stage2}
For every $c\in\mathcal{C}^{\text{txt}}$ and multimodal-backed chunk in $\mathcal{C}^{\text{mm}}$, the \textbf{Extraction Agent} prompts an LLM with a pathology-specialized instruction schema. The agent outputs a linearized list of records using delimiters $\langle\!\langle \# \rangle\!\rangle$ between tuple fields and a completion token $\langle\!\langle\mathrm{COMPLETE}\rangle\!\rangle$.
Each entity line has the form
\begin{equation}
\texttt{entity} \,\|\, \texttt{name} \,\|\, \texttt{type} \,\|\, \texttt{description},
\end{equation}
and each relation line unifies binary and hyperedges as
\begin{equation}
\texttt{relation} \,\|\, v_1 \,\|\, \cdots \,\|\, v_k \,\|\, \texttt{keywords} \,\|\, \texttt{description},
\quad k\ge 2,
\end{equation}
where $\{v_1,\dots,v_k\}$ are entity names that must align with previously emitted names. 
Entity types are strictly constrained to a comprehensive domain ontology $\mathcal{T}$, comprising exactly 24 classes:
\texttt{Disease}, \texttt{Symptom}, \texttt{PathologicalFinding}, \texttt{AnatomicalSite}, \texttt{CellType}, \texttt{HistologicalPattern}, \texttt{Gene}, \texttt{GeneticMutation}, \texttt{Biomarker}, \texttt{MolecularPathway}, \texttt{Pathogen}, \texttt{RiskFactor}, \texttt{Pathogenesis}, \texttt{DiagnosticMethod}, \texttt{StainingMethod}, \texttt{LabTest}, \texttt{DiagnosticCriteria}, \texttt{Treatment}, \texttt{Drug}, \texttt{MedicalDevice}, \texttt{Prognosis}, \texttt{Organization}, \texttt{Location}, and \texttt{Person}.

\subsubsection*{Rectifying Agent}
\label{sec:stage3}
To maximize extraction recall and ensure structural validity, a \textbf{Rectifying Agent} performs a secondary auditing pass. Conditioned on the initial extraction completion, the Rectifying Agent is instructed to emit only missed entities or repair malformed relation records.
Formally, let the initial extraction be $y^{(0)}=\mathcal{A}_{\mathrm{ext}}(s_x)$. With a predefined rectifying budget $G\in\mathbb{Z}_{\ge 0}$, the agent operates recursively:
\begin{equation}
y^{(g+1)} = \mathcal{A}_{\mathrm{rect}}\!\left(s_x,\, y^{(g)}\right),\quad g<G.
\end{equation}

\subsubsection*{Consolidation and Indexing Agent}
\label{sec:stage4}
The final \textbf{Consolidation and Indexing Agent} is responsible for merging the extracted components into a cohesive topological structure and generating dense embeddings for all elements to enable semantic retrieval.

\paragraph{Hypergraph Consolidation.} 
Entities are merged by canonical name with type normalization rules. Dedicated nodes $a(u)\in\mathcal{V}$ are created for multimodal units $u$. For each multimodal chunk, relations extracted from that chunk are filtered, and $a(u)$ is inserted into each surviving incident set to assemble multimodal-text hyperedges, ensuring $|{\mathrm{inc}(e)}|\ge 2$. Redundant binary edges nested within higher-arity hyperedges are pruned. The agent then upserts nodes and hyperedges into graph storage, merging provenance fields such as source chunk ids and concatenated descriptions.

\paragraph{Multimodal Dense Indexing.} 
To complete the knowledge engine, the agent computes and writes embeddings into a vector database (using pgvector HNSW indexes) across four granularities:
\begin{itemize}
    \item \textbf{Text Chunks:} Dense vectors for raw chunk contents $s_c$ are computed using a multilingual text encoder $\phi$ (e.g., BGE-M3\cite{bge-m3}).
    \item \textbf{Entities / Nodes:} For each entity $v$ with name $n_v$ and description $d_v$, the embedding is generated by concatenating its attributes: $\mathbf{e}_v = \phi\!\left(\texttt{concat}(n_v,d_v)\right)$.
    \item \textbf{Relations / Hyperedges:} For each hyperedge $e$ with keyword list $\kappa_e$, entity multiset $V_e$, and description $d_e$, the representation is synthesized as: $\mathbf{e}_e = \phi\!\left(\texttt{format}(\kappa_e, V_e, d_e)\right)$.
    \item \textbf{Images:} For nodes associated with visual modalities (i.e., parsed ROIs from document), visual embeddings are generated using a pathology-specific vision foundation model $\psi_{\mathrm{vis}}$ as: $\mathbf{e}_{\mathrm{img}} = \psi_{\mathrm{vis}}(u_{\mathrm{image}})$.
\end{itemize}

\subsection{Multi-Agent Pathology Reasoning}
\label{sec:method-inference}

Given a complex pathology query $q$, which typically comprises unstructured clinical narratives, diagnostic candidate options, and potentially multimodal visual inputs such as WSIs, PathPocket orchestrates a collaborative workflow of four specialized agents over the constructed multimodal pathology hypergraph $\mathcal{G}$.

\subsubsection*{Overview of the Agentic Workflow}
\label{subsec:overview}

The end-to-end diagnostic pipeline is formulated as:
\begin{equation}
  \hat{y} \;=\; \mathcal{A}_{\mathrm{rg}}\!\Big(
    \mathcal{A}_{\mathrm{ef}}\!\big(
      \mathcal{A}_{\mathrm{er}}\!\big(
        \mathcal{A}_{\mathrm{cu}}(q)
      \big)
    \big),\; q
  \Big),
\end{equation}
where $\mathcal{A}_{\mathrm{cu}}$ is the \textbf{Case Understanding Agent}, $\mathcal{A}_{\mathrm{er}}$ the \textbf{Evidence Retrieval Agent}, $\mathcal{A}_{\mathrm{ef}}$ the \textbf{Evidence Filtering Agent}, and $\mathcal{A}_{\mathrm{rg}}$ the \textbf{Response Generation Agent}. 

\subsubsection*{Case Understanding Agent}
\label{subsec:parse}

Real-world pathology questions are often convoluted and noisy. To prevent context pollution, the Case Understanding Agent $\mathcal{A}_{\mathrm{cu}}$ structurally decomposes the raw text $q$ into a highly specific JSON retrieval object $\sigma$:
\begin{equation}
  \sigma = \mathcal{A}_{\mathrm{cu}}(q)
  = \bigl(
    s,\;
    \mathbf{g},\; \mathbf{g}^{\mathrm{desc}},\;
    \mathbf{m},\; \mathbf{m}^{\mathrm{desc}},\;
    \mathbf{k},\; \mathbf{k}^{\mathrm{desc}},\;
    \mathbf{c},\; \mathbf{c}^{\mathrm{desc}},\;
    \mathbf{o},\;
    \mathbf{a}
  \bigr).
\end{equation}
The fields comprehensively capture the clinical picture: anatomical \textbf{site} $s$; lists of macroscopic (gross) entities $\mathbf{g}$, microscopic morphology $\mathbf{m}$, molecular/IHC markers $\mathbf{k}$, and clinical context $\mathbf{c}$, alongside their corresponding natural language descriptions ($\mathbf{g}^{\mathrm{desc}}, \mathbf{m}^{\mathrm{desc}}$, etc.); and the candidate diagnostic options $\mathbf{a}=\{a_1,\ldots,a_K\}$. 

When the input query is accompanied by a gigapixel WSI, processing the entire image is computationally prohibitive. In such cases, the agent invokes an integrated visual selection tool. Utilizing the attention maps extracted from a pre-trained pathology vision foundation model, the agent autonomously localizes and extracts representative ROIs that exhibit the highest diagnostic salience. These localized ROIs, alongside the structured text JSON $\sigma$, form the complete multimodal query intent.

\subsubsection*{Evidence Retrieval Agent}
\label{subsec:retrieval}

The Evidence Retrieval Agent $\mathcal{A}_{\mathrm{er}}$ executes a sophisticated hybrid search across the hypergraph $\mathcal{G}$ and semantic vector space $\mathcal{C}$, leveraging the unique high-dimensional topology of our knowledge base. Let $V$, $R$, and $D$ denote retrieved entities, hyperedge relations, and text chunks, respectively.

Using the parsed entities, the agent generates dynamic PostgreSQL instructions to traverse the hypergraph topologically (Appendix~\ref{sec:appendix_hypergraph_sql}). For any retrieved hyperedge $r$, its relevance to the query core set $C$ is measured via an Intersection over Union (IoU) of incident entities. Simultaneously, taking advantage of the hypergraph's unique structure where edges themselves encapsulate complex clinical narratives, the agent performs dense semantic retrieval directly over the hyperedge embeddings ($\mathrm{Enc}(R)$). Graph hits and vector hits are then intersected based on their entity sets to maximize precision.

Crucially, when the input query contains visual modalities (raw or selected ROIs), the agent generates visual embeddings via the foundation model and performs semantic search within the vector database. It retrieves highly similar multimodal nodes along with their meticulously paired textual descriptions (captions), ensuring that morphological patterns observed in the patient are matched with verified visual evidence from the literature.

\subsubsection*{Evidence Filtering Agent}
\label{subsec:filter}

The raw evidence bundle $\mathcal{E}_0 = (V_0, R_0, D_0)$ retrieved from the large-scale database inevitably contains noise. The Evidence Filtering Agent $\mathcal{A}_{\mathrm{ef}}$ acts as a strict gatekeeper, refining $\mathcal{E}_0$ into a high-fidelity, token-budgeted context $\mathcal{E}$.

Instead of relying on traditional cross-encoders, we deploy a reranking model\cite{qwen3embedding} to evaluate the relevance of each retrieved relation and chunk. By prompting a secondary LLM with the structured query intent $\sigma$ and candidate evidence, the model scores and filters the evidence based on deep clinical logic and pathological semantics. The exact prompt template utilized for this LLM-based reranking is provided in Appendix~\ref{sec:appendix_prompts}. Evidence scoring below a predefined confidence threshold or misaligned with the parsed anatomical site is discarded.

\subsubsection*{Response Generation Agent}
\label{subsec:response}

Finally, the Response Generation Agent $\mathcal{A}_{\mathrm{rg}}$ synthesizes the clinical narrative and formulates the final answer.

Within the generation prompt $\mathcal{P}_{\mathrm{rg}}$, the filtered context $\mathcal{E}$ is meticulously formatted. A critical feature of this formatting is that every piece of evidence is explicitly tagged with its authoritative source and evidence level. The Response Generation Agent is explicitly instructed to weigh these tiers during its reasoning process, prioritizing high-tier consensus over lower-tier observational reports when resolving conflicting information. The agent then formulates the final response strictly from this graded context, explicitly acknowledges insufficient information if required to prevent hallucination, and appends structured reference citations for total transparency.

\subsection{Comprehensive Pathology Evaluation Benchmark}
\label{sec:benchmark}

To rigorously validate the diagnostic capabilities and clinical utility of PathPocket, we construct a comprehensive pathology evaluation benchmark comprising over 21,000 rigorous evaluation items. The multimodal pathology hypergraph that supplies PathPocket's evidence is built from public and authorized documents. To reduce the risk of data leakage between this evidence source and the evaluation queries, most experiments are therefore conducted on private, real-world clinical cohorts sourced from hospital workflows, with a smaller set of public ROI datasets retained for external reference. The benchmark is structurally divided into three categories of escalating clinical complexity: text-only reasoning, ROI-level multimodal interpretation, and WSI-level diagnosis.

\subsubsection*{Text-Only Tasks}
This foundational tier evaluates the system's medical reasoning and natural language understanding capabilities in the absence of visual context. 

\textbf{USMLE 1-3.} 
The USMLE subset comprises 133 pathologist-generated English-language clinical vignettes in the style of the United States Medical Licensing Examination. Designed to isolate general clinical reasoning under standard prompt distributions, these multiple-choice items span mixed internal medicine systems without being restricted to a single organ. 

\textbf{NMLE 1-3.} 
The NMLE subset consists of 142 pathologist-generated Chinese-language cases in the style of the National Medical Licensing Examination of China from the nominal years 2022 to 2024. It provides general clinical coverage and evaluates foundational medical knowledge using standard single-best-answer multiple-choice prompts.

\textbf{SMU-Text 1-6.} 
To push beyond the generic scope of these examination-style items, the SMU-Text dataset provides a large-scale, private retrospective cohort of 1,854 hospital-derived cases curated at Southern Medical University (SMU). Packaged across six disjoint subsets, it heavily stresses complex, pathology-centric knowledge. The items focus on tumor biology, diagnostic pathology, and detailed clinical contexts, evaluating the models on real-world unstructured clinical reasoning via explicit multiple-choice formulations.

\subsubsection*{ROI-Level Multimodal Tasks}
This intermediate tier requires the system to jointly synthesize clinical text and localized ROI microscopic images. It includes established public benchmarks alongside massive private real-world cohorts. 

\textbf{BreakHis.} 
The BreakHis dataset \cite{spanhol2015dataset} for breast cancer histopathological image classification is utilized in this study for external validation. The dataset comprises two primary classes: benign tumors and malignant tumors. All ROIs are captured at four distinct magnification levels (40$\times$, 100$\times$, 200$\times$, and 400$\times$). For consistency, images are resized to 224$\times$224 pixels. For our benchmark evaluation, to ensure the assessment of true generalization on unseen data, we exclusively employ the official test set. This split consists of 1,582 ROIs (20\% of the total dataset), stratified to preserve the original label distribution. Detailed experimental results are provided in Appendix Table \ref{tab:bench:roi_public}.

\textbf{CCRCC.} 
The CCRCC dataset \cite{brummer2023computational} comprises 52,723 annotated histopathology ROIs (300$\times$300 pixels) derived from WSIs of clear cell renal cell carcinoma (CCRCC) specimens. These ROIs are randomly sampled from two independent sources: the TCGA-KIRC repository and the Helsinki cohort. The dataset encompasses six distinct histological classes: malignant tumor regions (13,057 ROIs), normal renal parenchyma (8,652 ROIs), stromal tissue (5,460 ROIs), red blood cell accumulations (996 ROIs), non-informative background areas (16,026 ROIs), and heterogeneous tissue types including necrosis and artifacts (8,522 ROIs). For robust classification modeling, we focused exclusively on four biologically meaningful classes (cancer, normal tissue, stroma, and blood), excluding ambiguous and non-informative ROIs. The dataset was formally partitioned into training (22,530 ROIs) and test sets while preserving class distributions. In our evaluation pipeline, the models are evaluated strictly on the 5,635 ROIs comprising the official unseen test set. Comprehensive performance metrics are provided in Appendix Table \ref{tab:bench:roi_public}.

\textbf{Chaoyang.} 
The Chaoyang dataset \cite{zhu2021hard} provides a comprehensive collection of histopathology ROIs for colorectal tissue analysis, comprising four clinically relevant classes: normal mucosa (1,816 ROIs), serrated lesions (1,163 ROIs), adenocarcinoma (2,244 ROIs), and adenoma (937 ROIs). All ROIs are standardized to 224$\times$224 pixels. Following the official dataset splits to ensure strict reproducibility and valid comparative benchmarking, our evaluation exclusively utilizes the designated unseen test set consisting of 2,139 ROIs. Performance metrics are detailed in Appendix Table \ref{tab:bench:roi_public}.

\textbf{SMU-VL 1-6.} 
A critical limitation of the aforementioned public datasets is their simplified paradigm, where each evaluation item provides only a single, highly curated image. To bridge the gap to actual clinical practice, we introduce the SMU-VL cohort, comprising 4,937 multimodal VQA instances curated at SMU. Representing a complex \emph{multi-image} paradigm, each clinical scenario requires the model to simultaneously process multiple fields of view, macroscopic illustrative panels, and clinical charts. The agent must fuse this varied visual evidence with auxiliary structured texts to deduce diagnoses, capturing the varied staining qualities and artifact noise typical of uncurated clinical environments.

\textbf{SMU-281.} 
The SMU-281 dataset is a meticulously curated private cohort of 281 complex, ROI-level diagnostic scenarios. Similar to SMU-VL, it challenges models with a rigorous \emph{multi-image} paradigm, requiring the integration of multiple H\&E microscopy patches per case alongside diagnostic-scenario text. For evaluation, cases are partitioned into three mutually exclusive anatomical strata that collectively span the diagnostic spectrum: \emph{Breast \& Chest} (68 cases), covering breast, respiratory, and mediastinal pathology; \emph{GU \& Skin \& H\&N} (80 cases), covering female genital, genitourinary, bone and soft tissue, and head and neck diseases; and \emph{GI \& Thyroid \& Other} (133 cases), encompassing gastrointestinal, hepatopancreatobiliary, thyroid and parathyroid, hematolymphoid, neurological, endocrine, cardiovascular, and miscellaneous organ patterns (Appendix Table~\ref{tab:bench:roi_smu281}).

\subsubsection*{WSI-Level Multimodal Tasks}
Representing the pinnacle of clinical authenticity and computational challenge, this final tier evaluates the agent's ability to navigate gigapixel WSIs. Crucially, this tier is composed entirely of private, real-world cases derived from surgical pathology workflows across multiple anonymized hospitals. We define \textbf{10} WSI-level multiple-choice VQA tasks totaling 4,080 cases, spanning colorectal, gastric, lung, and breast pathology and covering tumor grading, staging, and subtyping under both retrospective and prospective study designs (Appendix Table~\ref{tab:benchmark_datasets}).

\textbf{Colorectal Cancer (CRC).}
The colorectal tier comprises three complementary tasks (2,471 cases). \emph{CRC-T Staging} (444 retrospective WSIs) requires fine-grained T-stage assignment across T1--T4, demanding multi-scale gigapixel context resolution. \emph{CRC-Biopsy} (916 retrospective biopsy WSIs) tests multiple-choice differential diagnosis on colorectal biopsy specimens, including adenomatous lesions, invasive adenocarcinoma, and related diagnostic categories. To assess robustness under temporal distribution shift, \emph{CRC-Biopsy (Prosp.)} (667 prospective cases) mirrors this biopsy classification paradigm on recently accrued clinical intakes.

\textbf{Gastric Cancer.}
The gastric tier includes three tasks (1,277 cases) dedicated to holistic WSI-level interpretation of gastric pathology. \emph{Gastric-Grading} (400 retrospective WSIs) evaluates histologic grade assignment directly from surgical pathology slides. \emph{Gastric-Subtyping} (396 retrospective WSIs) challenges the model with pathologic subtyping across major gastric adenocarcinoma patterns. To rigorously validate clinical readiness beyond retrospective-only evaluation, \emph{Gastric-CAG (Prosp.)} is a strictly \emph{prospective} cohort of 481 gastric biopsy WSIs focused on screening and diagnosis of chronic atrophic gastritis versus non-atrophic or normal mucosa.

\textbf{Lung Cancer.}
\emph{Lung-NSCLC} comprises 599 retrospective lung WSIs evaluating non-small cell lung carcinoma (NSCLC) subtyping at the slide level, requiring the model to discriminate lung adenocarcinoma (LUAD) from lung squamous cell carcinoma (LUSC) by integrating morphology across representative WSI patches.

\textbf{Breast Cancer.}
Sourced from an independent clinical institution, the breast tier consists of two retrospective tasks (177 cases). \emph{Breast-Grading} (88 WSIs) evaluates Nottingham histologic grade assignment by comprehensively analyzing nuclear pleomorphism, tubule formation, and mitotic counts across the entire slide. \emph{Breast-TNM-T} (89 WSIs) assesses pathologic T-category estimation (T1--T3) for breast cancer, representing a high-complexity staging task that requires coherent integration of localized morphological features across gigapixel tissue contexts.

\subsubsection*{Compared Methods}

Although pathology-oriented conversational systems such as PathChat\cite{lu2024multimodal} and SlideSeek\cite{chen2025evidence} are closely related to our setting, they are not publicly accessible for independent evaluation. We therefore compare PathPocket against accessible baselines spanning three categories: general-purpose multimodal large language models (Grok~4.1 Fast, GPT-4.1 mini, GPT-5 mini, and Qwen3.7 Plus); biomedical and histopathology multimodal large language models (LLaVA-Med\cite{li2023llava} and Quilt-LLaVA\cite{quiltllava}); and pathology vision-language foundation models (PLIP\cite{plip} and CONCH\cite{conch}). Text-only tasks use the general-purpose models, whereas ROI- and WSI-level tasks additionally include the biomedical and pathology-specific baselines. 

\noindent \textbf{Grok 4.1 Fast.}
Grok 4.1 Fast constitutes a high-speed, cost-effective multimodal large language model developed by xAI. A key innovation of the model is its natively integrated dual-mode switching mechanism between "Reasoning" and "Non-Reasoning," which allows the system to dynamically invoke chain-of-thought (CoT) or directly generate logical outcomes based on task complexity. Featuring an ultra-long context window of up to 2M tokens alongside extensive coverage of deep industry knowledge in science and finance, Grok 4.1 Fast significantly enhances long-text processing and multimodal image-text analysis for complex, multi-step real-time problems while maintaining exceptionally low first-token latency.

\noindent \textbf{GPT-4.1 mini.}
GPT-4.1 mini constitutes a lightweight, low-latency multimodal large language model from OpenAI, designed as a highly cost-effective alternative to traditional small-scale models. While maintaining a 128K token context window, the model is specifically optimized for high-frequency, high-volume mixed text and image tasks, demonstrating exceptional capabilities in structured JSON output and function calling. Despite its streamlined parameter count, GPT-4.1 mini delivers intelligent performance that far exceeds peers in its class across various academic benchmarks, particularly in multimodal reasoning and cross-lingual understanding.

\noindent \textbf{GPT-5 mini.}
GPT-5 mini constitutes a compact, high-tier reasoning model within OpenAI's flagship lineage, representing a major breakthrough in the multi-step problem-solving capabilities. The model innovatively introduces an adaptive multi-stage routing protocol and sparse attention mechanisms, significantly reducing computational overhead across a dense 400K context while endowing the model with deep reasoning capacities via internally generated long thinking tokens. Supported by native instruction-following and reinforced safety alignment technologies, GPT-5 mini achieves rigorous logic performance comparable to early full-scale frontier models, all while substantially lowering enterprise deployment costs and latency.

\noindent \textbf{Qwen3.7 Plus.}
Qwen3.7 Plus constitutes a general-purpose language model within the Tongyi Qwen series, engineered for balanced high performance and advanced long-text evolution. Utilizing an upgraded Mixture-of-Experts (MoE) architecture and optimized activation function control, the model achieves high data throughput efficiency across massive multilingual general knowledge, advanced mathematical logic competitions, and complex multi-turn extended dialogues. Benefiting from deep training on full-stack development instruction sets and structured reasoning chains, Qwen3.7 Plus delivers flagship-level comprehension alongside excellent inference efficiency.

\noindent \textbf{CONCH.}
CONCH (CONtrastive learning from Captions for Histopathology)\cite{conch} constitutes a vision-language foundation model tailored specifically for the field of computational pathology. Leveraged by pre-training on a diverse proprietary dataset of over 1.17 million high-quality histopathology image-caption pairs using task-agnostic contrastive learning, the model effectively overcomes the over-reliance of traditional pathology AI on specific stains (such as H\&E) or public WSI  datasets. CONCH demonstrates state-of-the-art performance across diverse benchmarks, including tissue image classification and cross-modal retrieval, significantly accelerating zero-shot or few-shot intelligent clinical diagnostic workflows.

\noindent \textbf{PLIP.}
PLIP (Pathology Language and Image Pre-Training)\cite{plip} constitutes the first large-scale multimodal foundation model in pathology derived from open-access medical community knowledge. By introducing the OpenPath image-text dataset, the model applies deep pathology-domain fine-tuning and cross-modal feature space alignment to the classic CLIP\cite{clip} architecture. PLIP successfully circumvents the bottleneck of traditional pathology AI, which strictly requires dense, manual pixel-level annotations. With robust zero-shot classification and linear probing capabilities, it provides essential architectural support for global medical knowledge sharing and precision digital pathology education.

\noindent \textbf{Zero-shot prompting modes for CONCH and PLIP.}
Because CONCH and PLIP are dual-encoder vision-language models rather than generative MLLMs, multiple-choice VQA is evaluated by zero-shot image-text matching following the CLIP-style paradigm\cite{clip,conch,plip}: the image embedding is compared (cosine similarity) with a text embedding for every candidate choice, and the highest-scoring choice is taken as the prediction. To reflect sensitivity to prompt formulation, we evaluate two complementary text constructions for each option $c$:
\begin{itemize}
    \item \textbf{(O)} \emph{option-only}: the text prompt contains only the wording of choice $c$ (class label / answer option), without the clinical question stem;
    \item \textbf{(Q+O)} \emph{question+option}: the text prompt concatenates the question stem with choice $c$, so that option scoring is conditioned on the full query context.
\end{itemize}
For readability, main-text ROI and WSI figures report CONCH and PLIP using, for each model and task, the prompting mode with higher accuracy. Complete results for both (O) and (Q+O) modes are provided in Appendix Tables~\ref{tab:bench:roi_public}, \ref{tab:bench:roi_smu281}, \ref{tab:bench:roi_smuvl}, and \ref{tab:bench:wsi_private}.

\noindent \textbf{LLaVA-Med.}
LLaVA-Med \cite{li2023llava} emerges as a biomedical multimodal large language model specifically designed for medical image understanding and reasoning, developed through curriculum learning in the extensive collection of biomedical literature from PubMed Central. The model architecture builds upon LLaVA's foundation while incorporating domain-specific adaptations through two key phases: biomedical concept alignment using 1.6 million image-caption pairs from PMC-15M dataset, followed by instruction tuning with GPT-4 generated medical question-answer pairs. This approach enables LLaVA-Med to achieve state-of-the-art performance on biomedical VQA benchmarks.

\noindent \textbf{Quilt-LLaVA.}
Quilt-LLaVA \cite{quiltllava} represents a specialized vision-language model for histopathology analysis, built upon the LLaVA framework with significant domain-specific adaptations. The model employs a two-stage training approach: initial vision-text alignment using 107K curated histopathology question-answer pairs (Quilt-Instruct) derived from educational YouTube videos, followed by instruction tuning with spatially-grounded QA pairs generated through mouse cursor tracking.

\subsubsection*{Evaluation Metrics}
\label{sec:method-metrics}

\paragraph{Automated benchmark evaluation.}
To evaluate the diagnostic reasoning capabilities of PathPocket across the diverse text and multimodal tasks in our benchmark, we employ \textbf{Accuracy (ACC)} as the primary quantitative metric. 
ACC is strictly defined as the proportion of questions where the final output letter generated by the Response Generation Agent exactly matches the ground-truth diagnostic label. Formally, given a dataset of $N$ pathology cases $\{(q_i, y_i)\}_{i=1}^N$, where $q_i$ is the multimodal clinical query and $y_i$ is the ground-truth option, the predicted answer is denoted as $\hat{y}_i = \mathcal{A}_{\mathrm{rg}}(q_i, \mathcal{E}_i)$. The accuracy is calculated as:
\begin{equation}
\text{ACC} = \frac{1}{N} \sum_{i=1}^{N} \mathbb{I}(\hat{y}_i = y_i) \times 100\%,
\end{equation}
where $\mathbb{I}(\cdot)$ is the indicator function. Unless otherwise noted, binomial proportions are summarized with Wilson score 95\% confidence intervals\cite{wilson1927probable}. The generated rationale accompanying the final answer is qualitatively reviewed by senior pathologists to verify the absence of hallucinations and the correctness of the retrieved evidence citations.

\begin{figure}[H]
    \centering
    \includegraphics[width=\textwidth]{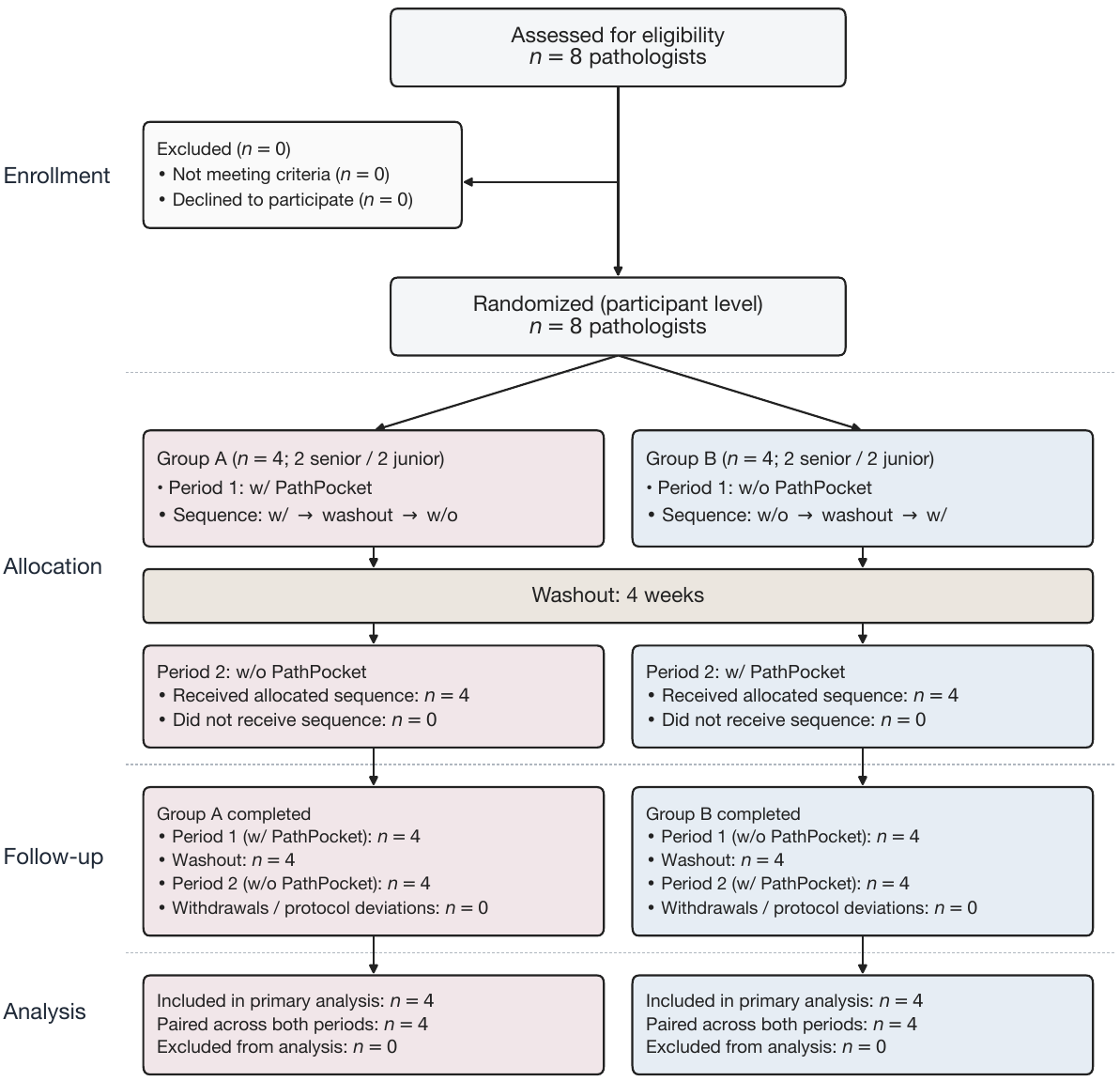}
    \caption{\textbf{CONSORT flow diagram for the PathPocket randomized two-period crossover reader study.}
    Eight board-certified pathologists were assessed for eligibility and randomized at the participant level to Group~A or Group~B ($n=4$ each; 2~senior / 2~junior per arm).
    Group~A completed Period~1 \textbf{with} PathPocket (w/), then a 4-week washout, then Period~2 \textbf{without} PathPocket (w/o); Group~B completed the reverse sequence.
    The w/o condition allows independent diagnosis or other search tools and is not restricted to unaided recall.
    All randomized participants completed both periods and were included in the paired primary analysis ($n=8$; none excluded).
    Item-bank composition and endpoints are detailed in Table~\ref{tab:RS_questions} and the Methods.
    Reporting follows CONSORT guidance\cite{turner2012consolidated}.}
    \label{fig:RS_consort}
\end{figure}

\paragraph{Pathologist reader study.}
In the crossover pathologist reader study, outcomes are defined at the level of scored response records. Participant enrollment, randomization, washout, and analysis inclusion follow CONSORT reporting (Appendix Figure~\ref{fig:RS_consort}). The study comprises 150 unique questions (100 text-only and 50 multimodal). Each text-only question contains two open-ended sub-questions that are scored independently and therefore contribute two scoring units (100$\times$2 = 200), which together with the 50 multimodal items yield 250 scored items and $M=2{,}000$ pathologist-level response records (250$\times$8 pathologists). Let $u=1,\ldots,M$ index these scoring units. Each unit receives a \textbf{normalized answer score} $s_u \in [0,1]$:
\begin{itemize}
    \item Multiple-choice items: $s_u = \mathbb{I}(\hat{y}_u = y_u)$, where $\hat{y}_u$ is the pathologist's selected option and $y_u$ is the reference key.
    \item Open-ended items: an independent LLM grader (GPT-5 mini) assigns an integer rubric score $r_u \in \{1,\ldots,5\}$ by comparing the pathologist answer against the reference answer; the score is normalized as $s_u = r_u/5$. For text-only items, the two sub-questions are graded independently within a single dual-rubric call and each contributes a separate $s_u$; multimodal open-ended items are scored once for the full response. The exact grading prompts are provided in Appendix~\ref{sec:appendix_reader_scoring_prompts}.
\end{itemize}
The normalized answer accuracy reported in the reader study is the arithmetic mean $\bar{s} = \frac{1}{M}\sum_{u=1}^{M} s_u$, expressed as a percentage. Condition comparisons for accuracy and confidence use one-sided paired $t$-tests on pathologist-level means\cite{student1908probable}.

\textbf{Diagnostic confidence} is the pathologist's self-reported certainty on a 1--10 Likert scale\cite{likert1932technique} immediately after each response, denoted $c_u$. Summary statistics are computed as the sample mean $\bar{c} = \frac{1}{M}\sum_{u=1}^{M} c_u$ within each assistance condition (with or without PathPocket).

\textbf{Inter-pathologist agreement} is quantified separately by question type. For open-ended items, because each scored item is rated by a fixed rater cohort of size $k$ under the crossover design, we report the average-measure two-way random-effects intraclass correlation \textbf{ICC(2,$k$)}\cite{shrout1979intraclass} from pathologist-aligned score matrices $\mathbf{X} \in \mathbb{R}^{n \times k}$ (rows = scored items, including text-only sub-questions; columns = pathologists within a rater cohort), and pool cohort-specific estimates by item-count weighting:
\begin{equation}
\text{ICC}(2,k) = \frac{MS_R - MS_E}{MS_R + \frac{1}{n}(MS_C - MS_E)},
\end{equation}
where $n$ is the number of items, $k$ is the number of pathologists in the cohort, and $MS_R$, $MS_C$, and $MS_E$ denote the mean squares for rows (items), columns (raters), and residual error, respectively. In parallel, we report \textbf{tolerance agreement}: the pairwise fraction of pathologist pairs whose normalized scores differ by at most one rubric step ($|s_i-s_j|\leq 0.2$ on the $1$--$5$ scale after $/5$ normalization).

For multiple-choice items, we compute \textbf{Fleiss' $\kappa$}\cite{fleiss1971measuring} across pathologists within each question. For question $i$ with $n$ raters and $C$ categories, let $n_{ij}$ denote the number of raters assigning category $j$. The per-question agreement is
\begin{equation}
P_i = \frac{1}{n(n-1)}\left(\sum_{j=1}^{C} n_{ij}^2 - n\right),
\end{equation}
and the overall observed agreement is $\bar{P} = \frac{1}{N}\sum_{i=1}^{N} P_i$ over $N$ eligible questions. With category prevalence $p_j = \frac{1}{Nn}\sum_{i=1}^{N} n_{ij}$, the chance agreement is $P_e = \sum_{j=1}^{C} p_j^2$, and
\begin{equation}
\kappa = \frac{\bar{P} - P_e}{1 - P_e}.
\end{equation}
We report option-level agreement with $C=5$ (categories A--E).

During PathPocket-assisted phases, pathologists additionally rate the clinical usefulness of the retrieved assistance on a 1--10 scale. The \textbf{assistance content quality rating} is summarized by its sample mean across assisted responses.

\subsection{Implementation Details}
\label{sec:method-implementation}
The framework of PathPocket is implemented in PyTorch\cite{paszke2019pytorch}. All computational experiments, including hypergraph construction, dense embedding generation, and multi-agent inference, are conducted on a high-performance on-premise computing cluster equipped with 8 $\times$ NVIDIA H800 GPUs (80GB VRAM per card). The specific foundational tools and models integrated into the multi-agent system are sourced from their official open-source repositories.
MinerU\cite{wang2024mineru}: Utilized by the Parsing Agent for layout-aware PDF structuralization and data extraction from pathology textbooks (\url{https://github.com/opendatalab/MinerU}).
TITAN\cite{ding2025multimodal}: Employed for attention-based multi-scale ROI selection from gigapixel WSIs, operating on CONCH~v1.5 patch features (Appendix~\ref{sec:appendix_wsi_patch}; \url{https://github.com/mahmoodlab/TITAN}).
Virchow2\cite{virchow2}: The pathology-specific vision foundation model used for embedding multimodal nodes and visual ROIs (\url{https://huggingface.co/paige-ai/Virchow2}).
BGE-m3\cite{bge-m3}: The state-of-the-art multilingual text encoder utilized for computing dense vector representations of text chunks and hyperedges (\url{https://github.com/FlagOpen/FlagEmbedding}).
Qwen3-Reranker\cite{qwen3embedding}: The LLM-based model utilized by the Evidence Filtering Agent to deeply evaluate semantic relevance and truncate context noise (\url{https://huggingface.co/Qwen/Qwen3-Reranker-8B}).
\textbf{Hypergraph construction.}
The LLM-driven stages of pathology hypergraph construction, including structured entity extraction, relation mining, and hyperedge synthesis from parsed literature， are served via vLLM (\url{https://github.com/vllm-project/vllm}), a high-throughput and memory-efficient inference engine, using the open-weight \textbf{Qwen3-30B}\cite{yang2025qwen3} model as the backbone generator. Given the scale of the evidence corpus ($\sim$110,000 documents yielding $>$4.5 million entities and $>$7 million relations), hypergraph construction is a high-volume offline workload in which inference cost and throughput are critical. In pilot ablations, Qwen3-30B already achieved strong structured extraction quality relative to larger models, providing a favorable accuracy--efficiency trade-off for large-scale deployment.
\textbf{Clinical inference.}
For benchmark evaluation, PathPocket's multi-agent reasoning workflow is powered by a task-adaptive MLLM. We first benchmark four API-accessible models in isolation on each evaluation task: Grok-4.1-Fast, GPT-4.1 mini, GPT-5 mini, and Qwen3.7 Plus. For every task, PathPocket adopts whichever model achieved the highest standalone accuracy on that task to drive the agentic inference pipeline. Concretely, Qwen3.7 Plus is used for most text-only tasks (pathologist-generated USMLE-style and NMLE-style items, and SMU-Text), public ROI datasets, SMU-281, and the majority of WSI tasks; GPT-5 mini for SMU-VL 1--6 and USMLE 2; and GPT-4.1 Mini for CRC-T Staging. In contrast, the pathologist reader study fixes PathPocket assistance to Qwen3.7 Plus for all questions and both assistance phases, ensuring a uniform human--AI interaction setting independent of per-task backbone selection.

\section*{Data Availability}
\label{sec:data_availability}

The data supporting this study fall into three categories: knowledge sources used for hypergraph construction, open-access datasets, and restricted private clinical data.

\textbf{Knowledge Sources for Hypergraph Construction.}
The evidence base for the multimodal pathology hypergraph was constructed from both open and copyrighted materials (Appendix Table~\ref{tab:evidence_sources}). The majority of the documents are freely accessible in the public domain or distributed under Creative Commons licenses (CC BY or CC BY-NC). A portion of proprietary guidelines and textbooks are acquired through official purchase and download channels. All knowledge sources are used strictly for scientific research and only distilled factual medical knowledge are extracted for hypergraph modeling without reproducing or redistributing original text or figures, and raw copyrighted materials will not be publicly shared.

\textbf{Open-Access Datasets.}
The public ROI datasets (BreakHis, CCRCC, and Chaoyang) are available from their original releases as detailed in Appendix Table~\ref{tab:evidence_sources}.

\textbf{Restricted Clinical Data.}
The private data, including the text-only examination-style tasks (NMLE~1--3 and USMLE~1--3), SMU-Text~1--6, SMU-281, SMU-VL~1--6, and the nine WSI-level tasks (CRC-T Staging, CRC-Biopsy, CRC-Biopsy~(Prosp.), Gastric-Grading, Gastric-Subtyping, Gastric-CAG~(Prosp.), Lung-NSCLC, Breast-Grading, and Breast-TNM-T; Appendix Table~\ref{tab:evidence_sources}), contain de-identified protected health information and proprietary whole-slide images. In compliance with institutional review board regulations and hospital data-protection policies, these datasets are not publicly released. Researchers may request access to de-identified data from the corresponding author, subject to ethical approval and institutional agreements.

\section*{Code Availability}
The code for PathPocket is available at \url{https://github.com/zhexu1997/PathPocket}. For the competing methods, we utilize a combination of cloud-based APIs and official model checkpoints. Specifically, evaluations for Grok 4.1 Fast, GPT-4.1 mini, and GPT-5 mini are performed via Microsoft Azure AI Foundry services (\url{https://ai.azure.com}), while Qwen3.7 Plus is accessed through the official Alibaba Cloud API (\url{https://bailian.console.aliyun.com}). For the specialized pathology foundation models and MLLMs, PLIP (\url{https://github.com/pathologyfoundation/plip}), CONCH (\url{https://github.com/mahmoodlab/CONCH}), LLaVA-Med (\url{https://github.com/microsoft/LLaVA-Med}), and Quilt-LLaVA (\url{https://github.com/aldraus/quilt-llava}) are evaluated using their official open-source weights and implementations.

\section*{Ethics Declarations}
This project has been reviewed and approved by the Human and Artefacts Research Ethics Committee (HAREC) of Hong Kong University of Science and Technology (protocol no. HREP-2026-0168). The prospective study protocol was approved
by the Medical Ethics Committee of Nanfang Hospital, Southern Medical University (protocol no. NFEC-2025-403). The
corresponding protocol was registered on ClinicalTrials.gov (ID: NCT07157618).

\section*{Author Contribution}
Z.X. and H.C. conceived and designed the work. Z.X. designed and performed the experiments, conducted statistical analysis, and wrote the manuscript with inputs from all authors. Z.Z. coordinated the prospective study and enrolled patients, provided pathological expertise, and curated private datasets. Z.X., Z.C., J.X., and Y.N. curated and preprocessed the public data included in the paper. Y.L., Z.L., J.H., and H.W. participated in discussions on experimental design. Z.X., Y.W., J.M., L.L., Y.X., and Z.G. preprocessed the private datasets. D.L., Z.Z., D.T., Z.C., Y.T., X.L., and Q.X. participated as pathologists in the reader study. All authors reviewed and approved the final paper. H.C. supervised the research.

\section*{Acknowledgment}
This work is supported by Hong Kong Innovation and Technology Commission (Project No. MHP/002/22, PRP/086/24FX and ITC-SKLNSD26SC01), Shenzhen Science and Technology Innovation Committee Fund (Project No. KCXFZ20230731094059008) and Research Grants Council of the Hong Kong Special Administrative Region, China (Project No. R6003-22 and C4024-22GF).

\bibliography{sample.bib}

\section*{Appendix}

\renewcommand{\thesubsection}{A.\arabic{subsection}}
\setcounter{subsection}{0}

\clearpage

\begin{table}[]
\centering
\caption{\textbf{Sources and URLs for the pathology evidence corpus and public evaluation datasets.} Evidence documents were stratified into an 8-tier evidence hierarchy adapted from standard evidence-based medicine frameworks (e.g., GRADE\cite{guyatt2008grade}). Literature resources were accessed via public open-access repositories or authorized institutional licenses; public ROI benchmark datasets were accessed from their original releases.}
\label{tab:evidence_sources}
\begin{tabularx}{\textwidth}{@{} p{7cm} X @{}}
\toprule
\textbf{Source} & \textbf{URL} \\
\midrule
\multicolumn{2}{@{}l}{\textbf{Guideline}} \\
PubMed Central & \url{https://www.ncbi.nlm.nih.gov/pmc/} \\
World Health Organization & \url{https://publications.iarc.fr/} \\
Chinese Anti-Cancer Association & \url{http://www.caca.org.cn/} \\
Chinese Society of Clinical Oncology & \url{http://www.csco.org.cn/} \\
National Health Commission of the PRC & \url{http://www.nhc.gov.cn/} \\
\midrule
\multicolumn{2}{@{}l}{\textbf{Systematic Review \& Meta Analysis}} \\
PubMed Central & \url{https://www.ncbi.nlm.nih.gov/pmc/} \\
\midrule
\multicolumn{2}{@{}l}{\textbf{Randomized Controlled Trial}} \\
PubMed Central & \url{https://www.ncbi.nlm.nih.gov/pmc/} \\
\midrule
\multicolumn{2}{@{}l}{\textbf{Observational \& Cohort Study}} \\
PubMed Central & \url{https://www.ncbi.nlm.nih.gov/pmc/} \\
\midrule
\multicolumn{2}{@{}l}{\textbf{Case Report}} \\
PubMed Central & \url{https://www.ncbi.nlm.nih.gov/pmc/} \\
\midrule
\multicolumn{2}{@{}l}{\textbf{Consensus}} \\
PubMed Central & \url{https://www.ncbi.nlm.nih.gov/pmc/} \\
\midrule
\multicolumn{2}{@{}l}{\textbf{Textbook}} \\
Springer Nature Open Access & \url{https://www.springernature.com/cn/open-science/journals-books/books} \\
The Directory of Open Access Books & \url{https://directory.doabooks.org} \\ 
NCBI Bookshelf & \url{https://www.ncbi.nlm.nih.gov/books/} \\ 
OpenStax & \url{https://openstax.org} \\
\midrule
\multicolumn{2}{@{}l}{\textbf{Expert Opinion}} \\
PubMed Central & \url{https://www.ncbi.nlm.nih.gov/pmc/} \\
\midrule
\multicolumn{2}{@{}l}{\textbf{Public Datasets}} \\
CCRCC \cite{brummer2023computational} & \url{https://zenodo.org/records/7898308} \\
Chaoyang \cite{zhu2021hard} & \url{https://github.com/bupt-ai-cz/HSA-NRL} \\
BreakHis \cite{spanhol2015dataset} & \url{https://www.kaggle.com/datasets/ambarish/breakhis} \\
\bottomrule
\end{tabularx}
\end{table}

\clearpage

\clearpage
\subsection{Prompt Templates for the Multi-Agent Workflow}
\label{sec:appendix_prompts}

This section details the exact prompt templates used by the specialized LLM agents within PathPocket's workflow. The prompts utilize specific delimiters (e.g., \texttt{<|\#|>}) and placeholders (e.g., \texttt{\{entity\_types\}}) which are dynamically populated during runtime.

\begin{tcolorbox}[
    breakable,
    colback=teal!5,
    colframe=teal!70,
    boxrule=0.5pt,
    arc=2pt,
    left=6pt, right=6pt, top=6pt, bottom=6pt,
    fonttitle=\bfseries,
    title={Hypergraph Construction: Extraction Agent}]
\begin{Verbatim}[breaklines=true, breakanywhere=true]
---Role---
You are an Expert Pathology Knowledge Graph Specialist. Your goal is to extract structured medical entities and relationships from texts. The knowledge must be accurate, universal, and useful for patients, students, and clinicians.

---Instructions---
1.  **Entity Extraction:**
    *   **Identification:** Identify clinically and pathologically significant entities.
    *   **Entity Details:**
        *   `entity_name`: Use standard medical terminology. Capitalize Title Case. Ensure consistency (e.g., use "Renal Cell Carcinoma" consistently).
        *   `entity_type`: Categorize using: `{entity_types}`. If none apply, use `Other`.
        *   `entity_description`: Provide a concise definition focusing on clinical or pathological significance (e.g., etiology, morphology, mechanism). Avoid context-specific temporality (e.g., "observed in this patient").
    *   **Output Format:** `entity{tuple_delimiter}entity_name{tuple_delimiter}entity_type{tuple_delimiter}entity_description`

2.  **Relationship Extraction:**
    *   **Identification:** Identify objective, factual connections such as etiology, pathogenesis, diagnostic criteria, treatment efficacy, or prognosis.
    *   **Multi-Entity Relations:** Capture relationships involving two or more entities (e.g., "Drug X combined with Drug Y treats Disease Z in Organ W" should be a single relation connecting all four entities).
    *   **Relationship Details:**
        *   `entities`: List of ALL entity names involved in this relationship, separated by `{tuple_delimiter}`. Must match extracted entity names exactly.
        *   `relationship_keywords`: High-level medical concepts. Comma-separated.
        *   `relationship_description`: Explain the medical logic connecting ALL the entities (e.g., mechanism of action, causal link, synergistic effects).
    *   **Output Format:** `relation{tuple_delimiter}entity_name_1{tuple_delimiter}entity_name_2{tuple_delimiter}...{tuple_delimiter}entity_name_n{tuple_delimiter}relationship_keywords{tuple_delimiter}relationship_description`
    *   **Note:** The last two fields are always keywords and description.

3.  **General Protocols:**
    *   **Delimiter:** Use `{tuple_delimiter}` strictly as a separator.
    *   **Directionality:** Relationships like "treats," "causes," or "indicates" are directed. Ensure logical flow.
    *   **Objectivity:** Use third-person medical language. No pronouns.
    *   **Language:** Output in {language}. Keep proper nouns (e.g., gene names like *EGFR*, drugs) in standard medical English.
    *   **Completion:** End with `{completion_delimiter}`.

---Examples---
{examples}

---Input Data---
Entity_types: [{entity_types}]
Text:
{input_text}
\end{Verbatim}
\end{tcolorbox}

\vspace{1em}

\begin{tcolorbox}[
    breakable,
    colback=teal!5,
    colframe=teal!70,
    boxrule=0.5pt,
    arc=2pt,
    left=6pt, right=6pt, top=6pt, bottom=6pt,
    fonttitle=\bfseries,
    title={Hypergraph Construction: Rectifying Agent}]
\begin{Verbatim}[breaklines=true, breakanywhere=true]
---Task---
Based on the last extraction task, identify and extract any **missed or incorrectly formatted** entities and relationships from the input text.

---Instructions---
1.  **Strict Adherence to System Format:** Strictly adhere to all format requirements for entity and relationship lists, including output order, field delimiters, and proper noun handling, as specified in the system instructions.
2.  **Focus on Corrections/Additions:**
    *   **Do NOT** re-output entities and relationships that were **correctly and fully** extracted in the last task.
    *   If an entity or relationship was **missed** in the last task, extract and output it now according to the system format.
    *   If an entity or relationship was **truncated, had missing fields, or was otherwise incorrectly formatted** in the last task, re-output the *corrected and complete* version in the specified format.
3.  **Output Format - Entities:** Output a total of 4 fields for each entity, delimited by `{tuple_delimiter}`, on a single line. The first field *must* be the literal string `entity`.
4.  **Output Format - Relations:** Output at least 5 fields for each relation, delimited by `{tuple_delimiter}`, on a single line. The first field *must* be the literal string `relation`. The last two fields are always keywords and description. All fields in between are entity names (2 or more entities).
5.  **Output Content Only:** Output *only* the extracted list of entities and relationships. Do not include any introductory or concluding remarks, explanations, or additional text before or after the list.
6.  **Completion Signal:** Output `{completion_delimiter}` as the final line after all relevant missing or corrected entities and relationships have been extracted and presented.
7.  **Output Language:** Ensure the output language is {language}. Proper nouns (e.g., personal names, place names, organization names) must be kept in their original language and not translated.

<Output>
\end{Verbatim}
\end{tcolorbox}

\vspace{1em}

\begin{tcolorbox}[
    breakable,
    colback=teal!5,
    colframe=teal!70,
    boxrule=0.5pt,
    arc=2pt,
    left=6pt, right=6pt, top=6pt, bottom=6pt,
    fonttitle=\bfseries,
    title={Pathology Reasoning: Case Understanding Agent}]
\begin{Verbatim}[breaklines=true, breakanywhere=true]
---Role---
You are an expert pathology query analyzer for a multimodal pathology knowledge hypergraph RAG system. Decompose the user question into a **single JSON object** for retrieval. Output language for string values: {language}. Keep standard drug/gene symbols in common medical English when natural.

---Goal---
Return one JSON object with **exactly** these keys:

- **site**: **One string only** — anatomical site, organ, specimen source, or localization. Use `""` if none.
- **gross_entities**: Macroscopic / specimen-level cue phrases — JSON **array of strings** or one string (e.g. specimen color, size, number of pieces, cut surface). Use `[]` if there is no gross/specimen narrative.
- **gross_description**: **One string only** — copy or lightly trim the **original gross / specimen ** description from the stem. 
- **morphology_entities**: Named **microscopic** morphology / histology entities — JSON **array of strings** or one string.
- **morphology_description**: **One string only** — copy or lightly trim the **original microscopic / histology** description from the stem. Keep the question’s wording; do not replace with abstract keywords. Use `""` if there is no microscopic text.
- **marker_entities**: IHC / molecular marker names — array of strings or one string.
- **marker_description**: **One string only** — copy the **original** immunohistochemistry / special stain / molecular lines from the stem, verbatim or with minimal cleanup. Use `""` if none.
- **clinical_entities**: Clinical / setting entities (besides age/sex) — array of strings or one string.
- **clinical_description**: **One string only** — start with **sex and age** exactly as in the stem. Use `""` only if the query truly has no such info. 
- **other_entities**: Other key entities/concepts from the stem that don't fit the above categories — array of strings or one string.
- **candidate_answer**: MCQ: **list** of option texts (one string per option). Otherwise `[]` or one short string.

---Instructions---
1. **Output**: Valid JSON only — no markdown fences, no text before/after.
2. **Source split (CRITICAL)**:
   - **Stem-only**: You MUST derive **site**, **gross_entities**, **gross_description**, **morphology_entities**, **morphology_description**, **marker_entities**, **marker_description**, **clinical_entities**, **clinical_description**, **other_entities** ONLY from the **stem**. Do NOT use the MCQ options to invent or supplement these fields.
   - **Options-only**: You MUST derive **candidate_answer** ONLY from the option texts when options are present.
   - If the query contains both stem and options, treat them as two separate sources with the above rules.
3. **Types**: `site`, `gross_description`, `morphology_description`, `marker_description`, and `clinical_description` must be **strings** (never JSON arrays). Use `""` when empty.
4. **Concise** for `site` and `*_entities`; description fields may stay close to full original sentences where helpful.
5. **Non-pathology / garbage queries**: Return all keys with `[]` or `""` as appropriate.

---Examples---
{examples}

---Real Data---
User Query: {query}

---Output---
Output:
\end{Verbatim}
\end{tcolorbox}

\vspace{1em}

\begin{tcolorbox}[
    breakable,
    colback=teal!5,
    colframe=teal!70,
    boxrule=0.5pt,
    arc=2pt,
    left=6pt, right=6pt, top=6pt, bottom=6pt,
    fonttitle=\bfseries,
    title={Pathology Reasoning: Response Generation Agent}]
\begin{Verbatim}[breaklines=true, breakanywhere=true]
---Role---

You are an expert AI assistant specializing in synthesizing information from a provided knowledge base. Your primary function is to answer user queries accurately by ONLY using the information within the provided **Context**.

---Goal---

Generate a comprehensive, well-structured answer to the user query.
The answer must integrate relevant facts from the Knowledge Graph, Document Chunks and Retrieved Similar Images found in the **Context**.
Consider the conversation history if provided to maintain conversational flow and avoid repeating information.

---Instructions---

1. Step-by-Step Instruction:
  - Carefully determine the user's query intent in the context of the conversation history to fully understand the user's information need.
  - Scrutinize both `Knowledge Graph Data`, `Document Chunks`, and `Retrieved Similar Images` in the **Context**. Identify and extract all pieces of information that are directly relevant to answering the user query.
  - **Image handling (critical)**: Do NOT rely on your own visual interpretation of any images. Treat the `Retrieved Similar Images` section as an external knowledge source rather than as raw pixels to interpret. 
  - Consider the evidence source, evidence level, image similarity (if any), and anatomical structure match degree. Higher evidence level (smaller number), higher image similarity, and higher anatomical match degree indicate more trustworthy evidence. Prioritize information from more credible images when synthesizing the answer.
  - When there is any ambiguity in images (e.g., patterns that could be read multiple ways), prefer the **retrieval-grounded descriptions** in `Retrieved Similar Images` and `Document Chunks` over any intuition. 
  - Weave the extracted facts into a coherent and logical response. Your own knowledge must ONLY be used to formulate fluent sentences and connect ideas, NOT to introduce any external information.
  - Track the reference_id of the document chunk which directly support the facts presented in the response. Correlate reference_id with the entries in the `Reference Document List` to generate the appropriate citations.
  - Generate a references section at the end of the response. Each reference document must directly support the facts presented in the response.
  - Do not generate anything after the reference section.

2. Content & Grounding:
  - Strictly adhere to the provided context from the **Context**; DO NOT invent, assume, or infer any information not explicitly stated.
  - When multiple sources conflict, rely on those with higher evidence level (smaller number), higher image similarity, and higher anatomical structure match degree.

3. Formatting & Language:
  - The response MUST be in the same language as the user query.
  - The response MUST utilize Markdown formatting for enhanced clarity and structure (e.g., headings, bold text, bullet points).
  - The response should be presented in {response_type}.

4. References Section Format:
  - The References section should be under heading: `### References`
  - Reference list entries should adhere to the format: `* [n] Document Title`. Do not include a caret (`^`) after opening square bracket (`[`).
  - The Document Title in the citation must retain its original language.
  - Output each citation on an individual line
  - Provide maximum of 5 most relevant citations.
  - Do not generate footnotes section or any comment, summary, or explanation after the references.

5. Reference Section Example:
```
### References

- [1] Document Title One
- [2] Document Title Two
- [3] Document Title Three
```
6. Additional Instructions: {user_prompt}

---Context---

{context_data}
\end{Verbatim}
\end{tcolorbox}

\vspace{1em}

\begin{tcolorbox}[
    breakable,
    colback=teal!5,
    colframe=teal!70,
    boxrule=0.5pt,
    arc=2pt,
    left=6pt, right=6pt, top=6pt, bottom=6pt,
    fonttitle=\bfseries,
    title={Pathology Reasoning: Evidence Filtering Agent}]
\begin{Verbatim}[breaklines=true, breakanywhere=true]
<|im_start|>system
Judge whether the Document meets the requirements based on the Query and the Instruct provided. Note that the answer can only be "yes" or "no".<|im_end|>
<|im_start|>user
<Instruct>: Given a web search query, retrieve relevant passages that answer the query.
<Query>: {query}
<Document>: {document}<|im_end|>
<|im_start|>assistant
\end{Verbatim}
\end{tcolorbox}

\clearpage
\subsection{Hypergraph Topological Query (SQL)}
\label{sec:appendix_hypergraph_sql}

During inference, the Evidence Retrieval Agent issues the following PostgreSQL statements against the hypergraph store (and, in parallel, dense indexes over hyperedge / entity / chunk embeddings).

\begin{tcolorbox}[
    breakable,
    colback=gray!4,
    colframe=gray!55,
    boxrule=0.5pt,
    arc=2pt,
    left=6pt, right=6pt, top=6pt, bottom=6pt,
    fonttitle=\bfseries,
    title={1. Hyperedge topological query (array overlap)}]
\begin{Verbatim}[breaklines=true, breakanywhere=true]
-- Inputs: $1 = workspace, $2 = core_entities (text[]), $3 = min_entity_count
SELECT edge_id, entities, entity_count, edge_data, created_at
FROM {hyperedges_table}
WHERE workspace = $1
  AND entities && $2::text[]
  AND entity_count >= $3
ORDER BY
  (SELECT COUNT(*) FROM unnest(entities) e WHERE e = ANY($2::text[])) DESC,
  entity_count ASC
LIMIT 500;
\end{Verbatim}
\end{tcolorbox}

\vspace{1em}

\begin{tcolorbox}[
    breakable,
    colback=gray!4,
    colframe=gray!55,
    boxrule=0.5pt,
    arc=2pt,
    left=6pt, right=6pt, top=6pt, bottom=6pt,
    fonttitle=\bfseries,
    title={2. Entity lookup from hypergraph node table}]
\begin{Verbatim}[breaklines=true, breakanywhere=true]
-- Inputs: $1 = workspace, $2 = entity_keywords (text[]), $3 = limit
SELECT node_id, node_data
FROM {hypergraph_nodes_table}
WHERE workspace = $1
  AND (
    node_id = ANY($2::text[])
    OR node_data->>'entity_name' = ANY($2::text[])
  )
LIMIT $3;
\end{Verbatim}
\end{tcolorbox}

\vspace{1em}

\begin{tcolorbox}[
    breakable,
    colback=gray!4,
    colframe=gray!55,
    boxrule=0.5pt,
    arc=2pt,
    left=6pt, right=6pt, top=6pt, bottom=6pt,
    fonttitle=\bfseries,
    title={3. Dense retrieval over hyperedge embeddings (pgvector)}]
\begin{Verbatim}[breaklines=true, breakanywhere=true]
-- Inputs: $1 = workspace, $2 = max_cos_distance, $3 = top_k
SELECT r.entities, r.entity_count, r.relation, r.keywords,
       r.description, r.chunk_ids, r.file_path,
       EXTRACT(EPOCH FROM r.create_time)::BIGINT AS created_at,
       (r.content_vector <=> '[{embedding}]'::vector) AS distance
FROM VDB_RELATION r
WHERE r.workspace = $1
  AND r.content_vector <=> '[{embedding}]'::vector <= $2
ORDER BY r.content_vector <=> '[{embedding}]'::vector
LIMIT $3;
\end{Verbatim}
\end{tcolorbox}

\vspace{1em}

\begin{tcolorbox}[
    breakable,
    colback=gray!4,
    colframe=gray!55,
    boxrule=0.5pt,
    arc=2pt,
    left=6pt, right=6pt, top=6pt, bottom=6pt,
    fonttitle=\bfseries,
    title={4. Dense retrieval over entity embeddings (pgvector)}]
\begin{Verbatim}[breaklines=true, breakanywhere=true]
SELECT e.entity_name,
       EXTRACT(EPOCH FROM e.create_time)::BIGINT AS created_at
FROM VDB_ENTITY e
WHERE e.workspace = $1
  AND e.content_vector <=> '[{embedding}]'::vector <= $2
ORDER BY e.content_vector <=> '[{embedding}]'::vector
LIMIT $3;
\end{Verbatim}
\end{tcolorbox}

\vspace{1em}

\begin{tcolorbox}[
    breakable,
    colback=gray!4,
    colframe=gray!55,
    boxrule=0.5pt,
    arc=2pt,
    left=6pt, right=6pt, top=6pt, bottom=6pt,
    fonttitle=\bfseries,
    title={5. Dense retrieval over document chunks (pgvector)}]
\begin{Verbatim}[breaklines=true, breakanywhere=true]
SELECT c.id, c.content, c.file_path,
       EXTRACT(EPOCH FROM c.create_time)::BIGINT AS created_at,
       (c.content_vector <=> '[{embedding}]'::vector) AS distance
FROM VDB_CHUNKS c
WHERE c.workspace = $1
  AND c.content_vector <=> '[{embedding}]'::vector <= $2
ORDER BY c.content_vector <=> '[{embedding}]'::vector
LIMIT $3;
\end{Verbatim}
\end{tcolorbox}

\clearpage
\subsection{TITAN-guided Multi-scale WSI Patch Selection}
\label{sec:appendix_wsi_patch}

Gigapixel WSIs cannot be fed directly into the multimodal agents. PathPocket therefore selects a small set of diagnostically salient, multi-scale ROIs before reasoning (Appendix~\ref{sec:appendix_wsi_patch}).
The procedure has three stages: (i)~patch feature encoding, (ii)~TITAN attention scoring, and (iii)~greedy multi-scale ROI selection followed by cropping for the MLLM.
Unless otherwise noted, we use a base patch size $p{=}512$ (level~0), maximum pairwise IoU $\tau_{\mathrm{IoU}}{=}0.5$, and minimum grid margin $m{=}2$.

\paragraph{Inputs.}
For each slide we assume precomputed CONCH~v1.5\cite{conch} patch embeddings $\mathbf{F}\in\mathbb{R}^{N\times C}$ and corresponding level-0 coordinates $\{(x_i,y_i)\}_{i=1}^{N}$.
Grid indices are $(g^x_i,g^y_i)=(\lfloor x_i/p\rfloor,\lfloor y_i/p\rfloor)$.

\paragraph{Attention scoring.}
The TITAN vision encoder is applied to all $N$ patches and returns per-patch importance scores $\{a_i\}_{i=1}^{N}$ via attentional pooling over CONCH features and coordinates\cite{ding2025multimodal}.
Averaging $a_i$ within each occupied grid cell yields an attention map $A(g^x,g^y)$.

\paragraph{Multi-scale non-overlapping selection.}
We consider window scales $w\in\{8,4,2,1\}$ (coarse$\rightarrow$fine), corresponding to physical side lengths $w\cdot p$ (e.g., $4096$, $2048$, $1024$, and $512$ pixels when $p{=}512$).
For each $w$, every feasible $w\times w$ window on the attention grid is scored by the mean of $A$ over its cells.
Windows are then selected greedily in descending score order under two constraints relative to already chosen windows (across scales): pairwise IoU $\le\tau_{\mathrm{IoU}}$ and Chebyshev gap $\ge m$ grid cells.
The highest-scoring feasible candidate is retained at each scale.
Selected windows are exported as level-0 boxes $(x,y,\mathrm{size})$ with attention scores, then cropped from the WSI and resized (typically capped at $1024\times 1024$) for downstream multimodal inference.
Algorithm~\ref{alg:wsi_patch} summarizes the procedure.

\begin{tcolorbox}[
    breakable,
    colback=gray!4,
    colframe=gray!55,
    boxrule=0.5pt,
    arc=2pt,
    left=6pt, right=6pt, top=6pt, bottom=6pt,
    fonttitle=\bfseries,
    title={Algorithm: TITAN-guided multi-scale WSI patch selection}]
\label{alg:wsi_patch}
\noindent\textbf{Input:} CONCH features $\mathbf{F}$, patch coordinates $\{(x_i,y_i)\}$, base size $p$; hyperparameters $\tau_{\mathrm{IoU}},m$.\\
\textbf{Output:} Multi-scale ROI set $\mathcal{R}=\{(x,y,\mathrm{size},\mathrm{score},w)\}$.
\begin{enumerate}
    \item Map each patch to grid cell $(g^x_i,g^y_i)\leftarrow(\lfloor x_i/p\rfloor,\lfloor y_i/p\rfloor)$.
    \item $\{a_i\}_{i=1}^{N}\leftarrow\mathrm{TITAN.GetPatchImportance}(\mathbf{F},\mathrm{coords},p)$.
    \item Build attention map $A\leftarrow$ mean of $\{a_i\}$ per grid cell.
    \item Initialize $\mathcal{R}\leftarrow\emptyset$, occupied windows $\mathcal{W}\leftarrow\emptyset$.
    \item \textbf{for} $w\in\{8,4,2,1\}$ \textbf{do}
    \begin{enumerate}
        \item Enumerate all $w\times w$ windows whose cells exist in $A$; score each by mean attention.
        \item Sort candidates by score (descending).
        \item Accept the highest-scoring feasible window that satisfies $\mathrm{IoU}(\cdot,\mathcal{W})\le\tau_{\mathrm{IoU}}$ and margin $\ge m$; append it to $\mathcal{W}$ and the corresponding level-0 box to $\mathcal{R}$.
    \end{enumerate}
    \item Crop each ROI in $\mathcal{R}$ from the WSI at level~0; resize for the MLLM context budget.
    \item \textbf{return} $\mathcal{R}$.
\end{enumerate}
\end{tcolorbox}

\clearpage
\subsection{Prompt Templates for Reader-Study Open-Ended Scoring}
\label{sec:appendix_reader_scoring_prompts}

Open-ended answers in the pathologist reader study are graded by an independent LLM (GPT-5 mini) against curated reference answers using a fixed 1--5 rubric. The item bank includes 100 text-only questions, each comprising two sub-questions that are scored independently and therefore counted as 200 scoring units; multimodal open-ended items are scored once per question. Text-only items use a dual-sub-question system prompt so that both sub-questions of a stem are graded in one call; multimodal open-ended items use a single-question system prompt. Placeholders such as \texttt{\{stem\}}, \texttt{\{reference\_answer\}}, and \texttt{\{candidate\_answer\}} are filled at runtime.

\vspace{1em}

\begin{tcolorbox}[
    breakable,
    colback=orange!5,
    colframe=orange!70,
    boxrule=0.5pt,
    arc=2pt,
    left=6pt, right=6pt, top=6pt, bottom=6pt,
    fonttitle=\bfseries,
    title={Reader Study: Open-Ended Scoring (System Prompt, Single Item)}]
\begin{Verbatim}[breaklines=true, breakanywhere=true]
You are a strict, objective pathology exam grader. Score the candidate's
open-ended answer against the reference answer.

## Grading Procedure
1. Extract 2-4 core points from the reference answer.
2. Extract content from the candidate answer that is directly relevant to
   the current question.
3. For each core point, judge: clearly correct / missing / incorrect.
4. Assign a score from coverage; keep rationale brief (<=50 characters;
   no quotation marks), stating hits and misses.

## Core Principles
- Apply a strict standard; prefer under-scoring to over-scoring. Vague,
  generic, or formulaic answers must not receive high scores.
- Grade only content directly relevant to the question; off-topic answers
  receive no credit.
- Synonymous phrasing is acceptable, but must not broaden or narrow the
  lesion scope; incorrect additions do not earn points.
- Key diagnostic / differential points, characteristic morphology, and
  core IHC or molecular markers must be explicit and correct.
- Contradictions with the reference, lesion confusion, or mismatched
  attributions must not score above 3 even if partly correct.

## Scoring Rubric (integer 1-5)
5: All core points correct; no omissions or errors (phrasing may differ).
4: All core points correct; only minor details missing.
3: More than half of the core points correct, or one non-fatal error.
2: Only peripheral points correct; many core omissions or important errors.
1: Completely wrong, irrelevant, or essentially unanswered.

## Output
Output JSON only, with no other text:
{"score": <integer 1-5>, "rationale": "<brief justification>"}
\end{Verbatim}
\end{tcolorbox}

\vspace{1em}

\begin{tcolorbox}[
    breakable,
    colback=orange!5,
    colframe=orange!70,
    boxrule=0.5pt,
    arc=2pt,
    left=6pt, right=6pt, top=6pt, bottom=6pt,
    fonttitle=\bfseries,
    title={Reader Study: Open-Ended Scoring (System Prompt, Dual Sub-Questions)}]
\begin{Verbatim}[breaklines=true, breakanywhere=true]
You are a strict, objective pathology exam grader. Each text-only item
contains two sub-questions; score them separately.

## Grading Procedure
1. Extract 2-4 core points from the reference answer.
2. Extract content from the candidate answer that is directly relevant to
   the current question.
3. For each core point, judge: clearly correct / missing / incorrect.
4. Assign a score from coverage; keep rationale brief (<=50 characters;
   no quotation marks), stating hits and misses.

## Core Principles
- Apply a strict standard; prefer under-scoring to over-scoring. Vague,
  generic, or formulaic answers must not receive high scores.
- Grade only content directly relevant to the question; off-topic answers
  receive no credit.
- Synonymous phrasing is acceptable, but must not broaden or narrow the
  lesion scope; incorrect additions do not earn points.
- Key diagnostic / differential points, characteristic morphology, and
  core IHC or molecular markers must be explicit and correct.
- Contradictions with the reference, lesion confusion, or mismatched
  attributions must not score above 3 even if partly correct.
- Sub-question 1 and sub-question 2 must be scored independently.
- When scoring sub-question N, consider only content relevant to that
  sub-question; correct content belonging to the other sub-question
  neither adds nor subtracts points.
- Score as off-topic / low only when the core points required by that
  sub-question are entirely absent.

## Scoring Rubric (integer 1-5)
5: All core points correct; no omissions or errors (phrasing may differ).
4: All core points correct; only minor details missing.
3: More than half of the core points correct, or one non-fatal error.
2: Only peripheral points correct; many core omissions or important errors.
1: Completely wrong, irrelevant, or essentially unanswered.

## Output
Output JSON only, with no other text:
{"sub_scores": [
  {"sub_question_index": 1, "score": <integer 1-5>,
   "rationale": "<justification for sub-question 1>"},
  {"sub_question_index": 2, "score": <integer 1-5>,
   "rationale": "<justification for sub-question 2>"}
]}
\end{Verbatim}
\end{tcolorbox}

\vspace{1em}

\begin{tcolorbox}[
    breakable,
    colback=orange!5,
    colframe=orange!70,
    boxrule=0.5pt,
    arc=2pt,
    left=6pt, right=6pt, top=6pt, bottom=6pt,
    fonttitle=\bfseries,
    title={Reader Study: Open-Ended Scoring (User Prompt, Single Item)}]
\begin{Verbatim}[breaklines=true, breakanywhere=true]
[Question]
{stem}

[Reference Answer]
{reference_answer}

[Candidate Answer]
{candidate_answer}

[Grading Requirement]
Follow the grading procedure to check each core point, then assign a
score. The score must be an integer from 1-5 under a strict standard.
\end{Verbatim}
\end{tcolorbox}

\vspace{1em}

\begin{tcolorbox}[
    breakable,
    colback=orange!5,
    colframe=orange!70,
    boxrule=0.5pt,
    arc=2pt,
    left=6pt, right=6pt, top=6pt, bottom=6pt,
    fonttitle=\bfseries,
    title={Reader Study: Open-Ended Scoring (User Prompt, Dual Sub-Questions)}]
\begin{Verbatim}[breaklines=true, breakanywhere=true]
[Full Stem]
{parent_stem}

[Sub-question 1]
Stem: {sub_question_1_stem}
Reference answer: {sub_question_1_reference}

[Sub-question 2]
Stem: {sub_question_2_stem}
Reference answer: {sub_question_2_reference}

[Candidate Answer]
{candidate_answer}

[Grading Requirement]
Follow the grading procedure to score sub-question 1 and sub-question 2
separately. Each sub-question score must be an integer from 1-5,
assigned independently without cross-question penalties.
\end{Verbatim}
\end{tcolorbox}

\begin{table}[htbp]
\centering
\caption{\textbf{Overview of the comprehensive pathology evaluation benchmark.} The benchmark comprises three task categories. Public datasets use official test splits. WSI-level tasks are drawn from multi-institution private cohorts, including prospective subsets.}
\label{tab:benchmark_datasets}
\begin{tabularx}{0.75\textwidth}{@{} >{\bfseries}l X c c r @{}}
\toprule
\textbf{Task Type} & \textbf{Dataset} & \textbf{Split} & \textbf{Source} & \textbf{\# Cases} \\
\midrule
\multirow{3}{*}{Text-only}
 & NMLE 1--3 & Retrospective & Private & 142 \\
 & USMLE 1--3 & Retrospective & Private & 133 \\
 & SMU-Text 1--6 & Retrospective & Private & 1,854 \\
\midrule
\multirow{5}{*}{ROI-level Multimodal}
 & BreakHis & Retrospective & Public & 1,582 \\
 & CCRCC & Retrospective & Public & 5,635 \\
 & Chaoyang & Retrospective & Public & 2,139 \\
 & SMU-281 & Retrospective & Private & 281 \\
 & SMU-VL 1--6 & Retrospective & Private & 4,937 \\
\midrule
\multirow{10}{*}{WSI-level Multimodal}
 & CRC-T Staging & Retrospective & Private & 444 \\
 & CRC-Biopsy & Retrospective & Private & 916 \\
 & CRC-Biopsy (Prosp.) & Prospective & Private & 667 \\
 & Gastric-Grading & Retrospective & Private & 400 \\
 & Gastric-Subtyping & Retrospective & Private & 396 \\
 & Gastric-CAG (Prosp.) & Prospective & Private & 481 \\
 & Lung-NSCLC & Retrospective & Private & 599 \\
 & Breast-Grading & Retrospective & Private & 88 \\
 & Breast-TNM-T & Retrospective & Private & 89 \\
\bottomrule
\end{tabularx}
\end{table}

\begin{table}[htbp]
\centering
\small
\setlength{\tabcolsep}{5pt}
\renewcommand{\arraystretch}{1.2}
\caption{\textbf{Example questions for text-only tasks.} }
\label{tab:prompt_examples_text}
\begin{tabularx}{\textwidth}{@{} >{\raggedright\arraybackslash}p{2cm} X @{}}
\toprule
\textbf{Dataset} & \textbf{Example Question} \\
\midrule
USMLE 1 & Some undifferentiated neoplastic cells resemble their embryonic counterparts and may elaborate proteins normally expressed only in embryonic or fetal life. Tartrate-resistant acid phosphatase (TRAP) is such a protein. Which of the following describes the symptoms and signs of the tumor that would be most likely to produce TRAP? A. Enlarging moles B. Fatigue and easy bruising C. Focal weakness D. Jaundice E. Pencil-thin stools \\
USMLE 2 & A 56-year-old man is diagnosed with metastatic prostate cancer. The physician prescribes a certain drug and explains that this drug requires careful monitoring as it will first increase hormone production before therapeutically decreasing hormone production. Which of the following is the most likely mechanism of action of this drug? A. Inhibition of corticotropin-releasing hormone B. Inhibition of estrogen receptors C. Inhibition of estrogen synthesis from cholesterol D. Inhibition of follicle-\ldots \\
USMLE 3 & A 47-year-old woman, gravida 2 para 2, comes to the office after noticing a pea-sized lump in her right breast while taking a shower. Her medical history is significant for 3 pack-years of cigarette use during her 20s. She underwent infertility treatment and in vitro fertilization for both of her pregnancies. The patient has no family history of breast or ovarian cancer. A clinical breast examination confirms the presence of a firm, fixed nodule in the right breast with a small patch of overlying puckering\ldots \\
NMLE 1 & Female, 50 years old. Presents with headache and vomiting for 2 weeks, and has had 3 episodes of paroxysmal convulsions. She underwent a radical right mastectomy 5 years ago. A cranial CT scan reveals multiple low-density lesions in the left frontal lobe with significant mass effect. What is the most likely primary diagnosis? Options: A. Metastatic brain tumor B. Encephalitis C. Cerebral infarction D. Pulmonary embolism E. Meningitis. Please explicitly provide a capital letter as the final answer (A\textasciitilde{}E), followed by the rationale. \\
NMLE 2 & A 5 cm, erosive lesion is observed, which is friable and bleeds easily upon contact. The uterus and bilateral adnexa appear normal upon palpation. Tri-manual examination reveals soft parametrial tissue. If the patient is diagnosed with squamous cell carcinoma with an invasion depth of 2 mm below the basement membrane, what is the clinical stage? Options: A. IA1 B. IA2 C. IIB1 D. IIB2 E. IIIB3. Please explicitly provide a capital letter as the final answer (A\textasciitilde{}E), followed by the rationale. \\
NMLE 3 & What is the most frequently mutated gene in pancreatic cancer? Options: A. TP53 B. KRAS C. MYC D. BRAF E. RB. Please explicitly provide a capital letter as the final answer (A\textasciitilde{}E), followed by the rationale. \\
SMU-Text 1 & Male, 32 years old. Specimen: (Posterior superior iliac spine) One grayish-white to grayish-brown cord-like tissue, approx. 0.5 cm in length and 0.4 cm in diameter, entirely submitted in 1 cassette. Microscopy: Bone marrow tissue fibrosis, occupying approx. 80\% of the hematopoietic area; tissue is severely crushed with indistinct cellular architecture. Immunohistochemistry: MPO (Myeloid+), CD235a (Erythroid+), CD61 (Megakaryocytic+), CD163 (Diffuse+), Ki-67 (+, approx. 1\%). Special Stain: Reticulin stain (3+). What is the most likely pathological diagnosis? Options: A. Metastatic carcinoma-induced fibrosis B. Acute leukemia with secondary myelofibrosis C. Hyperparathyroidism D. Chronic myeloid leukemia, blast phase E. Primary myelofibrosis. \\
SMU-Text 2 & Female, 57 years old. Specimen: (Brain tumor) A pile of grayish-white to grayish-brown fragmented tissues, measuring approx. 7.0*6.0*1.5 cm. Microscopy: Tumor cells are arranged in glandular or cribriform patterns. Cells are large with abundant eosinophilic cytoplasm, and the nuclei are large, hyperchromatic, with obvious atypia. Immunohistochemistry: CK (+), P16 (Diffuse+), ER (Moderate-Strong+, approx. 90\%), PR (Moderate-Strong+, approx. 65\%), P53 (Strong+, approx. 95\%, mutant pattern), PAX-8 (+), Ki-67 (+, hot spots approx. 80\%). What is the most likely pathological diagnosis? Options: A. Glioblastoma multiforme B. Brain abscess C. Primary central nervous system lymphoma D. Metastatic lung cancer E. Metastatic endometrial carcinoma. \\
SMU-Text 3 & Female, 31 years old. Specimen: (Left adrenal gland) One irregular grayish-yellow tissue mass, measuring approx. 6.0*4.0*3.0 cm, showing a nodule. Microscopy: Tumor cells are arranged in streaming or fascicular patterns. Cells are spindle-shaped or wavy, with scattered aggregates of numerous ganglion cells. What is the most likely pathological diagnosis? Options: A. Adrenocortical adenoma B. Ganglioneuroma C. Neuroblastoma D. Non-functioning adrenocortical carcinoma E. Pheochromocytoma. \\
\bottomrule
\end{tabularx}
\end{table}

\begin{table}[htbp]
\centering
\small
\setlength{\tabcolsep}{5pt}
\renewcommand{\arraystretch}{1.2}
\caption{\textbf{Example questions for ROI-level multimodal tasks.} }
\label{tab:prompt_examples_roi}
\begin{tabularx}{\textwidth}{@{} >{\raggedright\arraybackslash}p{3cm} X @{}}
\toprule
\textbf{Dataset} & \textbf{Example Question} \\
\midrule
BreakHis & Analyze this breast histopathology image and classify the tumor type: (A) Benign tumor., (B) Malignant tumor. \\
CCRCC & Analyze this H\&E-stained renal tissue tile. Which description best matches the image? (A) Renal cancer., (B) Normal renal., (C) Stromal, including smooth muscle, fibrous stroma and blood vessels., (D) Red blood cells. \\
Chaoyang & Classify this colon tissue patch from Chaoyang hospital into one of these diagnostic categories based on histological features: (A) Normal tissue., (B) Serrated., (C) Adenocarcinoma., (D) Adenoma. \\
SMU-281 Breast \& Chest & Female, 31 years old, presented with ``discovery of a right breast mass for 1 month''. The right nipple is slightly retracted. A 2.5 cm x 2.0 cm mass is palpable in the upper outer quadrant of the right breast; it is firm, has indistinct borders with surrounding tissue, exhibits poor mobility, and is tender to touch. What is the most likely diagnosis? A. Tuberculosis B. Breast abscess C. Granulomatous lobular mastitis D. Mammary duct ectasia E. Breast cancer \\
SMU-281 GU \& Skin \& H \& N  & Female, 40 years old, presented with a right submandibular mass discovered 2 years ago, with no significant self-reported enlargement. Histology is shown in the image. What is the most likely diagnosis? A. Pleomorphic adenoma B. Basal cell adenoma C. Myoepithelioma D. Mucoepidermoid carcinoma E. Myoepithelial carcinoma \\
GI \& Thyroid \& Other & Male, 35 years old. Presents with fever and cervical lymphadenopathy for 2 weeks. What is the most likely diagnosis? A. ALK-positive anaplastic large cell lymphoma B. Kikuchi-Fujimoto disease C. Diffuse large B-cell lymphoma D. Cat scratch disease \\
SMU-VL 1 & Male, 32 years old. Specimen: (Posterior superior iliac spine) One grayish-white to grayish-brown cord-like tissue, approx. 0.5 cm in length and 0.4 cm in diameter, entirely submitted in 1 cassette. Immunohistochemistry (01\#): MPO (Myeloid+), CD235a (Erythroid+), CD61 (Megakaryocytic+), CD34 (Scattered+), CD117 (Rare+), CD68 (Focal+), CD163 (Diffuse+), Ki-67 (+, approx. 1\%). Special Stain: Reticulin stain (3+). Please select the most likely pathological diagnosis from the following options (A\textasciitilde{}E). A. Metastatic carcinoma-induced fibrosis B. Acute leukemia with secondary myelofibrosis C. Hyperparathyroidism D. Chronic myeloid leukemia, blast phase E. Primary myelofibrosis \\
SMU-VL 2 & Female, 57 years old. Specimen: (Brain tumor) Fragmented tissues, approx. 7.0*6.0*1.5 cm. Cut surface is grayish-white to grayish-brown, solid, and soft. Immunohistochemistry: CK (+), CK7 (Focal+), P16 (Diffuse+), ER (Moderate-Strong+, approx. 90\%), PR (Moderate-Strong+, approx. 65\%), P53 (Strong+, approx. 95\%, mutant pattern), PAX-8 (+), $\beta$-catenin (Membrane+), Ki-67 (+, hot spots approx. 80\%). Please select the most likely pathological diagnosis from the following options (A\textasciitilde{}E). A. Metastatic endometrial carcinoma B. Primary central nervous system lymphoma C. Glioblastoma multiforme D. Brain abscess E. Metastatic lung cancer \\
SMU-VL 3 & Female, 31 years old. Specimen: (Left adrenal gland) Irregular grayish-yellow mass, measuring approx. 6.0*4.0*3.0 cm. A cut reveals a nodule with a max diameter of approx. 4.0 cm; cut surface is grayish-white, solid, soft, with an intact capsule. Please select the most likely pathological diagnosis from the following options (A\textasciitilde{}E). A. Adrenocortical adenoma B. Non-functioning adrenocortical carcinoma C. Pheochromocytoma D. Neuroblastoma E. Ganglioneuroma \\
SMU-VL 4 & Female, 51 years old. Specimen: (Left ankle lesion) Irregular grayish-white to grayish-red tissue piece, approx. 2.0*1.8*0.7 cm. Cut surface is grayish-white, solid, medium consistency. Please select the most likely pathological diagnosis from the following options (A\textasciitilde{}E). A. Synovial sarcoma B. Schwannoma C. Angioleiomyoma D. Leiomyosarcoma E. Tenosynovial giant cell tumor \\
SMU-VL 5 & Female, 28 years old. Specimen: (Posterior superior iliac spine) Two grayish-brown cord-like tissues, approx. 0.9-1.6 cm in length and 0.2 cm in diameter. Please select the most likely pathological diagnosis from the following options (A\textasciitilde{}E). A. Metastatic neuroblastoma B. Megaloblastic anemia C. Myelodysplastic syndrome D. Acute myeloid leukemia E. Aplastic anemia \\
SMU-VL 6 & Male, 56 years old. Specimen: (Gastric antrum) One grayish-white tissue piece, approx. 0.3 cm. (Esophagus, cardia) Five grayish-white tissue pieces, approx. 0.2-0.5 cm. Total 2 cassettes. Please select the most likely pathological diagnosis from the following options (A\textasciitilde{}E). A. Gastrointestinal stromal tumor B. Malignant lymphoma C. Moderately to poorly differentiated adenocarcinoma D. Reactive gastropathy E. Signet ring cell carcinoma \\
\bottomrule
\end{tabularx}
\end{table}

\begin{table}[htbp]
\centering
\setlength{\tabcolsep}{5pt}
\renewcommand{\arraystretch}{1.2}
\caption{\textbf{Example questions for WSI-level multimodal tasks.}}
\label{tab:prompt_examples_wsi}
\begin{tabularx}{\textwidth}{@{} >{\raggedright\arraybackslash}p{3.2cm} X @{}}
\toprule
\textbf{Dataset} & \textbf{Example Question} \\
\midrule
CRC-T Staging & Based on the pathological images, classify the T stage of this colon cancer: A. T1, B. T2, C. T3, D. T4 \\
CRC-Biopsy & Based on the given pathological images from the colorectal region, what is the most likely diagnosis? A. Adenomatous lesion, B. Invasive adenocarcinoma, C. Benign non-neoplastic lesion, D. High-grade dysplasia/early-stage cancer \\
CRC-Biopsy (Prosp.) & Based on the given pathological images from the colorectal region, what is the most likely diagnosis? A. Adenomatous lesion, B. Invasive adenocarcinoma, C. Benign non-neoplastic lesion, D. High-grade dysplasia/early-stage cancer \\
Gastric-Grading & Based on the histopathological images, which grade does this gastric cancer belong to? A. Grade 1 (high differentiation） B. Grade 2 (moderate differentiation) C. Grade 3 (poor differentiation) \\
Gastric-Subtyping & Based on the histopathological images, which subtype does this gastric cancer belong to? A. Signet Ring Cell Carcinoma of the Stomach B.Tubular Stomach Adenocarcinoma C. Stomach Adenocarcinoma \\
Gastric-CAG (Prosp.) & Classify this HE-stained gastric biopsy image as A or B based on the presence or absence of chronic atrophic gastritis. A. chronic atrophic gastritis, B. non-atrophic gastritis or normal mucosa \\
Lung-NSCLC & Considering all these pathology image patches from the same slide, which diagnosis is more consistent with the overall morphology? A. Lung adenocarcinoma (LUAD) B. Lung squamous cell carcinoma (LUSC) \\
Breast-Grading & Based on the histopathological images, which grade does this breast cancer belong to? A. Grade 1 (high differentiation）B. Grade 2 (moderately differentiated) C. Grade 3 (poorly differentiated) \\
Breast-TNM-T & Based on the pathological images, classify the T stage of this breast cancer: A. T1 B. T2 C. T3 \\
\bottomrule
\end{tabularx}
\end{table}

\begin{table}[htbp]
\centering
\small
\caption{\textbf{Text-only QA performance on 12 datasets.} 
 Best performing model is \textbf{bolded} and second-best is \underline{underlined}. The 95\% CI is included in parentheses.}
\label{tab:bench:text_only}
\begin{tabular}{lccc}
\toprule
\textbf{Method} & \textbf{USMLE 1} & \textbf{USMLE 2} & \textbf{USMLE 3}\\
\midrule
Grok 4.1 Fast & 0.844 (0.734, 0.955) & 0.816 (0.704, 0.929) & 0.846 (0.728, 0.965)\\
GPT-4.1 mini & 0.867 (0.763, 0.970) & \underline{0.959 (0.918, 1.000)} & 0.872 (0.762, 0.982)\\
GPT-5 mini & \underline{0.933 (0.867, 1.000)} & \textbf{1.000 (1.000, 1.000)} & \underline{0.923 (0.846, 1.000)}\\
Qwen3.7 Plus & \underline{0.933 (0.867, 1.000)} & \underline{0.959 (0.918, 1.000)} & 0.872 (0.762, 0.982)\\
PathPocket (Ours) & \textbf{0.956 (0.911, 1.000)} & \textbf{1.000 (1.000, 1.000)} & \textbf{1.000 (1.000, 1.000)}\\
\midrule
\textbf{Method} & \textbf{NMLE 1} & \textbf{NMLE 2} & \textbf{NMLE 3}\\
\midrule
Grok 4.1 Fast & 0.829 (0.697, 0.960) & 0.898 (0.810, 0.986) & 0.862 (0.771, 0.954)\\
GPT-4.1 mini & 0.800 (0.661, 0.939) & 0.857 (0.756, 0.959) & 0.879 (0.793, 0.966)\\
GPT-5 mini & \underline{0.857 (0.735, 0.979)} & 0.898 (0.810, 0.986) & \underline{0.897 (0.816, 0.977)}\\
Qwen3.7 Plus & \underline{0.857 (0.735, 0.979)} & \underline{0.918 (0.839, 0.998)} & 0.862 (0.771, 0.954)\\
PathPocket (Ours) & \textbf{0.914 (0.829, 1.000)} & \textbf{0.939 (0.878, 1.000)} & \textbf{0.914 (0.839, 0.988)}\\
\midrule
\textbf{Method} & \textbf{SMU-Text 1} & \textbf{SMU-Text 2} & \textbf{SMU-Text 3}\\
\midrule
Grok 4.1 Fast & 0.733 (0.685, 0.781) & 0.708 (0.661, 0.754) & 0.660 (0.614, 0.706)\\
GPT-4.1 mini & 0.798 (0.754, 0.841) & 0.791 (0.749, 0.832) & 0.776 (0.736, 0.816)\\
GPT-5 mini & 0.883 (0.848, 0.918) & 0.887 (0.855, 0.920) & 0.834 (0.798, 0.870)\\
Qwen3.7 Plus & \underline{0.902 (0.869, 0.934)} & \underline{0.895 (0.864, 0.927)} & \underline{0.877 (0.845, 0.909)}\\
PathPocket (Ours) & \textbf{0.923 (0.894, 0.952)} & \textbf{0.909 (0.880, 0.938)} & \textbf{0.913 (0.886, 0.940)}\\
\midrule
\textbf{Method} & \textbf{SMU-Text 4} & \textbf{SMU-Text 5} & \textbf{SMU-Text 6}\\
\midrule
Grok 4.1 Fast & 0.482 (0.426, 0.538) & 0.471 (0.421, 0.521) & 0.602 (0.531, 0.673)\\
GPT-4.1 mini & 0.689 (0.636, 0.741) & 0.639 (0.591, 0.688) & 0.747 (0.684, 0.810)\\
GPT-5 mini & 0.800 (0.755, 0.845) & \underline{0.889 (0.858, 0.921)} & 0.903 (0.860, 0.946)\\
Qwen3.7 Plus & \underline{0.856 (0.816, 0.895)} & 0.876 (0.843, 0.910) & \underline{0.919 (0.880, 0.959)}\\
PathPocket (Ours) & \textbf{0.931 (0.903, 0.960)} & \textbf{0.932 (0.906, 0.957)} & \textbf{0.946 (0.914, 0.979)}\\
\bottomrule
\end{tabular}
\end{table}

\begin{table}[htbp]
\centering
\small
\caption{\textbf{ROI-level VQA performance on 3 public datasets (BreakHis, CCRCC, Chaoyang).} Best performing model is \textbf{bolded} and second-best is \underline{underlined}. The 95\% CI is included in parentheses. For CONCH/PLIP, (O) denotes option-only zero-shot prompts and (Q+O) denotes question+option prompts (image--text cosine matching).}
\label{tab:bench:roi_public}
\begin{tabular}{lccc}
\toprule
\textbf{Method} & \textbf{BreakHis} & \textbf{CCRCC} & \textbf{Chaoyang}\\
\midrule
Grok 4.1 Fast & 0.524 (0.493, 0.555) & 0.329 (0.308, 0.349) & 0.426 (0.395, 0.457)\\
GPT-4.1 mini & 0.581 (0.550, 0.612) & 0.472 (0.450, 0.494) & 0.362 (0.332, 0.392)\\
GPT-5 mini & 0.544 (0.513, 0.575) & \underline{0.633 (0.612, 0.654)} & 0.450 (0.419, 0.481)\\
Qwen3.7 Plus & \underline{0.716 (0.688, 0.744)} & 0.512 (0.491, 0.534) & \underline{0.575 (0.544, 0.606)}\\
LLaVA-Med & 0.376 (0.346, 0.406) & 0.353 (0.332, 0.374) & 0.332 (0.303, 0.361)\\
Quilt-Llava & 0.580 (0.549, 0.611) & 0.391 (0.369, 0.412) & 0.306 (0.277, 0.335)\\

CONCH (O) & 0.687 (0.658, 0.716) & 0.377 (0.355, 0.398) & 0.556 (0.525, 0.587)\\
CONCH (Q+O) & 0.620 (0.590, 0.650) & 0.444 (0.422, 0.465) & 0.182 (0.158, 0.206)\\
PLIP (O) & 0.510 (0.479, 0.541) & 0.512 (0.490, 0.534) & 0.291 (0.263, 0.319)\\
PLIP (Q+O) & 0.672 (0.643, 0.701) & 0.511 (0.489, 0.533) & 0.322 (0.293, 0.351)\\
PathPocket (Ours) & \textbf{0.743 (0.716, 0.770)} & \textbf{0.695 (0.674, 0.715)} & \textbf{0.670 (0.641, 0.699)}\\
\bottomrule
\end{tabular}
\end{table}

\begin{table}[htbp]
\centering
\small
\caption{\textbf{ROI-level VQA performance on 3 private datasets (SMU-281).} Best performing model is \textbf{bolded} and second-best is \underline{underlined}. The 95\% CI is included in parentheses. For CONCH/PLIP, (O) denotes option-only zero-shot prompts and (Q+O) denotes question+option prompts.}
\label{tab:bench:roi_smu281}
\begin{tabular}{lccc}
\toprule
\textbf{Method} & \textbf{Breast \& Chest} & \textbf{GU \& Skin \& H\&N} & \textbf{GI \& Thyroid \& Other}\\
\midrule
Grok 4.1 Fast & 0.485 (0.363, 0.607) & 0.487 (0.376, 0.599) & 0.519 (0.432, 0.606)\\
GPT-4.1 mini & 0.529 (0.408, 0.651) & 0.662 (0.557, 0.768) & 0.618 (0.534, 0.703)\\
GPT-5 mini & \underline{0.647 (0.531, 0.764)} & 0.688 (0.584, 0.791) & 0.672 (0.590, 0.753)\\
Qwen3.7 Plus & 0.603 (0.484, 0.722) & \underline{0.775 (0.681, 0.869)} & \underline{0.733 (0.656, 0.810)}\\
LLaVA-Med & 0.250 (0.144, 0.356) & 0.212 (0.121, 0.304) & 0.145 (0.084, 0.206)\\
Quilt-Llava & 0.324 (0.209, 0.438) & 0.163 (0.080, 0.245) & 0.267 (0.190, 0.344)\\
CONCH (O) & 0.397 (0.278, 0.516) & 0.388 (0.278, 0.497) & 0.427 (0.342, 0.513)\\
CONCH (Q+O) & 0.368 (0.250, 0.485) & 0.463 (0.351, 0.574) & 0.489 (0.402, 0.575)\\
PLIP (O) & 0.426 (0.306, 0.547) & 0.463 (0.351, 0.574) & 0.397 (0.312, 0.482)\\
PLIP (Q+O) & 0.382 (0.264, 0.501) & 0.362 (0.255, 0.470) & 0.321 (0.240, 0.402)\\
PathPocket (Ours) & \textbf{0.750 (0.644, 0.856)} & \textbf{0.800 (0.710, 0.890)} & \textbf{0.748 (0.673, 0.823)}\\
\bottomrule
\end{tabular}
\end{table}

\begin{table}[htbp]
\centering
\small
\caption{\textbf{ROI-level VQA performance on 6 private datasets (SMU-VL 1--6).} 
Best performing model is \textbf{bolded} and second-best is \underline{underlined}. The 95\% CI is included in parentheses. For CONCH/PLIP, (O) denotes option-only zero-shot prompts and (Q+O) denotes question+option prompts.}
\label{tab:bench:roi_smuvl}
\begin{tabular}{lccc}
\toprule
\textbf{Method} & \textbf{SMU-VL 1} & \textbf{SMU-VL 2} & \textbf{SMU-VL 3}\\
\midrule
Grok 4.1 Fast & 0.414 (0.377, 0.451) & 0.386 (0.351, 0.421) & 0.330 (0.299, 0.360)\\
GPT-4.1 mini & \underline{0.684 (0.649, 0.719)} & 0.637 (0.602, 0.672) & 0.641 (0.609, 0.672)\\
GPT-5 mini & 0.645 (0.609, 0.682) & \underline{0.676 (0.642, 0.710)} & 0.631 (0.599, 0.662)\\
Qwen3.7 Plus & 0.671 (0.635, 0.706) & 0.625 (0.590, 0.660) & 0.611 (0.579, 0.643)\\
LLaVA-Med & 0.237 (0.205, 0.270) & 0.201 (0.172, 0.229) & 0.233 (0.205, 0.260)\\
Quilt-Llava & 0.257 (0.224, 0.290) & 0.229 (0.199, 0.259) & 0.248 (0.220, 0.276)\\
CONCH (O) & 0.266 (0.232, 0.299) & 0.266 (0.234, 0.298) & 0.217 (0.190, 0.244)\\
CONCH (Q+O) & 0.304 (0.269, 0.339) & 0.317 (0.283, 0.351) & 0.313 (0.283, 0.343)\\
PLIP (O) & 0.625 (0.588, 0.661) & 0.615 (0.580, 0.650) & \underline{0.645 (0.614, 0.676)}\\
PLIP (Q+O) & 0.221 (0.190, 0.252) & 0.215 (0.186, 0.245) & 0.209 (0.183, 0.236)\\
PathPocket (Ours) & \textbf{0.797 (0.766, 0.827)} & \textbf{0.795 (0.766, 0.825)} & \textbf{0.788 (0.762, 0.815)}\\
\midrule
\textbf{Method} & \textbf{SMU-VL 4} & \textbf{SMU-VL 5} & \textbf{SMU-VL 6}\\
\midrule
Grok 4.1 Fast & 0.317 (0.286, 0.348) & 0.319 (0.289, 0.350) & 0.333 (0.302, 0.365)\\
GPT-4.1 mini & 0.562 (0.529, 0.596) & 0.618 (0.587, 0.650) & 0.542 (0.509, 0.576)\\
GPT-5 mini & \underline{0.754 (0.725, 0.783)} & \underline{0.753 (0.725, 0.781)} & \underline{0.784 (0.756, 0.812)}\\
Qwen3.7 Plus & 0.599 (0.566, 0.632) & 0.640 (0.608, 0.671) & 0.591 (0.558, 0.624)\\
LLaVA-Med & 0.221 (0.193, 0.249) & 0.196 (0.170, 0.222) & 0.194 (0.167, 0.220)\\
Quilt-Llava & 0.204 (0.177, 0.231) & 0.245 (0.217, 0.273) & 0.246 (0.217, 0.275)\\
CONCH (O) & 0.197 (0.170, 0.224) & 0.187 (0.161, 0.212) & 0.194 (0.167, 0.220)\\
CONCH (Q+O) & 0.330 (0.298, 0.361) & 0.321 (0.291, 0.352) & 0.336 (0.304, 0.367)\\
PLIP (O) & 0.536 (0.503, 0.570) & 0.461 (0.428, 0.493) & 0.534 (0.501, 0.568)\\
PLIP (Q+O) & 0.212 (0.185, 0.240) & 0.189 (0.163, 0.215) & 0.180 (0.154, 0.206)\\
PathPocket (Ours) & \textbf{0.823 (0.797, 0.848)} & \textbf{0.844 (0.821, 0.868)} & \textbf{0.847 (0.823, 0.871)}\\
\bottomrule
\end{tabular}
\end{table}

\begin{table}[htbp]
\centering
\caption{\textbf{WSI-level VQA performance on 9 private cohorts.} Binomial accuracy with Wilson 95\% CI\cite{wilson1927probable}. For CONCH/PLIP, (O) denotes option-only zero-shot prompts and (Q+O) denotes question+option prompts (image--text cosine matching). Best performing model is \textbf{bolded} and second-best is \underline{underlined}. The 95\% CI is included in parentheses.}
\label{tab:bench:wsi_private}
\begin{tabular}{lccc}
\toprule
\textbf{Method} & \textbf{CRC-T Staging} & \textbf{CRC-Biopsy} & \textbf{CRC-Biopsy (Prosp.)}\\
\midrule
Grok 4.1 Fast & 0.390 (0.345, 0.436) & 0.334 (0.304, 0.365) & 0.357 (0.321, 0.394)\\
GPT-4.1 mini & 0.559 (0.512, 0.604) & 0.395 (0.364, 0.427) & 0.292 (0.259, 0.328)\\
GPT-5 mini & 0.556 (0.510, 0.602) & \underline{0.549 (0.517, 0.581)} & 0.628 (0.591, 0.664)\\
Qwen3.7 Plus & 0.500 (0.454, 0.546) & 0.519 (0.486, 0.551) & \underline{0.651 (0.614, 0.686)}\\
LLaVA-Med & \underline{0.709 (0.666, 0.750)} & 0.385 (0.354, 0.417) & 0.381 (0.345, 0.418)\\
Quilt-Llava & 0.541 (0.494, 0.586) & 0.385 (0.354, 0.417) & 0.385 (0.349, 0.423)\\
CONCH (O) & 0.579 (0.532, 0.624) & 0.233 (0.206, 0.261) & 0.471 (0.433, 0.509)\\
CONCH (Q+O) & 0.079 (0.057, 0.108) & 0.439 (0.407, 0.471) & 0.627 (0.589, 0.663)\\
PLIP (O) & 0.099 (0.075, 0.130) & 0.098 (0.081, 0.119) & 0.298 (0.265, 0.334)\\
PLIP (Q+O) & 0.115 (0.088, 0.148) & 0.129 (0.109, 0.152) & 0.049 (0.035, 0.069)\\
PathPocket (Ours) & \textbf{0.721 (0.677, 0.760)} & \textbf{0.604 (0.572, 0.635)} & \textbf{0.681 (0.644, 0.715)}\\
\midrule
\textbf{Method} & \textbf{Gastric-Grading} & \textbf{Gastric-Subtyping} & \textbf{Gastric-CAG (Prosp.)}\\
\midrule
Grok 4.1 Fast & 0.495 (0.446, 0.544) & 0.475 (0.426, 0.524) & 0.497 (0.452, 0.541)\\
GPT-4.1 mini & 0.492 (0.444, 0.541) & 0.467 (0.419, 0.516) & 0.532 (0.488, 0.576)\\
GPT-5 mini & 0.472 (0.424, 0.521) & \underline{0.480 (0.431, 0.529)} & \underline{0.555 (0.510, 0.599)}\\
Qwen3.7 Plus & 0.458 (0.409, 0.506) & 0.472 (0.424, 0.521) & 0.553 (0.508, 0.597)\\
LLaVA-Med & 0.210 (0.173, 0.253) & 0.422 (0.374, 0.471) & 0.486 (0.442, 0.531)\\
Quilt-Llava & 0.163 (0.130, 0.202) & 0.407 (0.359, 0.456) & 0.516 (0.471, 0.560)\\
CONCH (O) & \underline{0.600 (0.551, 0.647)} & 0.351 (0.306, 0.399) & 0.505 (0.461, 0.550)\\
CONCH (Q+O) & 0.205 (0.168, 0.247) & 0.394 (0.347, 0.443) & 0.538 (0.494, 0.583)\\
PLIP (O) & 0.177 (0.143, 0.218) & 0.384 (0.337, 0.433) & 0.497 (0.452, 0.541)\\
PLIP (Q+O) & 0.035 (0.021, 0.058) & 0.338 (0.294, 0.386) & 0.497 (0.452, 0.541)\\
PathPocket (Ours) & \textbf{0.637 (0.589, 0.683)} & \textbf{0.485 (0.436, 0.534)} & \textbf{0.586 (0.542, 0.629)}\\
\midrule
\textbf{Method} & \textbf{Lung-NSCLC} & \textbf{Breast-Grading} & \textbf{Breast-TNM-T}\\
\midrule
Grok 4.1 Fast & 0.576 (0.536, 0.615) & \underline{0.659 (0.555, 0.750)} & 0.157 (0.096, 0.247)\\
GPT-4.1 mini & 0.634 (0.595, 0.672) & 0.648 (0.544, 0.739) & 0.281 (0.198, 0.382)\\
GPT-5 mini & 0.609 (0.570, 0.648) & 0.591 (0.486, 0.688) & 0.348 (0.257, 0.452)\\
Qwen3.7 Plus & 0.626 (0.587, 0.664) & 0.534 (0.431, 0.635) & \underline{0.494 (0.393, 0.596)}\\
LLaVA-Med & 0.503 (0.463, 0.542) & 0.580 (0.475, 0.677) & 0.011 (0.002, 0.061)\\
Quilt-Llava & 0.516 (0.476, 0.556) & 0.307 (0.220, 0.410) & 0.472 (0.372, 0.575)\\
CONCH (O) & 0.624 (0.585, 0.662) & 0.307 (0.220, 0.410) & 0.101 (0.054, 0.181)\\
CONCH (Q+O) & 0.649 (0.610, 0.687) & 0.307 (0.220, 0.410) & 0.393 (0.298, 0.497)\\
PLIP (O) & \textbf{0.666 (0.627, 0.703)} & 0.375 (0.281, 0.479) & 0.393 (0.298, 0.497)\\
PLIP (Q+O) & 0.646 (0.607, 0.683) & 0.398 (0.302, 0.502) & \textbf{0.528 (0.425, 0.628)}\\
PathPocket (Ours) & \underline{0.664 (0.626, 0.701)} & \textbf{0.670 (0.567, 0.760)} & \underline{0.494 (0.393, 0.596)}\\
\bottomrule
\end{tabular}
\end{table}

\clearpage

\subsection{Ablation of Evidence Retrieval on SMU-281}
\label{sec:appendix_ablation}

To quantify the contribution of PathPocket's hypergraph evidence relative to simpler retrieval strategies, we conduct an ablation on the private SMU-281 ROI benchmark under the same three anatomically exclusive organ strata used in the main ROI evaluation (Breast \& Resp \& Mediastinum; Gyn \& GU \& Bone \& Soft Tissue \& H\&N; GI \& HPB \& Thyroid \& Heme \& CNS \& Endo \& CV \& Other), together with the overall pool ($n=281$). All variants share the same multimodal backbone (Qwen3.7 Plus) and answer format; only the evidence-retrieval pathway is changed:
\begin{itemize}
    \item \textbf{Qwen3.7 Plus}: the backbone multimodal model without external retrieval.
    \item \textbf{Naive-RAG}: retrieves similar text chunks directly from the query and conditions generation on those chunks as evidence, without structured hypergraph traversal.
    \item \textbf{Graph-RAG}: retains graph-style retrieval but removes multi-entity relational evidence from the hypergraph pathway (i.e., ablated multi-entity relation evidence).
    \item \textbf{PathPocket}: the full system with multimodal hypergraph retrieval, multi-entity relations, and agentic evidence filtering.
\end{itemize}

As summarized in Appendix Figure~\ref{fig:ablation_smu281}, PathPocket attains the best overall accuracy (76.3\%), outperforming Graph-RAG (73.1\%), Qwen3.7 Plus (71.3\%), and Naive-RAG (65.9\%). Across organ strata, PathPocket leads on Breast \& Resp \& Mediastinum (75.0\% vs.\ 70.6\% / 63.2\% / 60.3\% for Graph-RAG, Naive-RAG, and Qwen3.7 Plus) and on GI \& HPB \& Thyroid \& Heme \& CNS \& Endo \& CV \& Other (74.8\% vs.\ 73.3\% / 69.5\% / 61.8\% for Qwen3.7 Plus, Graph-RAG, and Naive-RAG). On Gyn \& GU \& Bone \& Soft Tissue \& H\&N, Graph-RAG is marginally highest (81.2\%), with PathPocket close behind (80.0\%), followed by Qwen3.7 Plus (77.5\%) and Naive-RAG (75.0\%). Notably, Naive-RAG underperforms the no-retrieval backbone overall, suggesting that unstructured chunk retrieval can inject noisy evidence, whereas Graph-RAG improves over Naive-RAG but remains below PathPocket in the aggregate. These results support that PathPocket's full hypergraph pathway with multi-entity relational evidence provides the most reliable gains across anatomical subsets.

\clearpage

\begin{figure}[H]
    \centering
    \includegraphics[width=0.75\textwidth]{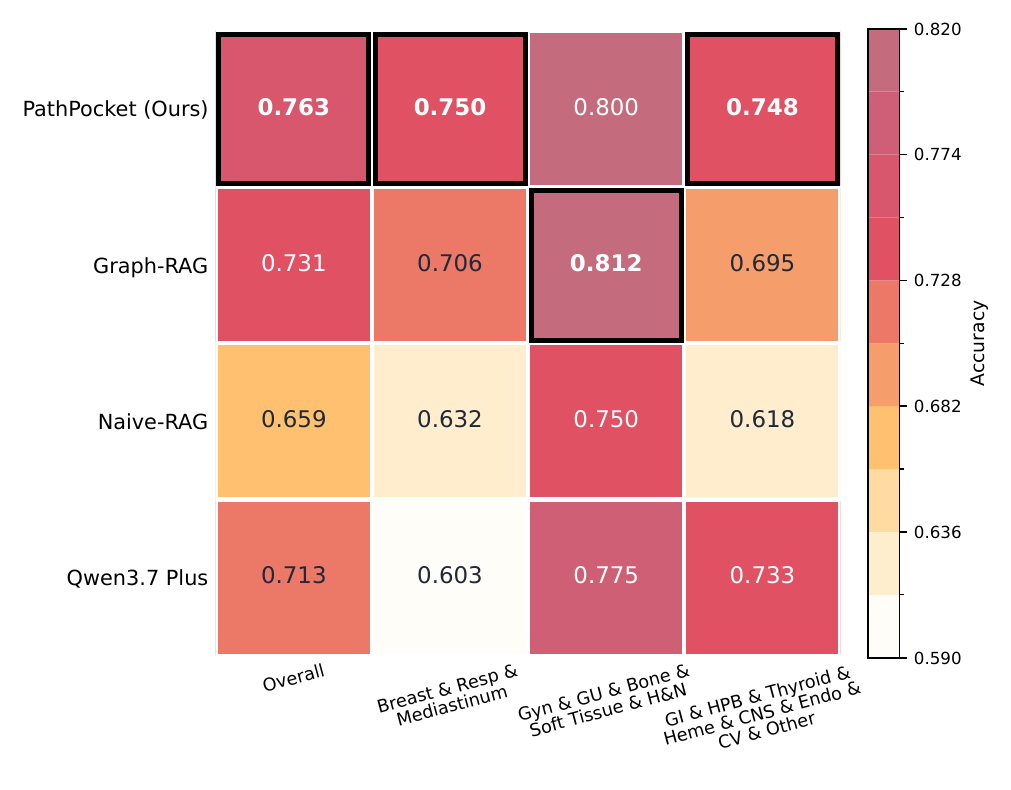}
    \caption{\textbf{Ablation of evidence retrieval on SMU-281.}
    Accuracy heatmap comparing PathPocket with Graph-RAG, Naive-RAG, and the Qwen3.7 Plus backbone across Overall and three organ strata (Breast \& Resp \& Mediastinum; Gyn \& GU \& Bone \& Soft Tissue \& H\&N; GI \& HPB \& Thyroid \& Heme \& CNS \& Endo \& CV \& Other; $n=281$).
    \textbf{Naive-RAG} retrieves similar text chunks directly from the question as evidence.
    \textbf{Graph-RAG} removes multi-entity relational evidence from the hypergraph retrieval pathway.
    Black outlines mark the best method in each column.}
    \label{fig:ablation_smu281}
\end{figure}




\begin{figure}[H]
    \centering
    \includegraphics[width=0.92\textwidth]{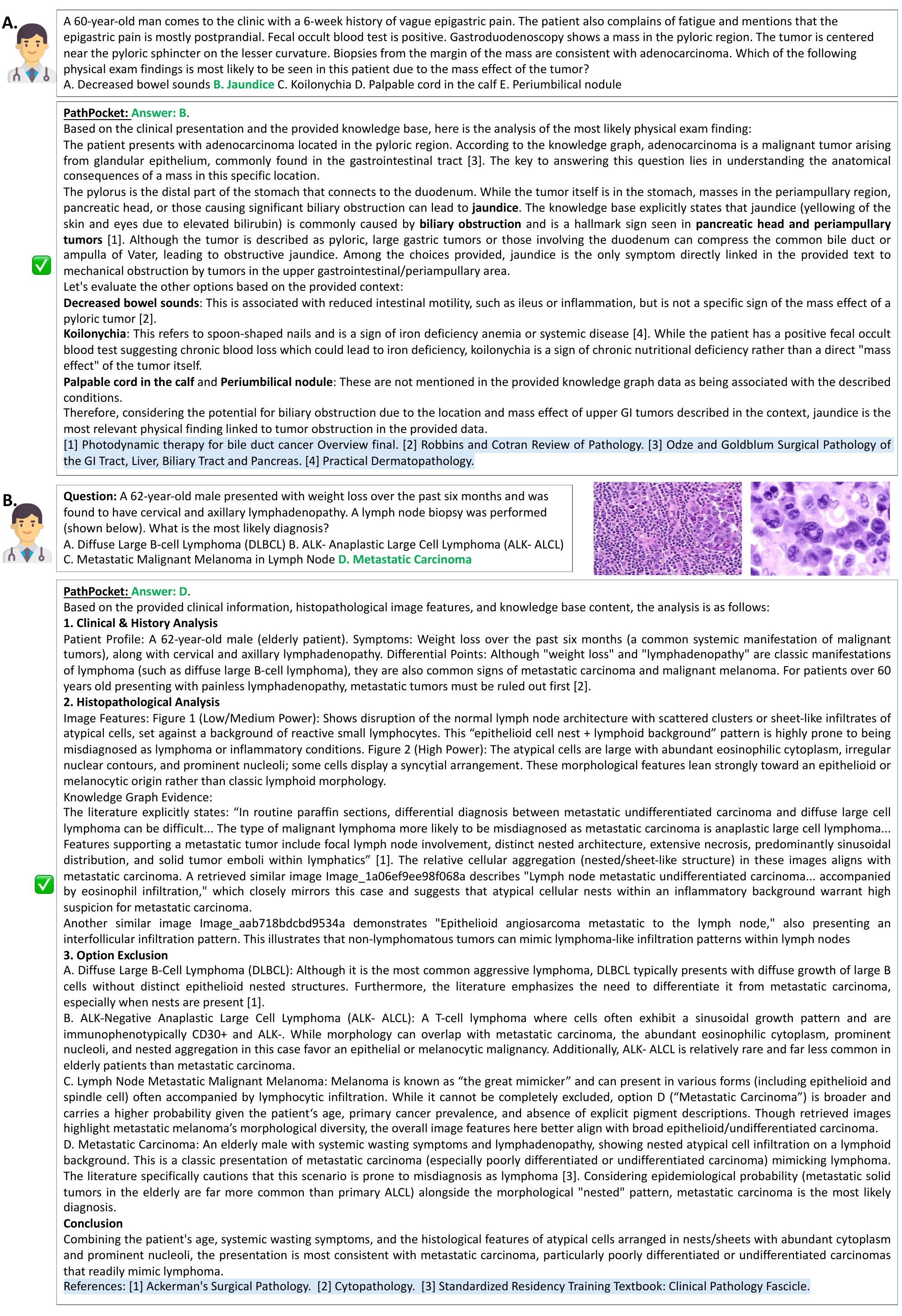}
\end{figure}

\begin{figure}[H]
    \centering
    \caption{\textbf{Full PathPocket qualitative responses corresponding to the main-text case studies.}
    Unabridged PathPocket outputs for the two cases summarized in Figure~\ref{fig:qualitative_examples}.
    \textbf{Top:} text-only pyloric-mass case in which PathPocket selects Jaundice (B) with anatomy-grounded reasoning and literature citations.
    \textbf{Bottom:} multimodal lymph-node ROI case in which PathPocket diagnoses Metastatic Carcinoma (D), systematically excluding lymphoma differentials with cited surgical pathology evidence.
    Relative to the main-text figure, this panel retains the complete stepwise rationale rather than a condensed summary.}
    \label{fig:qualitative_full}
\end{figure}

\clearpage



\begin{figure}[H]
    \centering
    \includegraphics[width=0.92\textwidth]{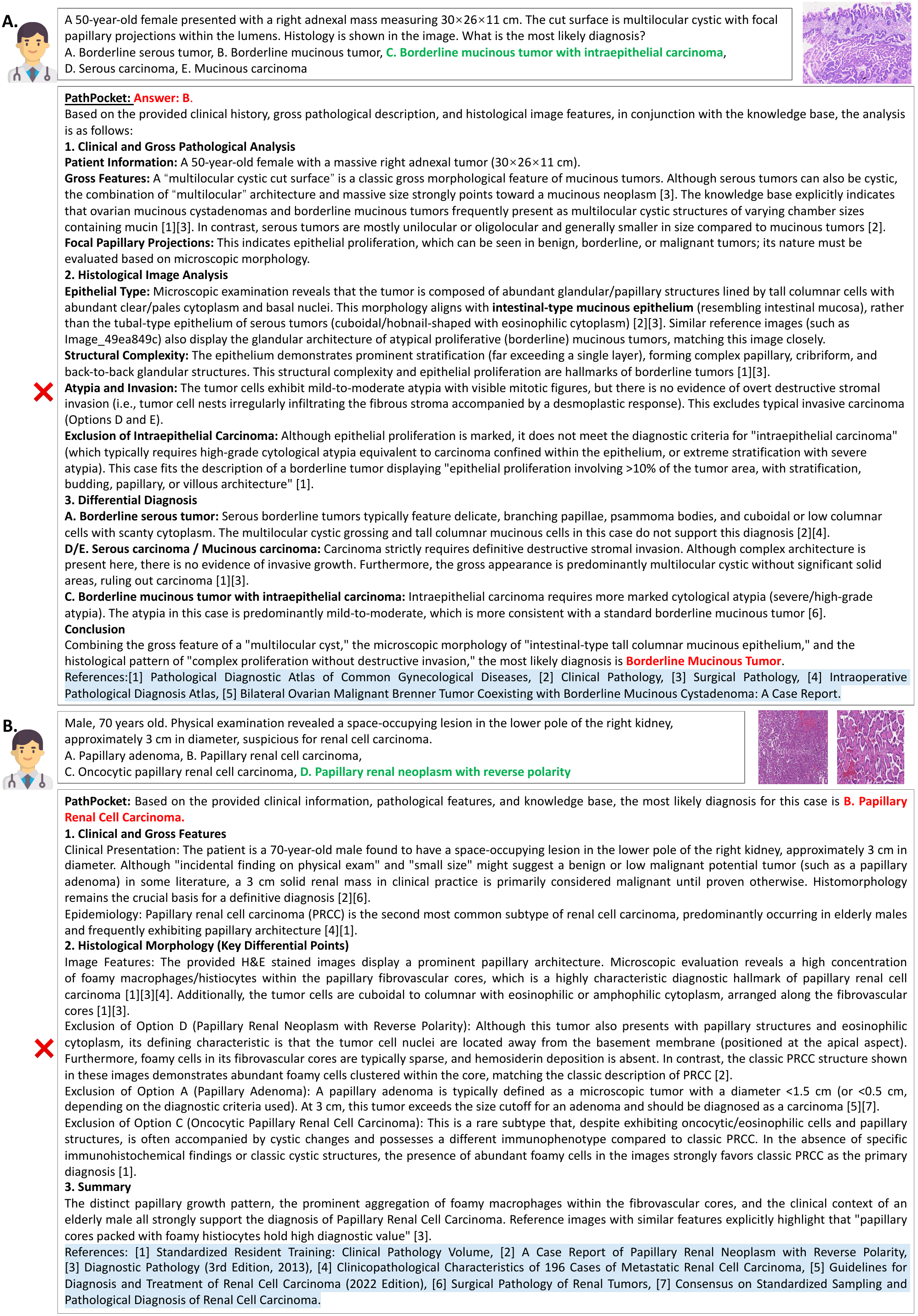}
\end{figure}

\begin{figure}[H]
    \centering
    \caption{\textbf{Representative PathPocket failure cases on hard near-neighbor differentials.}
    Full response transcripts for two incorrect predictions in which the chosen option remains a clinically adjacent entity.
    \textbf{Top:} borderline mucinous tumor with intraepithelial carcinoma is under-called as ordinary borderline mucinous tumor after PathPocket judges the atypia insufficient for intraepithelial carcinoma.
    \textbf{Bottom:} papillary renal neoplasm with reverse polarity is misclassified as classic papillary renal cell carcinoma, guided by foamy-macrophage cores that are a PRCC cue.
    Red crosses mark incorrect answers. These cases exemplify residual errors that evidence retrieval alone does not resolve when morphologic criteria are intrinsically borderline.}
    \label{fig:failure_cases}
\end{figure}

\clearpage

\begin{table}[H]
  \centering
  \captionsetup{width=\textwidth, justification=centering, singlelinecheck=false}
  \caption{\textbf{Characteristics of participating pathologists in the PathPocket reader study.}}
  \label{tab:RS_pathologist}
  \begin{tabular}{lr}
    \toprule
    \textbf{Characteristics} & \textbf{Num.} \\
    \midrule
    \textbf{Total pathologists} & 8 \\
    \textbf{Experience level} & \\
    \hspace{1em} Senior & 4 \\
    \hspace{1em} Junior & 4 \\
    \textbf{Sequence group allocation} & \\
    \hspace{1em} Group A (PathPocket in Phase 1) & 4 \\
    \hspace{1em} Group B (PathPocket in Phase 2) & 4 \\
    \textbf{Experience by sequence group} & \\
    \hspace{1em} Group A: Senior / Junior & 2 / 2 \\
    \hspace{1em} Group B: Senior / Junior & 2 / 2 \\
    \bottomrule
  \end{tabular}
\end{table}

\begin{table}[H]
  \centering
  \captionsetup{width=\textwidth, justification=centering, singlelinecheck=false}
  \caption{\textbf{Characteristics of the reader-study question bank and response records.} The study comprised 150 unique questions (100 text-only + 50 multimodal). Each text-only question has two independently scored sub-questions and is counted as two scored items (100$\times$2 = 200), yielding 250 scored items and 2,000 pathologist-level response records (250 scored items $\times$ 8 pathologists).}
  \label{tab:RS_questions}
  \begin{tabular}{lr}
    \toprule
    \textbf{Characteristics} & \textbf{Num.} \\
    \midrule
    \textbf{Unique questions} & 150 \\
    \hspace{1em} Text-only & 100 \\
    \hspace{1em} Multimodal & 50 \\
    \textbf{Scored items} & 250 \\
    \hspace{1em} Text-only (100 questions $\times$ 2 sub-questions) & 200 \\
    \hspace{1em} Multimodal & 50 \\
    \textbf{Response records} & 2000 \\
    \midrule
    \textbf{By modality (response records)} & \\
    \hspace{1em} Text-only response records & 1600 \\
    \hspace{1em} Multimodal response records & 400 \\
    \midrule
    \textbf{By question type (response records)} & \\
    \hspace{1em} Open-ended response records & 1760 \\
    \hspace{1em} Multiple-choice response records & 240 \\
    \midrule
    \textbf{By phase (unique questions per bank)} & \\
    \hspace{1em} Text-only, Phase 1 / Phase 2 & 50 / 50 \\
    \hspace{1em} Multimodal, Phase 1 / Phase 2 & 25 / 25 \\
    \bottomrule
  \end{tabular}
\end{table}

\clearpage

\begin{table}[H]
  \centering
  \captionsetup{width=\textwidth, justification=centering, singlelinecheck=false}
  \caption{\textbf{Primary outcome: normalized answer accuracy with and without PathPocket assistance.}}
  \label{tab:RS_primary}
  \begin{adjustbox}{max width=0.88\textwidth,center}
  \begin{threeparttable}
    \begin{tabular}{lcccc}
      \toprule
      & \textbf{N\tnote{1}} & \textbf{w/o PathPocket} & \textbf{w/ PathPocket} & \textbf{P value\tnote{2}} \\
      \midrule
      \textbf{Overall accuracy} & 1000 & 80.4\% & 85.8\% & 0.020 \\
      \midrule
      \textbf{By modality} & & & & \\
      \hspace{1em} Text-only & 800 & 83.8\% & 89.1\% & 0.042 \\
      \hspace{1em} Multimodal & 200 & 66.8\% & 72.4\% & 0.018 \\
      \midrule
      \textbf{By question type} & & & & \\
      \hspace{1em} Open-ended & 880 & 82.0\% & 87.0\% & 0.034 \\
      \hspace{1em} Multiple choice & 120 & 68.3\% & 76.7\% & 0.014 \\
      \bottomrule
    \end{tabular}
    \begin{tablenotes}
      \small
      \item[1] N = number of scored response records per condition. Text-only N includes two independently scored sub-questions per parent item (100 text-only questions $\times$ 2 = 200 scored items; 800 pathologist-level records per condition).
      \item[2] One-sided paired $t$-test on pathologist-level mean accuracy (w/ vs.\ w/o PathPocket).
      \item[3] Accuracy = GPT-5 mini normalized score (0--100\%). Open-ended answers scored on a 1--5 rubric and divided by 5; each text-only sub-question enters the mean as its own scoring unit.
    \end{tablenotes}
  \end{threeparttable}
  \end{adjustbox}
\end{table}

\begin{table}[H]
  \centering
  \captionsetup{width=\textwidth, justification=centering, singlelinecheck=false}
  \caption{\textbf{Secondary outcome: self-reported answer confidence (1--10) with and without PathPocket assistance.}}
  \label{tab:RS_secondary}
  \begin{adjustbox}{max width=0.82\textwidth,center}
  \begin{threeparttable}
    \begin{tabular}{lcccc}
      \toprule
      & \textbf{w/o PathPocket} & \textbf{w/ PathPocket} & \textbf{Mean difference\tnote{1}} & \textbf{P value\tnote{2}} \\
      \midrule
      \textbf{Overall confidence} & 7.2 & 8.1 & +0.8 & 0.026 \\
      \midrule
      \textit{By modality} & & & & \\
      \hspace{1em} Text-only & 7.3 & 8.1 & +0.9 & 0.022 \\
      \hspace{1em} Multimodal & 7.1 & 7.8 & +0.7 & 0.089 \\
      \midrule
      \textit{By question type} & & & & \\
      \hspace{1em} Open-ended & 7.2 & 8.1 & +0.9 & 0.021 \\
      \hspace{1em} Multiple choice & 7.2 & 7.9 & +0.7 & 0.144 \\
      \bottomrule
    \end{tabular}
    \begin{tablenotes}
      \small
      \item[1] Mean difference = mean confidence with PathPocket $-$ mean confidence without PathPocket (same units, 1--10 scale).
      \item[2] One-sided paired $t$-test on pathologist-level mean confidence.
    \end{tablenotes}
  \end{threeparttable}
  \end{adjustbox}
\end{table}

\clearpage

\begin{table}[H]
  \centering
  \captionsetup{width=\textwidth, justification=centering, singlelinecheck=false}
  \caption{\textbf{Individual pathologist accuracy and confidence with and without PathPocket assistance.}}
  \label{tab:RS_individual}
  \begin{adjustbox}{max width=0.98\textwidth,center}
  \begin{threeparttable}
    \begin{tabular}{lllcccccc}
      \toprule
      \textbf{ID} & \textbf{Exp.\tnote{1}} & \textbf{Seq.\tnote{2}} & \textbf{Accuracy w/o} & \textbf{Accuracy w/} & \textbf{$\Delta$Acc, pp\tnote{3}} & \textbf{Confidence w/o} & \textbf{Confidence w/} & \textbf{$\Delta$Conf\tnote{4}} \\
      \midrule
      P1 & Junior & A & 83.7\% & 89.3\% & +5.6 & 6.0 & 7.0 & +1.1 \\
      P2 & Junior & A & 81.9\% & 80.8\% & -1.1 & 8.1 & 8.4 & +0.3 \\
      P3 & Junior & B & 65.6\% & 83.2\% & +17.6 & 7.0 & 8.6 & +1.6 \\
      P4 & Junior & B & 81.6\% & 84.2\% & +2.6 & 6.0 & 4.6 & -1.4 \\
      P5 & Senior & A & 81.3\% & 86.1\% & +4.8 & 8.1 & 8.9 & +0.9 \\
      P6 & Senior & A & 88.2\% & 89.8\% & +1.6 & 6.6 & 8.3 & +1.7 \\
      P7 & Senior & B & 84.5\% & 85.8\% & +1.3 & 8.3 & 9.7 & +1.4 \\
      P8 & Senior & B & 76.2\% & 87.4\% & +11.2 & 7.9 & 9.2 & +1.2 \\
      \midrule
      \textbf{Mean} & & & 80.4\% & 85.8\% & +5.4 & 7.2 & 8.1 & +0.8 \\
      \bottomrule
    \end{tabular}
    \begin{tablenotes}
      \small
      \item[1] Exp.\ = experience level (Junior: P1--P4; Senior: P5--P8).
      \item[2] Seq.\ = sequence group (A: PathPocket in Phase 1; B: PathPocket in Phase 2).
      \item[3] $\Delta$Acc = accuracy with PathPocket $-$ accuracy without PathPocket, in percentage points.
      \item[4] $\Delta$Conf = mean confidence with PathPocket $-$ mean confidence without PathPocket (1--10 scale).
      \item[5] 7 of 8 pathologists improved in overall accuracy, and 7 of 8 improved in confidence, with PathPocket.
    \end{tablenotes}
  \end{threeparttable}
  \end{adjustbox}
\end{table}

\begin{table}[H]
  \centering
  \captionsetup{width=\textwidth, justification=centering, singlelinecheck=false}
  \caption{\textbf{Accuracy and confidence by anatomy system with and without PathPocket assistance.}}
  \label{tab:RS_organ}
  \begin{adjustbox}{max width=0.98\textwidth,center}
  \begin{threeparttable}
    \begin{tabular}{lccccccc}
      \toprule
      \textbf{Anatomy system} & \textbf{N\tnote{1}} & \textbf{Acc.\ w/o} & \textbf{Acc.\ w/} & \textbf{$\Delta$Acc, pp} & \textbf{Conf.\ w/o} & \textbf{Conf.\ w/} & \textbf{$\Delta$Conf\tnote{2}} \\
      \midrule
      Cardiovascular & 20 & 83.0\% & 96.0\% & +13.0 & 7.3 & 8.3 & +1.0 \\
      Other & 84 & 61.0\% & 72.6\% & +11.7 & 7.0 & 7.6 & +0.7 \\
      Head and Neck & 28 & 79.3\% & 90.0\% & +10.7 & 7.5 & 8.3 & +0.8 \\
      Soft Tissue & 64 & 85.6\% & 93.4\% & +7.8 & 7.2 & 7.8 & +0.7 \\
      Urinary & 68 & 75.0\% & 82.4\% & +7.4 & 7.4 & 8.2 & +0.8 \\
      Bone and Joint & 32 & 75.0\% & 81.2\% & +6.2 & 6.6 & 7.6 & +1.0 \\
      Breast & 80 & 79.0\% & 84.2\% & +5.2 & 7.8 & 8.4 & +0.5 \\
      Gynecologic & 100 & 86.4\% & 91.4\% & +5.0 & 7.3 & 8.2 & +0.9 \\
      Lymphohematopoietic & 84 & 87.6\% & 92.4\% & +4.8 & 7.0 & 8.1 & +1.1 \\
      Gastrointestinal & 168 & 75.7\% & 79.5\% & +3.8 & 7.4 & 8.0 & +0.7 \\
      Respiratory & 124 & 86.5\% & 90.0\% & +3.5 & 7.3 & 8.2 & +0.9 \\
      Endocrine & 44 & 83.6\% & 85.9\% & +2.3 & 7.1 & 8.1 & +1.1 \\
      Nervous & 52 & 89.2\% & 91.2\% & +1.9 & 7.0 & 8.3 & +1.2 \\
      Skin & 52 & 82.7\% & 84.6\% & +1.9 & 6.9 & 8.0 & +1.1 \\
      \bottomrule
    \end{tabular}
    \begin{tablenotes}
      \small
      \item[1] N = number of scored response records per condition within each anatomy system.
      \item[2] $\Delta$Conf = mean confidence with PathPocket $-$ mean confidence without PathPocket (1--10 scale).
    \end{tablenotes}
  \end{threeparttable}
  \end{adjustbox}
\end{table}

\clearpage

\begin{table}[H]
  \centering
  \captionsetup{width=\textwidth, justification=centering, singlelinecheck=false}
  \caption{\textbf{Inter-pathologist agreement with and without PathPocket assistance.}}
  \label{tab:RS_interrater}
  \begin{adjustbox}{max width=0.92\textwidth,center}
  \begin{threeparttable}
    \begin{tabular}{lccc}
      \toprule
      \textbf{Metric} & \textbf{w/o PathPocket} & \textbf{w/ PathPocket} & \textbf{$\Delta$\tnote{1}} \\
      \midrule
      Open-ended ICC(2,$k$) & 0.549 & 0.719 & +0.170 \\
      Open-ended tolerance agreement ($\leq$1 rubric step) & 76.4\% & 86.7\% & +10.2 pp \\
      Multiple-choice Fleiss' $\kappa$ (A--E) & 0.466 & 0.692 & +0.226 \\
      \bottomrule
    \end{tabular}
    \begin{tablenotes}
      \small
      \item[1] $\Delta$ = metric with PathPocket $-$ metric without PathPocket.
      \item[2] ICC(2,$k$)\cite{shrout1979intraclass}: average-measure two-way random-effects ICC on pathologist-aligned score matrices (weighted across crossover rater cohorts). Tolerance agreement = pairwise fraction with $|$normalized score difference$| \leq 0.2$ (one step on the 1--5 rubric). Fleiss' $\kappa$\cite{fleiss1971measuring} computed across pathologists within each multiple-choice item.
    \end{tablenotes}
  \end{threeparttable}
  \end{adjustbox}
\end{table}

\begin{table}[H]
  \centering
  \captionsetup{width=\textwidth, justification=centering, singlelinecheck=false}
  \caption{\textbf{Absolute accuracy gap between senior and junior pathologists.}}
  \label{tab:RS_experience_gap}
  \begin{adjustbox}{max width=0.72\textwidth,center}
  \begin{threeparttable}
    \begin{tabular}{lccc}
      \toprule
      \textbf{Condition} & \textbf{Junior mean} & \textbf{Senior mean} & \textbf{$|$Senior $-$ Junior$|$} \\
      \midrule
      w/o PathPocket & 78.2\% & 82.5\% & 4.3\% \\
      w/ PathPocket & 84.4\% & 87.2\% & 2.9\% \\
      \bottomrule
    \end{tabular}
    \begin{tablenotes}
      \small
      \item[1] Gap narrowed from 4.3\% to 2.9\% (32.6\% relative reduction).
    \end{tablenotes}
  \end{threeparttable}
  \end{adjustbox}
\end{table}

\clearpage

\begin{table}[H]
  \centering
  \captionsetup{width=\textwidth, justification=centering, singlelinecheck=false}
  \caption{\textbf{PathPocket assistance content quality ratings reported by pathologists during assisted phases.}}
  \label{tab:RS_content_rating}
  \begin{adjustbox}{max width=0.7\textwidth,center}
  \begin{threeparttable}
    \begin{tabular}{lc}
      \toprule
      \textbf{Metric} & \textbf{Value} \\
      \midrule
      Overall mean rating (1--10) & 8.94 \\
      Overall median rating & 9 \\
      Rated assisted responses, n & 1000 \\
      \midrule
      \textbf{Mean rating by pathologist} & \\
      \hspace{1em} P1 & 7.64 \\
      \hspace{1em} P2 & 9.44 \\
      \hspace{1em} P3 & 8.68 \\
      \hspace{1em} P4 & 8.17 \\
      \hspace{1em} P5 & 9.51 \\
      \hspace{1em} P6 & 9.16 \\
      \hspace{1em} P7 & 9.45 \\
      \hspace{1em} P8 & 8.40 \\
      \bottomrule
    \end{tabular}
    \begin{tablenotes}
      \small
      \item[1] Ratings were collected only in PathPocket-assisted phases.
    \end{tablenotes}
  \end{threeparttable}
  \end{adjustbox}
\end{table}

\end{document}